\documentclass{article}

\usepackage{arxiv}

\usepackage[utf8]{inputenc} 
\usepackage[T1]{fontenc}    
\usepackage{hyperref}       
\usepackage{url}            
\usepackage{booktabs}       
\usepackage{amsfonts}       
\usepackage{nicefrac}       
\usepackage{microtype}      
\usepackage{lipsum}
\usepackage{graphicx}
\usepackage{doi}

\usepackage{subfig}
\usepackage{amsmath,amssymb,amsfonts}
\usepackage{nicefrac}
\usepackage{enumitem}
\usepackage{xcolor}
\usepackage{booktabs}
\usepackage{multirow}
\usepackage{bbm}
\usepackage{placeins}
\usepackage{marvosym}
\usepackage{scalerel}
\usepackage{textcomp}
\usepackage{verbatim}
\usepackage{hyperref}

\usepackage{mathtools}

\usepackage{cite}  
\usepackage{graphicx,array}
\graphicspath{{figures/}{pictures/}{images/}{./}} 

\SetLabelAlign{myright}{\hss\llap{$#1$}}
\newlist{where}{description}{1}
\setlist[where]{labelwidth=2cm,labelsep=0.5em,leftmargin=!,align=myright,font=\normalfont}

\title{PheroCom: Decentralised and asynchronous swarm robotics coordination based on virtual pheromone and vibroacoustic communication}

\author{ \href{https://orcid.org/0000-0003-2540-1735}{\includegraphics[scale=0.06]{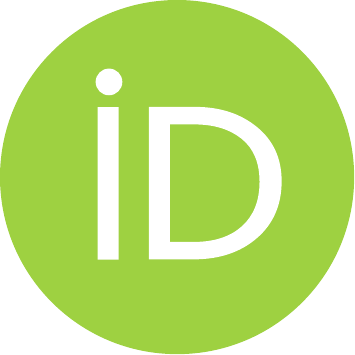}\hspace{1mm}Claudiney R. Tinoco}\thanks{This work was supported by the Funda\c{c}\~{a}o de Amparo \`{a} Pesquisa do Estado de Minas Gerais (FAPEMIG), Conselho Nacional de Desenvolvimento Cient\'{i}fico e Tecnol\'{o}gico (CNPq) and Coordena\c{c}\~{a}o de Aperfei\c{c}oamento de Pessoal de N\'{i}vel Superior (CAPES).} \\
	School of Computer Science\\
	Federal University of Uberl\^andia\\
	Uberl\^andia/MG, Brazil\\
	\texttt{claudineyrt@gmail.com} \\
	\And
	\href{https://orcid.org/0000-0003-0384-1879}{\includegraphics[scale=0.06]{orcid.pdf}\hspace{1mm}Gina M. B. Oliveira} \\
	School of Computer Science\\
	Federal University of Uberl\^andia\\
	Uberl\^andia/MG, Brazil\\
	\texttt{gina@ufu.br}
}

\begin{document}
\maketitle

\begin{abstract}
Representation and control of the dynamics of stigmergic substances used by bio-inspired approaches is a challenge when applied to robotics. In order to overcome this challenge, this work proposes a model to coordinate swarms of robots based on the virtualisation and control of these substances in a local scope. The model presents a new pheromone modelling, which enables the decentralisation and asynchronicity of navigation decisions. Each robot maintains an independent virtual pheromone map, which is continuously updated with the robot’s deposits and pheromone evaporation. Moreover, the individual pheromone map is also updated by aggregating information from other robots that are exploring nearby areas. Thus, individual and independent maps replace the need of a centralising agent that controls and distributes the pheromone information, which is not always practicable. Pheromone information propagation is inspired by ants' vibroacoustic communication, which, in turn, is characterised as an indirect communication through a type of gossip protocol. The proposed model was evaluated through an agent simulation software, implemented by the authors, and in the Webots platform. Experiments were carried out to validate the model in different environments, with different shapes and sizes, as well as varying the number of robots. The analysis of the results has shown that the model was able to perform the coordination of the swarm, and the robots have exhibited an expressive performance executing the surveillance task.
\end{abstract}


\section{Introduction}
\label{intro}
Centralisation makes it easy to control shared information. Furthermore, it ensures that the information is fully current and consistent, producing a decision-making process more accurate. However, centralisation is not a desired feature in swarm robotics, since robots must be independent and make their decisions based only on their knowledge and information received indirectly \cite{brambilla2013swarm}. On the other hand, a decentralised model warrants that the main characteristics of swarm intelligence are fulfilled: robustness, scalability and flexibility \cite{camazine2003self}. Robustness is related to the ability to keep the task performing, even when a member of the swarm fails or an external interference occurs. This is because the system is redundant, i.e., if a robot fails, its task would be performed by another robot. Flexibility refers to the ability of the swarm to adapt to different tasks and environments. Finally, scalability is related to the ability of the swarm to perform the task satisfactorily with different numbers of robots. In other words, even whether robots are removed from the swarm or new robots are inserted, the ability to perform the task would be maintained. Thus, decentralised robotics models have decentralised and asynchronous coordination, robots are simple, and sensing capabilities are distributed.

Bio-inspired strategies represent a large part of the research in the related literature, e.g., ant pheromone-inspired models \cite{sauter2005performance, bontzorlos2017bioinspired} and Cellular Automata (CA) based models \cite{ioannidis2011cellular, sirakoulis2015robots, lima2017cellular}. A review of several applications of CA models in robotics is presented by \citeonline{ferreira2014improved}. Recently, CA-based models for swarm robotics and multi-robot systems have been investigated for navigation under formation control \cite{oliveira2018evolutionary}, foraging \cite{lima2017cellular}, crowd evacuation \cite{boukas2015robot} and path-planning \cite{chaves2017improved, martins2018improved}. In our previous work, the IACA-DI \cite{tinoco2017improved} model was proposed to coordinate swarms of robots, based on two bio-inspired techniques: Inverted Ant System (IAS)~\cite{calvo2012bioinspired} and CA. The IACA-DI model implements a centralising robot to virtually represent the pheromone dynamics, and this robot is responsible for synchronising the information of the environment in which the task is performed and where the robots detect and deposit the pheromone. This approach was proposed as an alternative to mimic the pheromone dynamics in the environment. However, as mentioned before, centralisation and synchronisation are not desirable characteristics for swarm robotics, resulting in practical restrictions for the real implementation.

Several researches seek to develop efficient ways to mimic pheromones. In the literature, it is possible to find works involving the use of volatile substances, RFID tags, intelligent environments and virtual pheromones~\cite{trianni2006self}. However, most of these techniques have some degree of difficulty to be implemented, are expensive or are specific to a particular task, which results in low adaptability~\cite{cambier2020language}. Besides, this would reduce the autonomy of the swarm, especially in those techniques that have some type of centralisation (e.g., intelligent environments and control servers), since it would be limited to only one type of scenario.

Considering the adversities, previously presented, of reproduce the dynamics of stigmergic substances, this work proposes a decentralised and asynchronous method to represent and control the dynamics of pheromones applied to the coordination of swarms of robots. Denominated as PheroCom, the model seeks to mimic the pheromone dynamics through local virtual maps stored in the memory of each robot. The global behaviour emerges once the robots propagate information about their local pheromone maps. Besides, information dissemination is bio-inspired in ant's vibroacoustic communication, which maintains the main characteristics of the swarm intelligence.

The organisation of this paper is as follows: Sect.~\ref{sec:theoretical-framework} presents the Theoretical Framework, describing some pheromone mimicry techniques found in the literature, the biological inspiration of the model, and the description of the surveillance task. In Sect.~\ref{sec:pherocom-model} the PheroCom model is proposed and specified in detail, presenting the individual control mechanism, the interface of each robot to interact with the external environment, the global propagation of the pheromone information and the communication strategy. Experimental results are discussed in Sect.~\ref{sec:experiments-analysis}. Section~\ref{sec:assessment-discussion} presents an overview of the model, with its advantages, disadvantages and limitations. Section \ref{conclusions} presents the main conclusions and future works.

\section{Theoretical Framework}
\label{sec:theoretical-framework}
The theoretical basis is divided into three points: related works, biological inspiration and surveillance in Robotics. In the Related Works, it is analysed some important techniques linked to pheromone mimicry in robotics. The Biological Inspiration describes the inspiration bases used in the proposition of our model. Finally, Surveillance in Robotics describes the surveillance task, since it is the chosen task to be analysed in the experiments.

\subsection{Related Works}
\label{sec:literature-review}
The mimicry of the dynamics of pheromones (deposition, sensing, meaning and evaporation) is the object of several researches. The main approaches to this purpose use RFID tags, volatile chemicals, centralised control servers, intelligent environments, pheromone communication by light and sound sources, and, as in this work, the virtualisation of the pheromone.

The use of RFID tags is extremely widespread. \citeonline{mamei2007pervasive} presented the implementation of a pheromone-based mechanism using RFID tags. According to the authors, the mechanism is simple, low-cost and general-purpose. It was validated through the object-tracking task, in which a prototype of the mechanism showed its feasibility. In turn, \citeonline{khaliq2015stigmergy} proposed a stigmergic approach to goal-directed navigation in indoor environments. Through the use of RFID tags, e-puck robots store pheromone information in the tags to guide them towards their goals. The authors stated that the results proved to be robust and easy to be replicated. It is possible to notice that RFID tags have a distributed feature and are capable of storing pheromone information. However, some problems arise in this application: environments must be known and pre-prepared, and there is no possibility of reading and writing by area, only one tag at a time.

Other proposals include the use of volatile chemicals. \citeonline{purnamadjaja2007guiding} investigated the mimicking of pheromones using tin oxide gas in the light seeking task. According to its results, it was possible to observe that chemical signals can be used as a form of communication and directing the robots's actions. In the work of \citeonline{fujisawa2014designing}, it was proposed the use of chemical compounds in the foraging task. \citeonline{salman2020phormica} proposes a strategy in which the floor of the environment is coated with a photochromic substance to implementing stigmergic coordination. One can notice that the use of chemical substances to mimic pheromones, is the closest technique to the biological inspiration. However, it becomes unfeasible in most environments, since it requires the actual spread of some chemical substance, to be a known environment, the cost generated, and the risks arising from some of these substances being, for example, flammable or toxic.

Intelligent environments, in swarm robotics, are environments capable of visually simulate pheromones. \citeonline{sugawara2004foraging} proposes in his work the simulation of chemical signals through a graphic projection on the floor of the environment. \citeonline{arvin2015cosvarphi} and \citeonline{na2019extended} proposed an artificial pheromone system based on the use of a pheromone screen and a centralising camera. It can be seen that intelligent environments allow pheromone mimicry without the inherent difficulties of chemical substances. However, these environments are only possible to be used in known and controlled environments. Furthermore, these environments have a high cost of implementation, and again brings up the problem of centralisation.

Furthermore, many authors seek to virtualise the pheromone to facilitate its mimicry. In \citeonline{payton2001pheromone}, the idea of virtual pheromone was introduced in the proposition of a system to control multiple robots. Pheromone-related messages are transmitted through infrared communication. Similar works have shown variations of virtual pheromones, e.g., in coordinating robots in finding short paths \cite{campo2010artificial} and in controlling drones to cover unknown environments \cite{aznar2018modelling}. However, these models do not present a decentralised approach to the pheromone virtualisation. Other techniques are very promising in this field \cite{rubenstein2012kilobot, turkoral2017indoor}. In the work of \citeonline{song2020novel}, it was proposed pheromone model for swarm foraging based on neural networks. The authors have developed an optimisation method to determine the parameters of the robots performing the foraging task. Results verified the effectiveness of their model.

\subsection{Biological Inspiration}
\label{sec:bio-inspiration}
In nature, some animals have developed the ability to live in groups to achieve a common goal, since it would not be possible to be reached by individual efforts. They are called ``social animals''~\cite{silk2017understanding}.

Among these groups or societies, we can highlight the groups of birds and fishes. Some species of birds form flocks to increase foraging speed and safety from predators, keep them warm at low temperatures, and decrease air resistance when flying at ``V'' formation, which increases energy conservation in migration~\cite{gill2007ornithology}. Likewise, some species of fishes form schools to resemble a larger animal and ward off predators~\cite{partridge1982structure, parrish2002self}. Many insects also reproduce this behaviour of living in societies, forming extremely organised groups often regarded as superorganisms~\cite{wheeler1928social, wilson1989reviving}. It is important to emphasise that there is no central individual coordinating the actions of each insect. Each individual of a social insect colony interacts locally with other individuals, and this local interaction generates a global behaviour. Ants are an example of social insects. Through local interactions, they can accomplish complex tasks that only one ant would be unable to perform on its own. Nest building, food foraging, and the defence of the colony against invaders are all examples of collective tasks \cite{holldobler1990ants, detrain2006self, heyman2017ants}. Bees also live in societies: defending their nests, increasing foraging area and dividing labour are some examples of tasks they develop to live in colonies \cite{seeley1989honey, seeley2009wisdom}. There are several other types of social insects, for example, wasps, termites, fireflies, firebugs, etc.~\cite{wilson1971insect}.

\begin{figure}[t]
	\centering
	\includegraphics[width=3in]{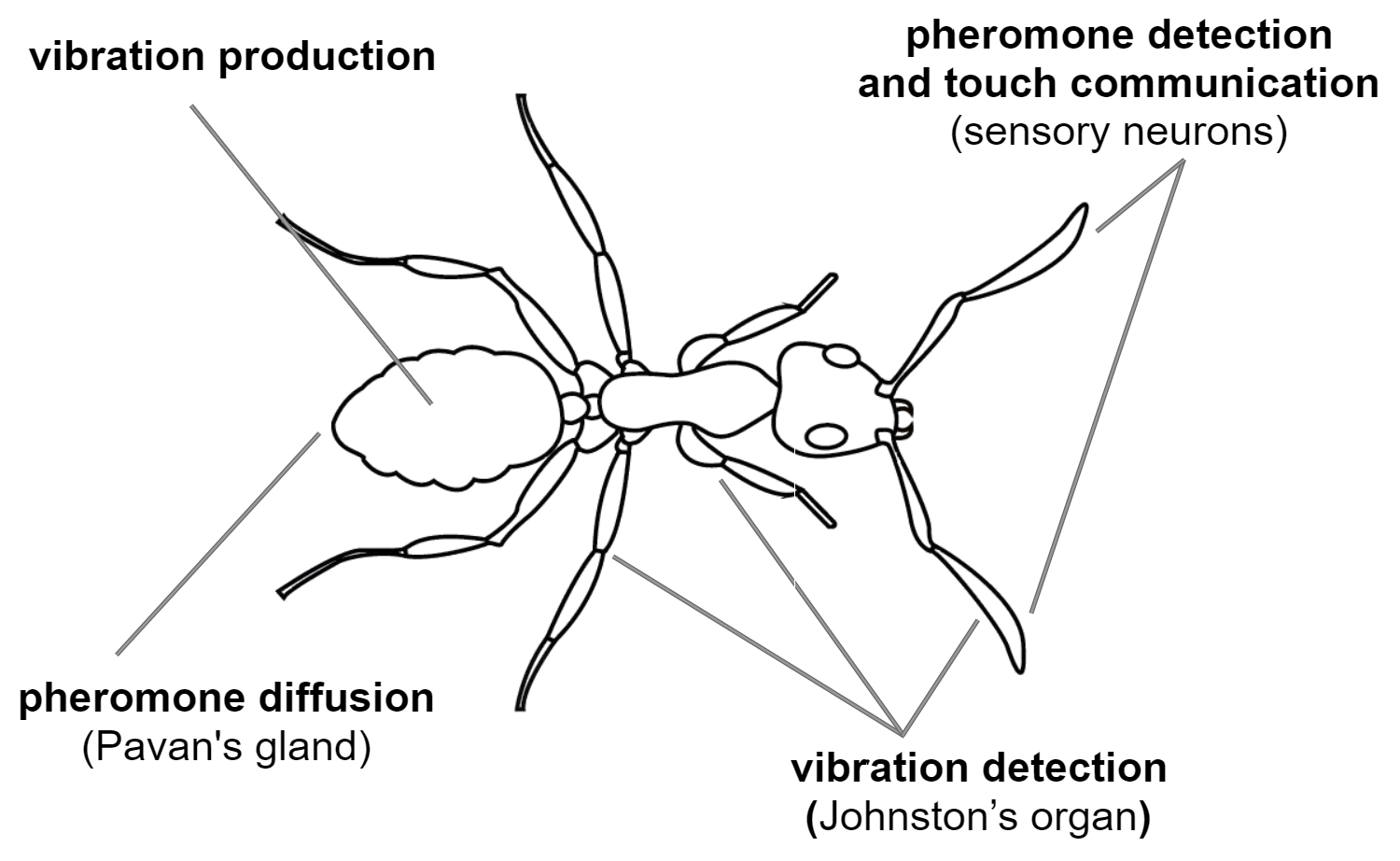}%
	\caption{Ants' communication mechanisms. Messages are transmitted through antenna touch, vibration production and pheromone deposition, and are received through antenna touch, vibration detection and pheromone detection [Based on \citeonline{hunt2013intracolony}].}
	\label{theory:ant_communication}
\end{figure}

Each different group of animals uses specific means of communication. In this work, the communication mechanisms used by the ants will be highlighted. The main communication mechanism used by them is the exchange of chemical signals. In addition to this mechanism, some ants also use vibroacoustic signals and antenna touch \cite {holldobler1999multimodal}. Figure~\ref{theory:ant_communication} illustrates each of the aforementioned communication mechanisms, pointing up in which part of the ants body these mechanisms are produced and detected.

Chemical communication is the primary communication mechanism used by ants. Ants produce certain chemicals and spread them throughout the environment to communicate~\cite{richard2013intracolony}. These substances, called pheromones, are produced in exocrine glands located at the end of ants' abdomen, and are detected in the environment through sensory neurons present on the tips of their antennae (Fig.~\ref{theory:ant_communication}). Pheromone was discovered and described by~\citeonline{karlson1959pheromones} and can be specified as a volatile chemical used for communication between individuals of the same species. Deposit-based communication and pheromone sensing are called stigmergy. In turn, stigmergy, defined by~\citeonline{grasse1959reconstruction}, is a type of indirect communication and involves making changes in the environment, in this case through pheromone deposits: ants deposit pheromones so that other ants of the swarm can interpret the message when they come in contact with it. As the pheromone is a volatile substance, its concentration at a given location indicates, in addition to the proper meaning given by its type, the amount of time that the pheromone has been deposited, where it is proportional to its concentration. In other words, the greater the amount of pheromone detected, the more recently an ant has passed and deposited it; on the other hand, the lower the pheromone concentration, the longer the site was not visited.

Another important communication mechanism is the vibroacoustic. Vibroacoustic communication is characterised by the exchange of information through the reproduction of sounds from vibrations. Ants produce sounds through stridulation (friction of body parts) and drumming (beating of the abdomen against the substrate/envi\-ronment). Sensing vibroacoustic signals is performed by specific organs in their paws~\cite{hunt2013intracolony}. Additionally, it is commonly applied as an alarm system, arousal indication and to enhance the effectiveness of pheromones~\cite{golden2016evolution}.

Considering the information previously presented, in this work, vibroacoustic communication is characterised as an indirect communication through a type of gossip protocol~\cite{haas2002gossip}. Communication is defined as indirect since it occurs while the ant is transmitting the message within a noticeable area (local communication), but it does not have a specific receiver (indirect). All ants within the transmission area, taking into account the availability of receiving messages, may use the information captured in their own decision-making process. Furthermore, most protocols based on gossiping aims to disseminate information in such a way that, in a given moment, all agents have the same information (it happens like the spreading of a virus in biological communities~\cite{demers1987epidemic}). However, in the case of vibroacoustic communication, it is sufficient that the information is disseminated, but it is not necessary that it propagates to the whole swarm. Therefore, the method of communication inspired by the transmission of information by vibroacoustic signals, being a protocol sub-type based on gossiping, is proposed and defined as Vibroacoustic Based Indirect Transmission (ViBIT).

\subsection{Surveillance in Robotics}
\label{sec:surveillance_robotics}
The investigations and analysis carried out in this work, take into account the surveillance task. This task was also applied in our previous works~\cite{tinoco2017improved, tinoco2018pheromone, tinoco2019heterogeneous, tinoco2020parameter}, which allows analytical comparisons. However, it is noteworthy that our proposed model can be applied to other tasks (e.g., foraging, coverage and search-and-rescue). In the surveillance task with robots, the objective (an indoor or outdoor environment) must be monitored for defined/undefined period of time~\cite{witwicki2017autonomous}. This monitoring is carried out to maintain, in most cases, safety against invaders. In general, the task can be divided into four phases: monitoring, detection, pursuit and capture. 

Here, it is investigated the monitoring phase in indoor environments. The goal is to spread the robots of the swarm throughout the environment as best as possible, ensuring that unmonitored areas become monitored, and monitored areas continue to be visited on a cyclical basis. In order to verify the effectiveness and efficiency of the swarm, its composition always takes place with a number of robots smaller than the number of rooms. This makes the movement between rooms necessary, seeking to optimise task execution.

\section{PheroCom Model}
\label{sec:pherocom-model}
Considering the centralisation of the IACA-DI model \cite{tinoco2017improved} and the bio-inspiration in social animals, this work proposes a new model to represent the pheromone dynamics in the environment in a decentralised and asynchronous way, with the main purpose of coordinating swarms of robots.

Denominated as PheroCom model, it seeks to solve the centralisation and synchronicity constraints of the IACA-DI model by virtualising the pheromone dynamics through decentralised local maps. In this approach, there is no need of a centralising agent (robot or control server) that manages the pheromone information, an intelligent environment or real chemical substances. Each robot has in its local memory a representation of the environment, and they are able to access and modify their own pheromone information (detection, deposition, evaporation and updating). Through the application of the ViBIT protocol (see Section~\ref{sec:bio-inspiration}), which is biologically inspired in ants' vibroacoustic communication, the local information regarding the pheromone is shared with other nearby robots. Information transmission causes the pheromone to propagate globally, characterising the symbiosis of the model.

Applying our bio-inspired communication mechanism (ViBIT) along with the pheromone dynamics is the essence of the proposed model. This is due to the fact that this mechanism guarantees the main swarm intelligence prerogatives: robustness, flexibility, and scalability. Since in vibroacoustic communication there is no need for a direct connection between ants, when mimicked, it allows the robots to be independent. Moreover, the model maintains a virtual representation of the pheromone, which facilitates its real-world reproduction, as the representation techniques proposed in other researches are expensive and/or difficult to reproduce in any type of environment (for examples of other techniques, see Section~\ref{sec:literature-review}).

\subsection{Individual control system}
\label{sec:fsm} 
\begin{figure}[ht]
	\centering
	\subfloat[Robot control mechanism represented by a FSM.]{
		\includegraphics[width=4.6in]{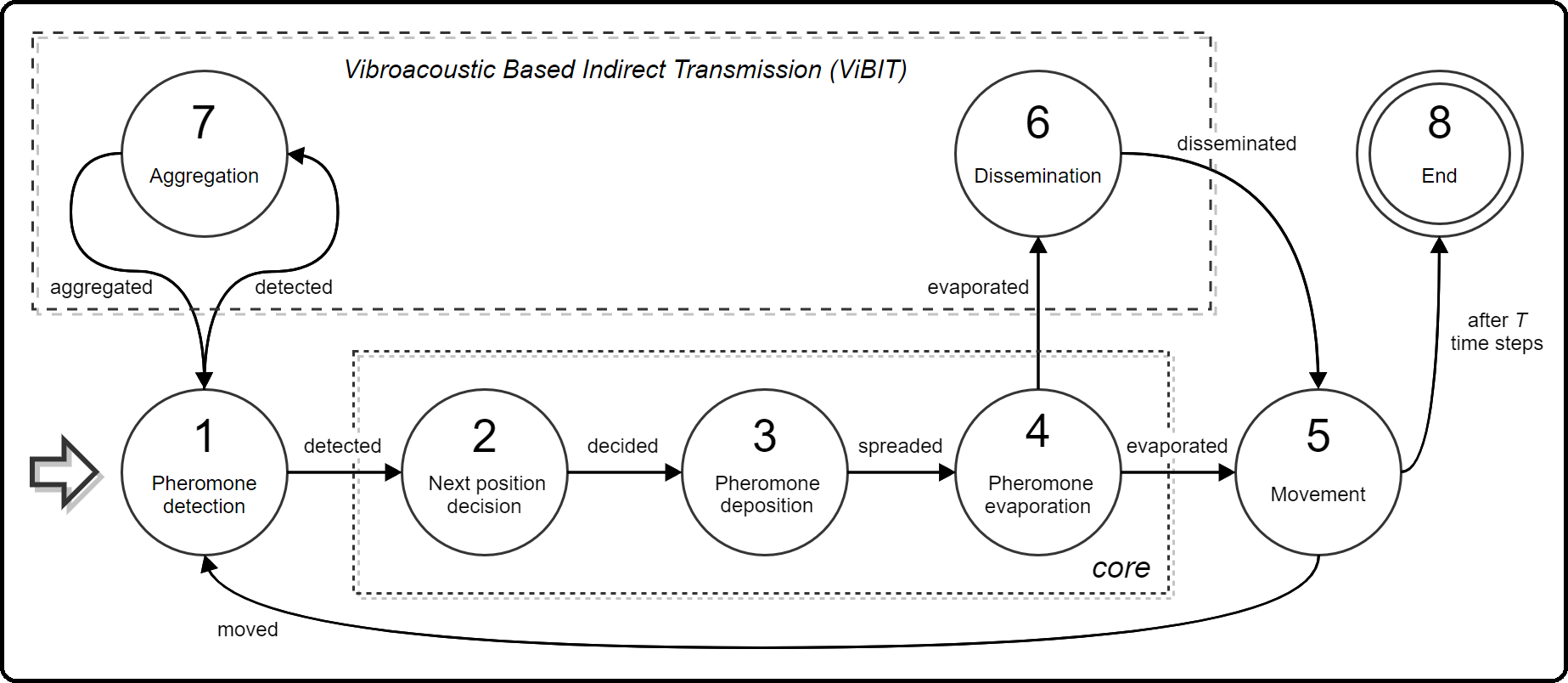}%
		\label{fig:fsm}}\\
	\subfloat[Physical grid.]{
		\includegraphics[width=1.8in]{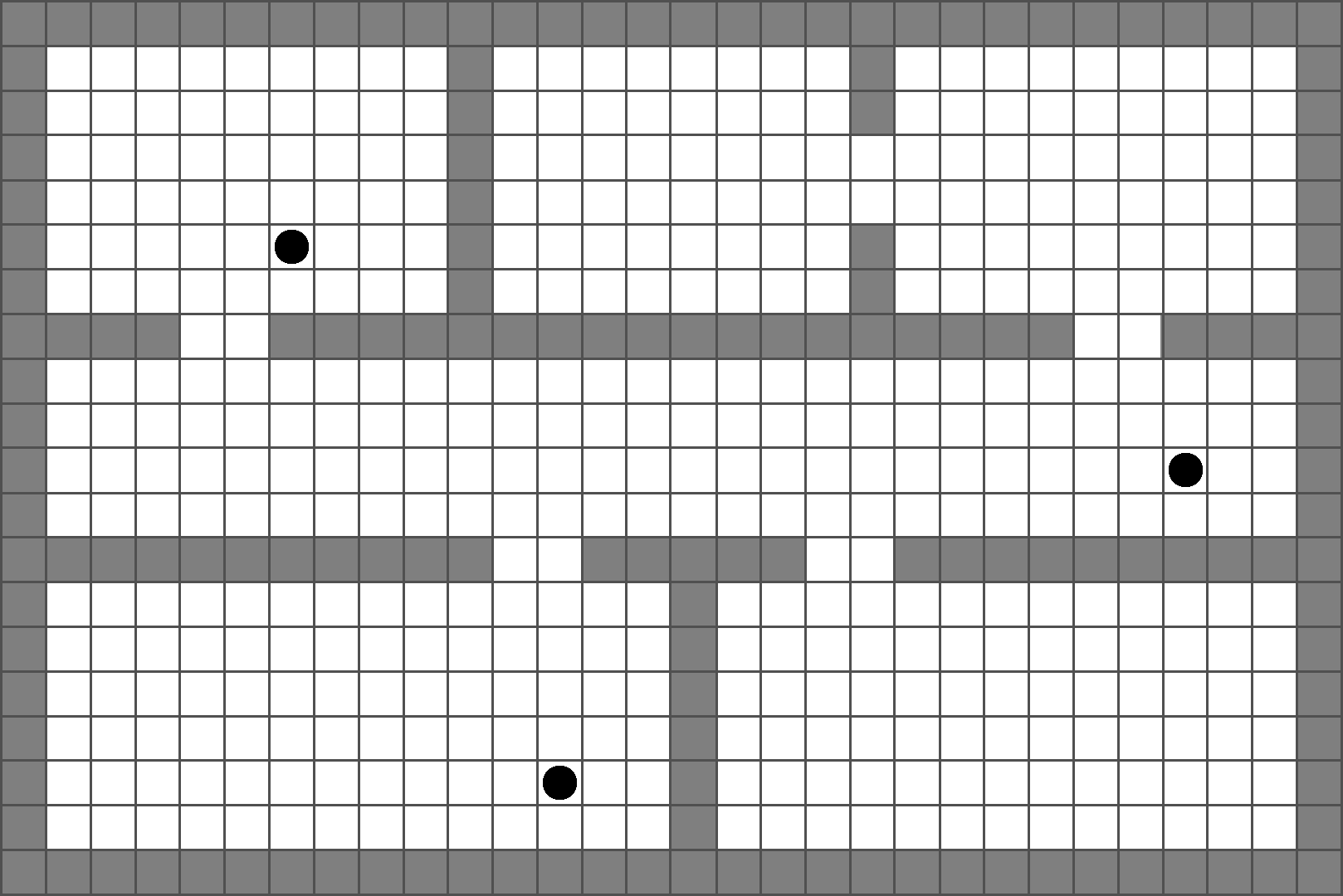}%
		\label{fig:physical_grid}}
	\hfil
	\subfloat[Pheromone grid.]{
		\includegraphics[width=1.8in]{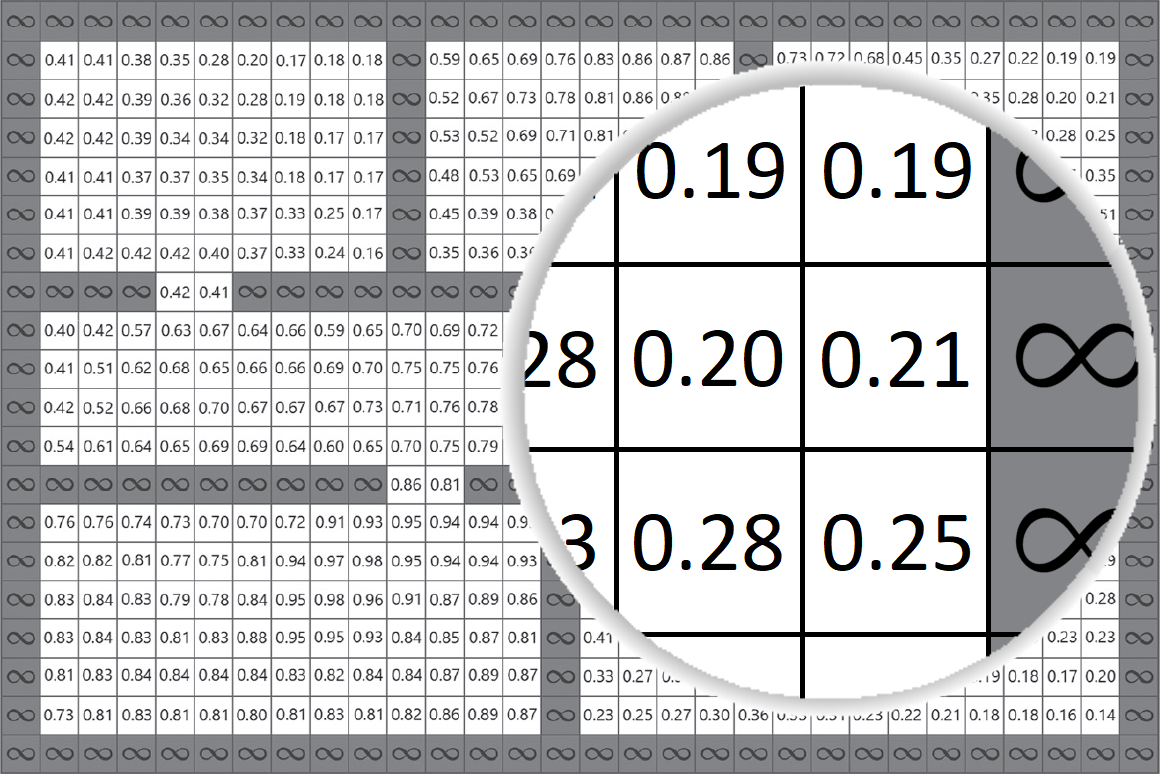}
		\label{fig:pheromone_grid}}
	\caption{Figure \ref{fig:fsm} describes the individual behaviour of each robot through a FSM with 8 states. Figures \ref{fig:physical_grid} and \ref{fig:pheromone_grid} shows examples of virtual grids/maps on which the FSM operates (grids with six rooms and size (20$\times$30) cells).}
	\label{fig:pherocom}
\end{figure}

The individual behaviour of the robots is defined through a Finite State Machine (FSM), shown in Figure~\ref{fig:fsm}. The FSM models the decision-making process of each robot when it is moving throughout the environment, and how/when it will disseminate information to other robots. Although these two processes (decision-making and information dissemination) have local scope, it is possible to generate a complex global behaviour capable of performing the proposed task. Each robot has an identical FSM, which guarantees autonomy and decentralisation in the swarm's decisions. A fully decentralised swarm reflects a strong robustness, which will ensure that the task continues to execute, even if just one active robot remains. 

The FSM consists of eight states: five cyclically executed representing interactions with the pheromone to decide individual motions and the movement itself, two on-demand states representing the communication system (applying the ViBIT protocol), and a final state. Each time a robot goes through all five states of the FSM's main cycle, a discrete CA time step is counted. Thus, even if the swarm is in a continuous environment for data processing, the FSM acts on discrete information, both in time and space. After this processing, the discrete information is transformed back into continuous, and then it is passed on to the robot's actuators to perform the physical movement.

Inside each robot's internal memory, the environment is discretised into two cell grids: a physical grid (Fig.~\ref{fig:physical_grid}) and a pheromone grid (Fig.~\ref{fig:pheromone_grid}). Each state of the FSM in Figure~\ref{fig:fsm} represents an action that must be performed considering these two grids. In this way, the global pheromone grid used in the IACA-DI model is replaced by asynchronously evolving local grids in each robot. The asynchronicity is due to the fact that each robot may be at a different time step on their internal machines, which, in turn, are not synchronised by a global clock (in the IACA-DI model, where a single pheromone map was used and shared, there was a need to synchronise the steps of all robots).

Some states of the PheroCom's FSM are based on the IACA-DI's FSM: states 1, 2, 3, 5 and 8. Thus, these pre-existing states are briefly exposed, and the new FSM states (states 4, 6 and 7), resulting from the new coordination model, proposed in this work, are presented and carefully described.

\subsubsection{States based on the IACA-DI model \{1, 2, 3, 5, 8\}}
In Pheromone Detection (State 1), each robot reads on its local map the pheromone concentration within its neighbourhood, which is delimited by its vision radius. The robots apply choice strategies (State 2) to decide which cells will be the destination of their next moves. Pheromone concentration in a given cell $x_{ij}$ at time step $t$, defines the probability $P(x_{ij})$ (in stochastic strategies) that this cell will be chosen at time step $(t+1)$. Five different choice strategies were evaluated in \citeonline{tinoco2018pheromone}: Random (all cells have the same probability of being chosen), Deterministic (the cell with the lowest pheromone concentration is always chosen), Simple Probabilistic (the probability of each cell is based on the pheromone concentration), Elitist Probabilistic (based on pheromone concentration, some cells are prioritised) and Inertial Probabilistic (in addition to elitism, the cell in the current direction of the robot is more likely to be selected). These strategies have been subjected to a large number of experiments and analyses in our previous works, and it was possible to define that a swarm composed of $\nicefrac{2}{3}$ of the robots with inertial strategy and $\nicefrac{1}{3}$ with deterministic strategy, outperforms other compositions~\cite{tinoco2019heterogeneous}. In turn, Pheromone Deposition (State 3) is the state in which each robot deposits pheromone, through Equation~\ref{eq:robotdeposition2}~\cite{tinoco2017improved}, in the environment within its deposition area. The objective is to inform all robots of the swarm that the area was monitored for an amount of time proportional to the concentration of pheromone detected. The state Movement (State 5) represents the robot's transition from an origin cell $x_{ij}$ to a target cell $x_{(i+a)(i+b)}$ in its neighbourhood, such that $\{a,b \in \mathbbm{Z}\}$. Finally, the FSM final state (State 8) is activated when the robot completes the proposed task or reaches a limit of $T$ time steps. It is worth considering that some tasks (e.g., surveillance), can be performed cyclically without a predefined ending moment.

\begin{equation}
	\begin{aligned}
	\Delta_{ij}^k = (\psi_{max} - \psi_{ij}^{t})\cdot \left[\alpha \cdot (\delta \cdot e)^{\eta \cdot \tfrac{r}{\pi}}\right]
	\end{aligned}
	\label{eq:robotdeposition2}
\end{equation}
where:
\begin{where}
    \item [\Delta_{ij}^k]: pheromone deposited by the robot $\{k\}$ in the cell $\{ij\}$
    \vspace{0.05cm}
    \item [\psi_{max}]: maximum concentration of pheromone in a cell
    \vspace{0.05cm}
    \item [\psi_{ij}^{t}]: concentration of pheromone in the cell $\{ij\}$ at time step $\{t\}$
    \vspace{0.05cm}
    \item [\alpha]: maximum possible amount of deposited pheromone
    \vspace{0.05cm}
    \item [\delta]: influence of the pheromone deposition rate
    \vspace{0.05cm}
    \item [\eta]: compensation of the environment evaporation rate
\end{where}

\subsubsection{State 4 - Pheromone Evaporation}
Although evaporation is a global external phenomenon, naturally ruled by the volatility of the substance in contact with the environment, in the PheroCom model this phenomenon becomes local and is simulated by each robot. The simulation of pheromone dynamics in a local map is one of the main contributions of this work, since it allows the implementation of a decentralised model and does not require physical means to mimic the substance. Since each robot has in its internal memory an individual pheromone map, all robots must evaporate the pheromone present in their maps at each time step (considering that a task that needs evaporation is being performed). The final calculation of the pheromone concentration present in each cell $x_{ij}$ is given by Equation~\ref{prop:ferofinal}.

\begin{equation}
    \psi^{t+1}_{ij} = \left[ \psi^{t}_{ij} - (\pmb{\beta} \cdot \psi^{t}_{ij})\right] + \sum_{k=1}^N{\Delta_{ij}^k}
    \label{prop:ferofinal}
\end{equation}

\noindent where, the pheromone concentration $\psi$ in a cell $x_{ij}$ at time step $(t + 1)$ is equal to the pheromone concentration present at time step $t$, minus the percentage $\beta$ of evaporated pheromone $\{\beta \in \mathbb{R}\ \|\ 0.0 < \beta \leq 1.0\}$, added to the deposit contribution made by the $N$ robots to this cell. Recalling that, a robot contributes to the pheromone in a cell $x_{ij}$ \textit{iff} this cell is within its deposition radius.

\subsubsection{State 6 - Dissemination (ViBIT)}
\label{sec:state6_dissemination}
Pheromone information propagates inside the swarm via dissemination (process of the ViBIT protocol (see Sec. \ref{sec:bio-inspiration})). Even though each robot has a local map containing information about its past deposits, aggregating pheromone data from other robots increases the efficiency of the decision-making process. This increase is due to the fact that the robots will not only be aware of their own pheromone deposits, but also of the deposits made by other robots. It allows a more accurate decision-making, taking into account that the robots will access information with global characteristics. Thus, each time a robot enters in State 6 (dissemination) of its FSM, a information dissemination of occurs, containing specific data about its local pheromone map.

Each robot has a transmission radius (defined in Section~\ref{individual_interface}) to disseminate information. This radius builds the area in which each robot's information disseminates and, as a consequence, the limit at which other robots can aggregate information. Thus, it is noted that there would be a difference between the maps of a decentralised and a globally centralised model. This conclusion is due to the fact that local maps do not have the same accuracy of information as a global map. In addition to the fact that the information disseminated does not contain the data of the complete map of each robot (since communication is a costly task), there is a limit in the communication area, which results in time-sparse information sharing, as it occurs only when the robots are in the neighbourhood of each other.

Information is disseminated without a peer-to-peer connection. Thus, the transmitting robot is unaware if any other robot has aggregated its data. This way, the transmitting robot only releases the information in a communication channel, and it is up to the other robots, that are in range, to decide if they will use it to improve their own local maps. Furthermore, the structure of the messages is described by Equation~\ref{eq:message}, where it is composed of the ``id'' of the transmitting robot, its current time step and the pheromone data (derived from its local pheromone map). The ``id'' and time step are used by the robots in order to control the information flow, which makes it possible to manage, for example, duplicate and echoed transmissions.

\begin{equation}
    \label{eq:message}
    message = [\ id\ +\ time\ step\ +\ (pheromone\ data)\ ]
\end{equation}

\subsubsection{State 7 - Aggregation (ViBIT)}
Considering that robots disseminate information in State 6, in State 7, they are able to aggregate information transmitted by other robots that are in their neighbourhood. Thus, the communication mechanism is not active continuously but at specific time intervals, for both data dissemination and aggregation. Since information about local pheromone maps is often disseminated, robots in State 7 aim to aggregate it into their own local maps, with the purpose of performing the task more efficiently. Moreover, considering that there is no peer-to-peer connection, the robots should manage which data to filter and which data to use to improve their local pheromone maps.

In order to control data aggregation, each robot keeps in its memory a list containing information about the robots in which it has received data in the past. This list consists of the ``id'' and the time step of the transmitting robot. Thereby, the aggregating robot can ensure that duplicate information is not reflected in its pheromone map, which can often happen as there are no direct connections and handshake processes. It can be formally defined as follows: \textit{Suppose that a robot A has aggregated information from a robot B, and robot B has stated that it was at the time step $t$. In a next transmission, robot A will aggregate new information from robot B, if and only if, robot B states that it is in a time step $(t + i)$, such that, $\{i \in \mathbbm{N}^{*}\}$.}

Accordingly, two conditions must be achieved to occur a transmission between two robots: (i) a robot has to be within the dissemination area of another robot, i.e., the distance between the two robots must be less than or equal to its transmission radius, and (ii) the data received must be new. The definition of ``new data'' for the robots is related to the absence of the ``id'' of the robot that is disseminating the data in the list of past transmissions or the disseminating robot's time step is greater than the time step recorded in the aggregating robot's transmissions history. In addition, the pheromone concentration of a cell that is being updated must be lower than the pheromone concentration received from the transmitting robot. Since the environment has an evaporation rate and this rate is simulated in the local pheromone maps of each robot, it can be stated that, comparing cells in the same positions from two different pheromone maps, the cell with the highest pheromone concentration holds the most up-to-date information.

Therefore, updating local pheromone maps enables the emergence of complex global behaviour\footnote{In surveillance, a complex global behaviour is related to the swarm's capacity for self-organisation so that all areas of the environment are constantly and cyclically monitored. See Section 2.3 for more information related to the surveillance task.}, as the movements and deposits of each robot may influence the behaviour of the whole swarm in the future. However, as stated before, this updating is not mandatory: upon receiving a data package, the robot will use evaluation metrics to decide which cells to update. This is due to the fact that there is a possibility that the aggregating robot itself already has more up-to-date data on its local map. In this case, the update of the parsed cells would be ignored.

\subsection{Individual robot interface}
\label{individual_interface}
\begin{figure}[h]
	\centering
	\includegraphics[width=3.5in]{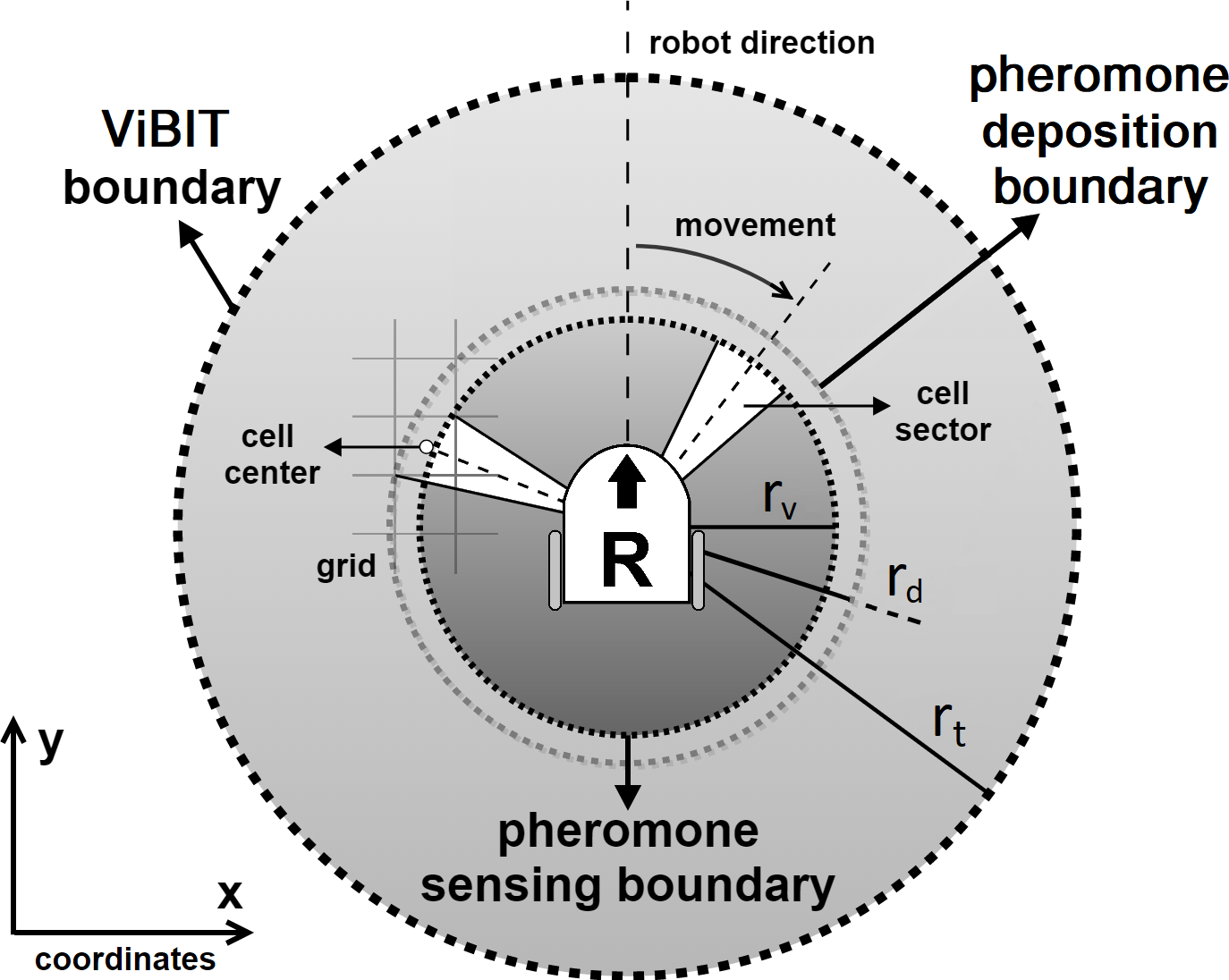}
	\caption{Visual representation of the robots' interface, used for interaction with the external environment and with other robots of the swarm. Three limits are highlighted: pheromone sensing, pheromone deposition and ViBIT boundaries.}
	\label{prop:robotinterface}
\end{figure}

The individual robot interface with the external environment is illustrated in Figure~\ref{prop:robotinterface}. In other words, the figure illustrates the structure that each robot has in order to interact with the surrounding environment, highlighting the transmission information system. In the centre of the figure is a representation of a robot and its current direction. Each robot has three external boundaries: pheromone sensing, pheromone deposition and the ViBIT boundaries.

The pheromone sensing boundary, defined by the vision radius ($r_v$), represents the sensing limit of the pheromone concentration around the robot (pheromone detection is performed in FSM State 1 - Fig.~\ref{fig:fsm}). While a robot is choosing the target cell of the next movement, the sensing boundary is divided into ``cell sectors'', where each sector is related to a cell of the pheromone grid. This cell should be at the edge of the area formed by $r_v$, and even if a cell is not completely within this area but is tangential to its perimeter, the cell will enter the set of candidates to be a possible destination of the robot. Thus, it is noted that the boundaries of sectors may overlap, but this does not interfere with the overall operation of the system. Once the target cell has been chosen, the robot, if necessary, performs a redirect movement and move as close as possible to the centre of this cell.

The pheromone deposition boundary, defined by the deposition radius ($r_d$), delimits the area that robots deposit pheromone (pheromone deposition is performed on FSM State 3 - Fig.~\ref{fig:fsm}). Recalling that pheromone deposits indicate in future time steps, for the robot itself and other robots of the swarm, that this area was monitored for an amount of time proportional to the detected pheromone concentration. The radius $r_d$ is set to be greater than or equal to the vision radius $r_v$. This is due to the fact that $r_v$ represents the area that was monitored by a robot, therefore, the deposition radius (defined by $r_d$) must at least allow signalling of an area equal to the monitored area (defined by~$r_v$). In general, in this work, $(r_d = r_v)$ was applied to redirect the experiments to more specific variables.

Finally, the ViBIT boundary, defined by the transmission radius ($r_t$), represents the limit of the transmission system's coverage area (used in FSM states 6-Dissemination and 7-Aggregation - Fig.~\ref{fig:fsm}). In this work, since the experiments were not performed with real robots, there was no need to simulate all characteristics of a real communication network, for example, the application of disturbances in the physical environment. However, it was assumed that it is possible to transmit information involving robots via wireless networks. According to~\citeonline{winfield2000distributed}, in order to apply a wireless network, three points must be previously assumed: (i) the transmission antenna present in each robot is uniformly omnidirectional in the horizontal plane; (ii) there must be a transmission threshold, which in this case is represented by the transmission radius $r_t$, i.e., if two or more robots are within the dissemination area of each other, data transmissions may occur, otherwise, there is no data propagation; and (iii) the wireless network must implement some type of time, channel or code difference so that the swarm can communicate on the same radio-frequency spectrum. All points raised are fulfilled in this work.

\subsection{Information propagation within the swarm}
\label{prop:complex_net}

\begin{figure}[h]
	\centering
	\includegraphics[width=3.7in]{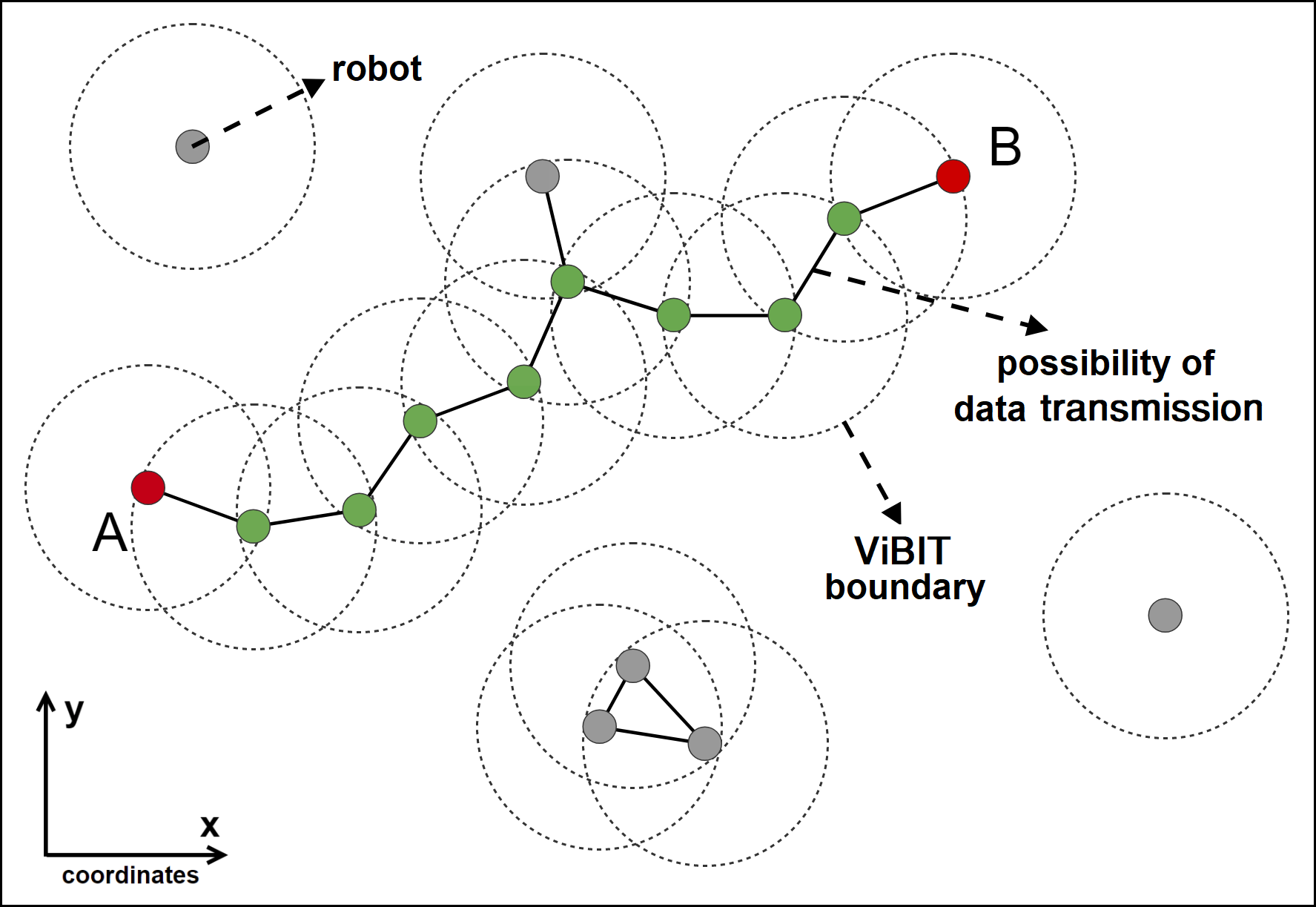}
	\caption{Formation of communication networks and information propagation. Vertices represent robots and edges, a possibility of data dissemination.}
	\label{prop:swarmrede}
\end{figure}

The robots' ability to indirectly communicate and transfer information within the swarm, results in a communication network, as exemplified in Figure~\ref{prop:swarmrede}. A network corresponds to an abstraction that allows systematising a relationship between pairs of objects. In computing, a network can be described by a graph~\cite{bondy1976graph}. In the figure, each vertex represents a robot, edges a possibility of information transmission, i.e., a relationship, and dotted circumferences the transmission limits of each robot (defined by the radius $r_t$ - Fig.~\ref{prop:robotinterface}). Since each robot has a limited transmission radius, network connections are created if a robot is within another robot's transmission area. In other words, there may be a connection between two robots \textit{iff} the distance $d$ among them is less than or equal to the transmission radius $(d \leq r_t)$. Thus, at each time step, there is a likelihood that there will be a complete network in the swarm (all robots can communicate with all other robots), a connected network (there is a path between all pairs of robots), disconnected networks forming groups of robots (connected components) or even no connections at all, if no robot is within the transmission area of any other robot. It is worthy to highlight that this is not a two-way communication, as there are no confirmations or responses in the ViBIT protocol.

Information spreads throughout the swarm without the need for direct interaction with the disseminating robot. This is due to the existence of multiple transmissions, even if they occur at different time steps. Consider robots `A' and `B' in Figure~\ref{prop:swarmrede} (red vertices). The pheromone information of robot `A' is propagated to robot `B' without the existence of a direct interaction. This information propagates along the path that connects the two robots (green vertices) during model evolution. As mentioned, even if this path is not present at a specific time step, but transmissions are happening with the evolution of the system, the information disseminated by `A' can reach `B' without a direct interaction. This is because robots aggregate pheromone data, that has been disseminated, in their local maps and then, in future time steps, disseminate it to other robots.

\subsection{Information dissemination by coverage area}
The states of the FSM (Fig.~\ref{fig:fsm}) that correspond to the robots' communication mechanism (State 6 - Dissemination and State 7 - Aggregation), make it possible to maintain local pheromone maps aggregating third party information (ViBIT protocol - Sec. \ref{sec:bio-inspiration}). Besides, disseminated data contain information about pheromone, which also represents a type of indirect communication (stigmergy). Thus, the two communication mechanisms combined characterise the PheroCom model.

\begin{figure}[h]
	\centering
	\includegraphics[width=4.2in]{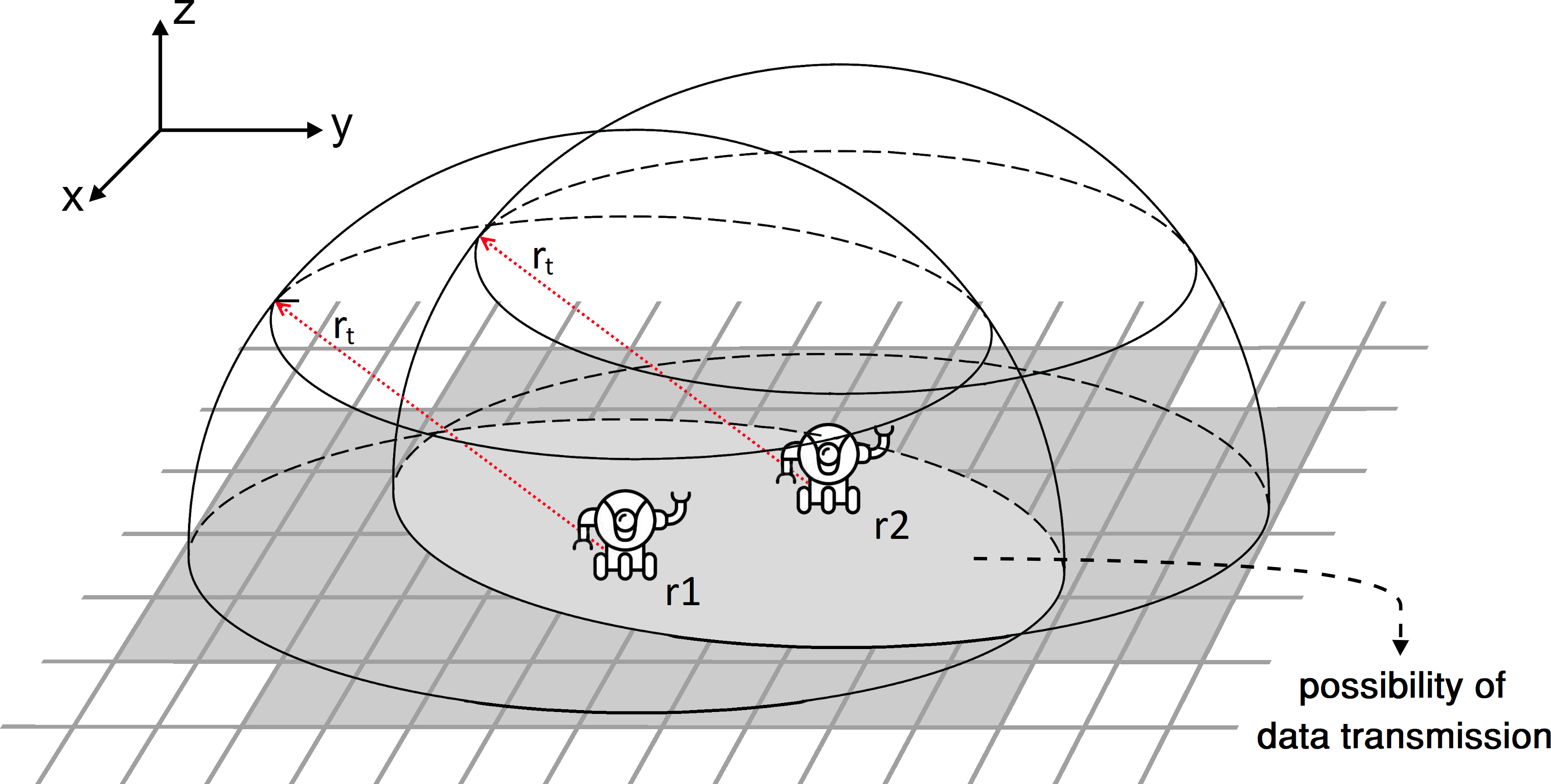}
	\caption{Information transmission scheme by coverage area. The figure illustrates two robots $r_1$ and $r_2$ and their respective transmission radius $r_t$ represented by the semi-spheres. Considering that they are within the dissemination area of each other, there is a possibility of data transmission.}
	\label{prop:area_01}
\end{figure}

Information transmitted by the robots corresponds to the pheromone data in their local maps. The PheroCom model, proposed in this work, presents a new way of pheromone representation and distribution for social insect-inspired algorithms applied in robotics. Information exchange takes place by coverage area, where only specific cells from the transmitter robot's local map are transmitted. Given the transmission radius $r_t$ of each robot, if another robot is within the area formed by this radius, there will be a possibility of these robots transmit information to each other. Cells selected for dissemination correspond to cells that are within the transmission radius $r_t$. Therefore, if a cell is within the area formed by $r_t$, even if it is not completely within this area but is tangent to its perimeter, the information of its pheromone concentration will be disseminated, as exemplified in Figure~\ref{prop:area_01}. The figure illustrates two robots ``$r_1$'' and ``$r_2$'' and their dissemination area, represented by the semi-spheres. Since the robots are within each other's dissemination area, data may be transmitted. If any data is transmitted, it will correspond to the pheromone concentration of all cells within their respective dissemination areas (in the figure, represented by the cells in grey).

Furthermore, this strategy applies the concept of communication networks presented in Section~\ref{prop:complex_net}, where even if two robots have no direct interaction, one can aggregate data from the other through the propagation of information. This propagation is only possible because the information transmitted by the robots represents exactly the pheromone concentration within their neighbourhoods. The dynamics of the pheromone make this phenomenon possible, because the concentration of the pheromone in a cell does not correspond only to the deposits made by a single robot. It corresponds to the aggregation of deposits made by all robots that passed through this cell in the past, excluding the evaporated portion. As previously mentioned, it is noteworthy that before aggregate third party pheromone information, the robots check which one is more up-to-date.

\section{Experiments and Analysis}
\label{sec:experiments-analysis}
The experiments presented in this section were executed applying the PheroCom model in the surveillance task. In this task, robots must spread throughout the environment, seeking to monitor areas that are not visited for a long time or that have never been visited (see Sec.~\ref{sec:surveillance_robotics}).

It is worth highlighting the main objectives of the experiments carried out in this section as: (i) to analyse the variations of the PheroCom model, investigating both its characteristics and its behaviour in different environments and with swarms of different sizes, as well as its limitations; and, more importantly, (ii) to verify whether the PheroCom model is a viable alternative and has similar performance to the centralised and synchronous model IACA-DI~\cite{tinoco2017improved}. Furthermore, experiments were executed applying the PheroCom model in a Multi-agent Simulation System (MaSS), implemented in C language by the authors, and in the Webots platform. 

\subsection{Experimental setup}
\begin{figure}[h]
	\centering
	\subfloat[Environment E1]{
		\label{environment1}
		\includegraphics[width=1.6in]{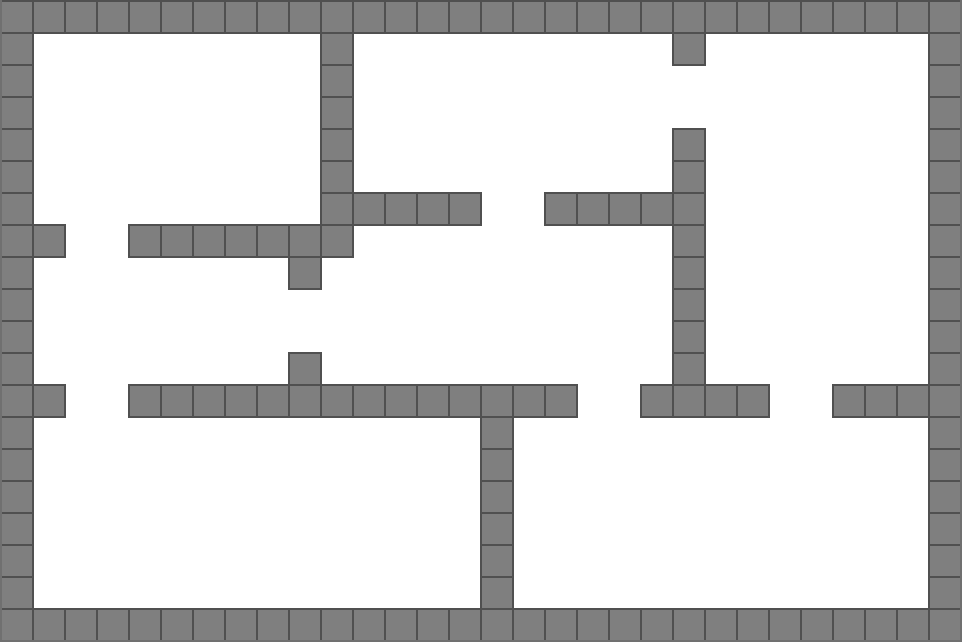}}
	\hfil
	\subfloat[Environment E2]{
		\label{environment2}
		\includegraphics[width=1.6in]{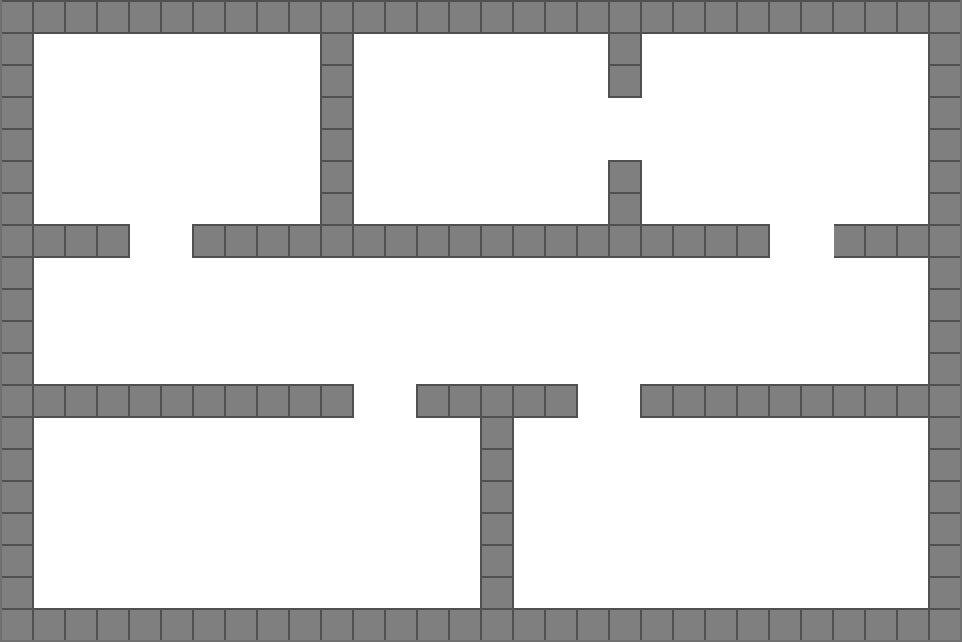}}
	\hfil
	\subfloat[Environment E3]{
		\label{environment3}
		\includegraphics[width=1.6in]{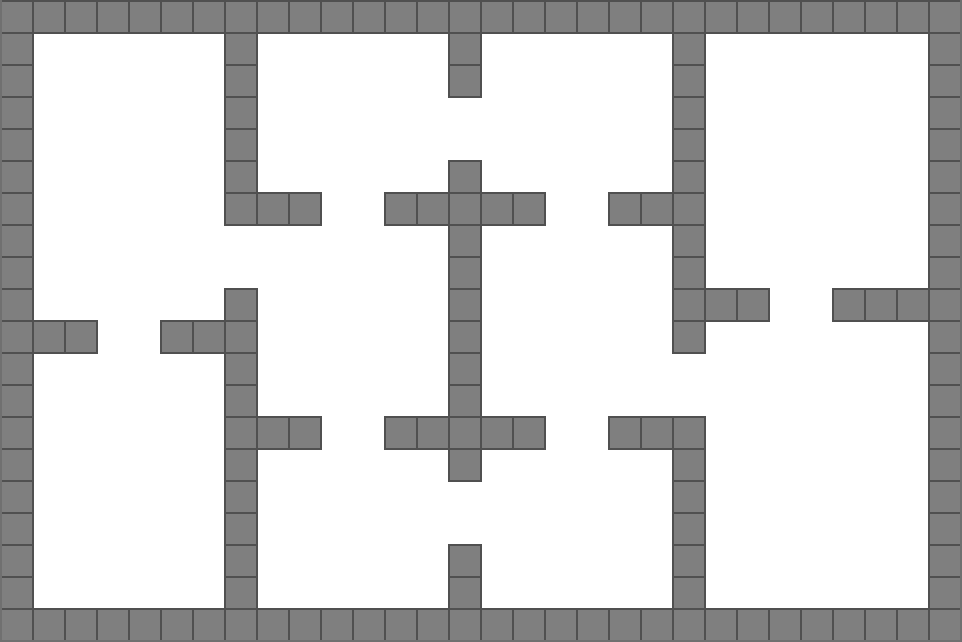}} \\
		
	\subfloat[Environment E4]{
		\label{environment4}
		\includegraphics[width=2.4in]{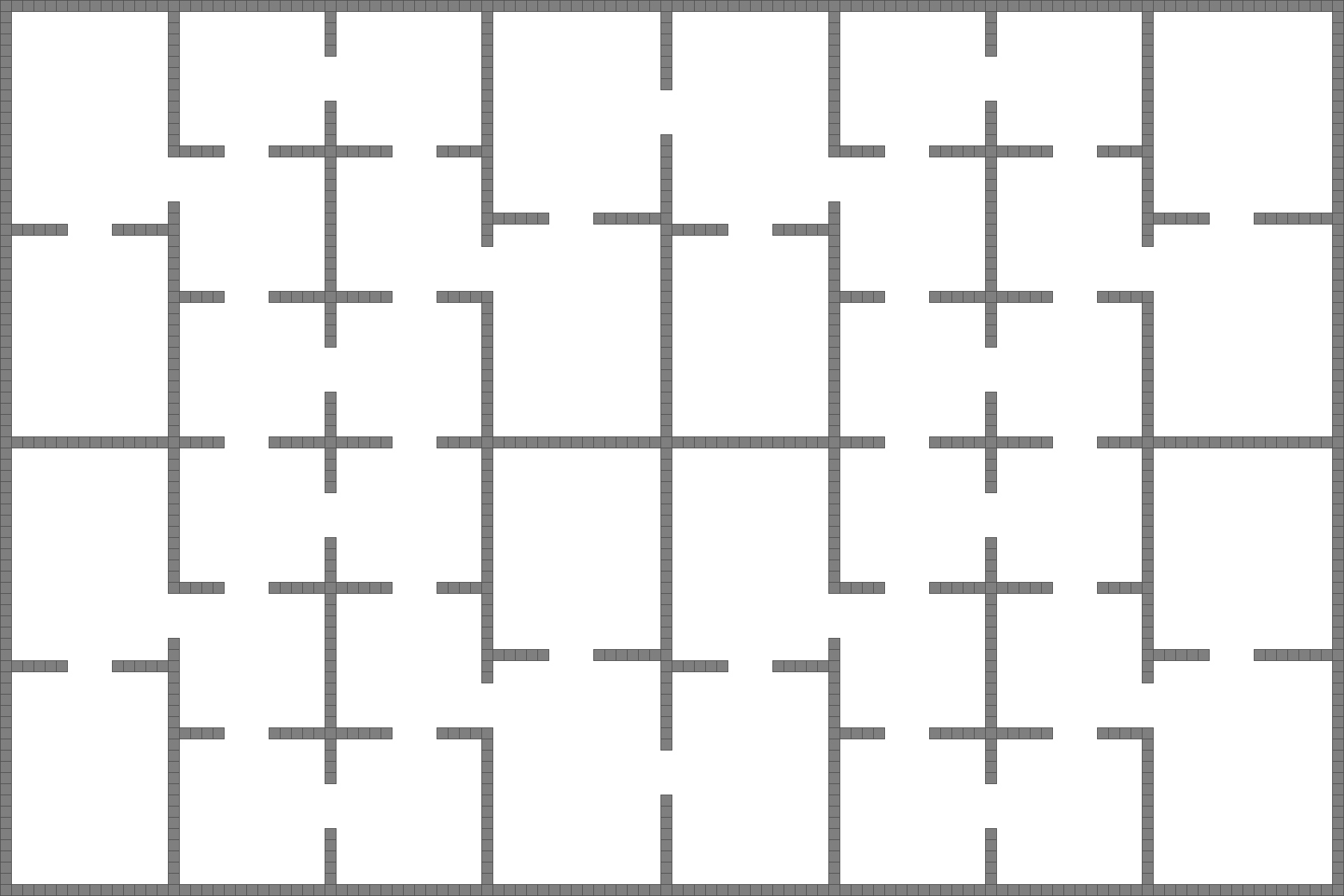}}
	\caption{Graphical representation of the environments: E1 - 7 rooms (Fig. \ref{environment1}); E2 - 6 rooms (Fig. \ref{environment2}); E3 - 10 rooms (Fig. \ref{environment3}); and, E4 - 40 rooms (Fig. \ref{environment4}).}
	\label{environments}
\end{figure}

The experiments\footnote{All experiments were carried out on an Intel Core i7-7700 3.6GHz CPU, 16GB RAM, NVIDIA GeForce GTX 1070 Ti, running Ubuntu 18.10 - Cosmic Cuttlefish.} were performed in seven different environments, as illustrated in Figure~\ref{environments}. These environments were divided into squared cells, in which free spaces are represented by white cells, i.e., cells in which robots can move, and obstacles (walls) are represented by grey cells. The environments have different sizes and shapes, where E1 (Fig. \ref{environment1}) has 7 rooms, E2 (Fig. \ref{environment2}) 6 rooms, E3 (Fig. \ref{environment3}) 10 rooms and E4 (Fig. \ref{environment4}) 40 rooms. Environments E1, E2 and E3 have dimensions equal to $(20 \times 30)$. Besides, experiments were also carried out in three other environments with the same room distributions, but with dimensions equal to $(40 \times 60)$ (environments E1', E2' and E3'), i.e., the scale of the environment was doubled in both axes. Figure \ref{environment4} shows the E4 environment with dimensions equal to $(80 \times 120)$ and 40 rooms. The main objective of using the E4 environment with these dimensions was to allow the execution of experiments with a larger number of robots, seeking to bring the team of robots closer to the dynamics of a swarm.

The number of time steps $T$ of each experiment is influenced by the dimensions of the environment: the larger is the size of the environment, the greater is the number of time steps required to observe the emergence of a global behaviour. In environments with dimension equal to $(20 \times 30)$ were used $10.000$ time steps per execution, with dimension equal to $(40 \times 60)$, $40.000$ time steps, and with dimension equal to $(80 \times 120)$, $120.000$ time steps.

In the experiments that have used the environments E1, E1', E2, E2', E3 and E3', the swarm was composed of 3 robots, whereas in the environment E4, 12 and 36 robots. This increase in the number of robots allows a scalability analysis of the model, heading towards the characteristics of a swarm. In environments E1, E2 and E3, the evaporation rate was maintained at $0.005\ (\beta = 0.5\%)$~\cite{tinoco2017improved}, while in the environments E1', E2', E3' and E4, was applied a rate of $0.001\ (\beta~=~0.1\%)$, defined empirically. Besides, based on our previous work~\cite{tinoco2019heterogeneous}, considering the movement strategies, a heterogeneous decision-making process has a higher performance. Thereby, proportionally, \nicefrac{2}{3} of the robots are composed by the inertial strategy and \nicefrac{1}{3} by the deterministic strategy. As a comparison measure, the results also represent experiments applying our previous model IACA-DI~\cite{tinoco2017improved}, with centralised and synchronous characteristics. Moreover, all data, excluding the heatmaps, represent the mean of $100$ samples using different seeds to avoid outliers.

It is noteworthy that the number of robots, evaporation rate and the time steps are environment-dependent variables. Accordingly, for each environment configuration, they have a different value in order to maintain the efficiency of the model. On the other hand, the variables $\{\mu,\ \nu,\ \alpha,\ \delta,\ \eta\}$, are independent. Thus, based on the analyses and optimisations carried out in our previous work~\cite{tinoco2020parameter}, the values of these variables were fixed in all experiments. Table~\ref{tab:experimental_setup01} summarises the setups of four different experiments (task-points, communication, heatmaps and cellsteps) performed in the MaSS.

Lastly, in the experiments carried out on the Webots platform~\cite{Webots04}, which is a free and open-source 3D robot simulator, it was used the e-puck\footnote{The e-puck is a robot developed for research and educational purposes. Technical specifications: STM32F4 168MHz, 2 stepper motors, 8 infrared sensors, 10 LEDs, 3D accelerometers (for more information: \url{http://www.e-puck.org/}, accessed on: 02/02/2022).} robot~\cite{mondada2009puck}. The experiments have consisted of adapting a marker on each robot, so they can stamp, on the floor, the path they have travelled during the simulation. The following parameters were adopted: a swarm with three robots ($N = 3$), an evaporation rate of $0.1 \%\ (\beta = 0.001)$, the environment E3' and ten hour simulation time. Each robot was equipped with a RGB marker with contrasting colours (red, green and blue). Different choice strategies were applied to enhance the analysis: three swarms with homogeneous strategies (random, deterministic and inertial) and one with heterogeneous strategies (\nicefrac{2}{3} of the robots with inertial strategy and \nicefrac{1}{3} with deterministic strategy). The random strategy was used as a lower limit of comparison for the other strategies and, since it does not use the pheromone information, it was applied only in the PheroCom model. In turn, using the other movement strategies, the proposed model PheroCom was compared with the centralised and synchronous model IACA-DI~\cite{tinoco2017improved}. 

\begin{table}[]
\centering
\resizebox{\textwidth}{!}{%
\begin{tabular}{@{}ccccccccccc@{}}
\toprule
\multicolumn{1}{c|}{\textbf{experiment}} &
  \textbf{environments} &
  \textbf{rb} &
  \textbf{time steps} &
  \textbf{$r_c$} &
  \textbf{$\beta$} &
  \textbf{$\mu$} &
  \textbf{$\nu$} &
  \textbf{$\alpha$} &
  \textbf{$\delta$} &
  \textbf{$\eta$} \\ \midrule
\multicolumn{1}{c|}{\multirow{4}{*}{task-points}} &
  E1, E2, E3 &
  03 &
  10,000 &
  \multirow{4}{*}{0$\sim$20} &
  0.005 &
  0.3 &
  0.3 &
  0.5 &
  0.1 &
  2.0 \\
\multicolumn{1}{c|}{}                           & E1’, E2’, E3’ & 03 & 40,000  &    & 0.001 & 0.3 & 0.3 & 0.5 & 0.1 & 2.0 \\
\multicolumn{1}{c|}{}                           & E4            & 12 & 120,000 &    & 0.001 & 0.3 & 0.3 & 0.5 & 0.1 & 2.0 \\
\multicolumn{1}{c|}{}                           & E4            & 36 & 120,000 &    & 0.001 & 0.3 & 0.3 & 0.5 & 0.1 & 2.0 \\ \cmidrule(r){1-1}
\multicolumn{1}{c|}{communication}              & E4            & 12 & 120,000 & 13 & 0.001 & 0.3 & 0.3 & 0.5 & 0.1 & 2.0 \\ \cmidrule(r){1-1}
\multicolumn{1}{c|}{heatmaps}                   & E4            & 12 & 120,000 & 13 & 0.001 & 0.3 & 0.3 & 0.5 & 0.1 & 2.0 \\ \cmidrule(r){1-1}
\multicolumn{1}{c|}{\multirow{2}{*}{cellsteps}} & E1, E2, E3    & 03 & 10,000  & 6  & 0.005 & 0.3 & 0.3 & 0.5 & 0.1 & 2.0 \\
\multicolumn{1}{c|}{}                           & E4            & 12 & 120,000 & 13 & 0.001 & 0.3 & 0.3 & 0.5 & 0.1 & 2.0 \\ \bottomrule
\end{tabular}%
}
\vspace{0.15cm}
\caption{Experimental setups of each task evaluated in the MaSS.}
\label{tab:experimental_setup01}
\end{table}

\subsection{Multi-agent Simulation System (MaSS)}
\label{exp:sec_simple_platform}
The MaSS aims to carry out mass experiments\footnote{The essence of the source code and the fully set of experimental data can be found public online: \url{https://github.com/claudineyrt/PheroCom-ViBIT} (Accessed on: 02/02/2022).}. This is due to the fact that this system does not take into account the physical characteristics of the robots and the environment, i.e., the objective is only to simulate the behaviour of the swarm against the proposed model. Besides, given the great diversification in the parameters and number of time steps of the experiments performed in this simulation system, it was possible to refine the model, so that it could be implemented in a 3D virtual robot simulation platform, where new factors (e.g., physics and hardware) are also taken into account.

\subsubsection{Task-points performance experiment}
\label{exp:task_points}
Task-points charts show how many times the proposed task has been completed, given a maximum amount of time steps. In surveillance of indoor environments, a task-point is achieved when all rooms have been visited by at least one robot. On one hand, effectiveness of this task is associated to having rooms to be cyclically visited, i.e., to reach task-points. On the other hand, efficiency consists in decreasing the time between two consecutive task-points, i.e., higher frequency implies that information about rooms' state is generally not outdated. While efficiency requires effectiveness, the latter does not imply the former. A task-point can be formally defined as follows: \textit{``Let $E$ be an environment composed of `$m$' rooms and $S$ a swarm composed of `$n$' robots. A room $i$ belonging to the environment $E$ is described as $\{r_i\ /\ (i \leq m)\ and\ (i \in \mathbbm{N}^*)\}$. Similarly, a robot $i$ belonging to the swarm $S$ is described as $\{s_i\ /\ (i \leq n)\ and\ (i \in \mathbbm{N}^*)\}$. Thus, a task-point is reached \textit{iff} every room $r_i \in E$ receives a visit from at least one robot $s_i \in S$.''} It is noteworthy that, when a task-point is reached, the count of visited rooms is restarted to start the counting of a new task-point. Finally, in the time step subsequent to the reset of the counting, all rooms that have the presence of robots will be considered a visited room in the current task-point.

\begin{figure}[b]
    \centering
	\subfloat[Environment E1 (3 robots).]{
		\includegraphics[width=2.7in]{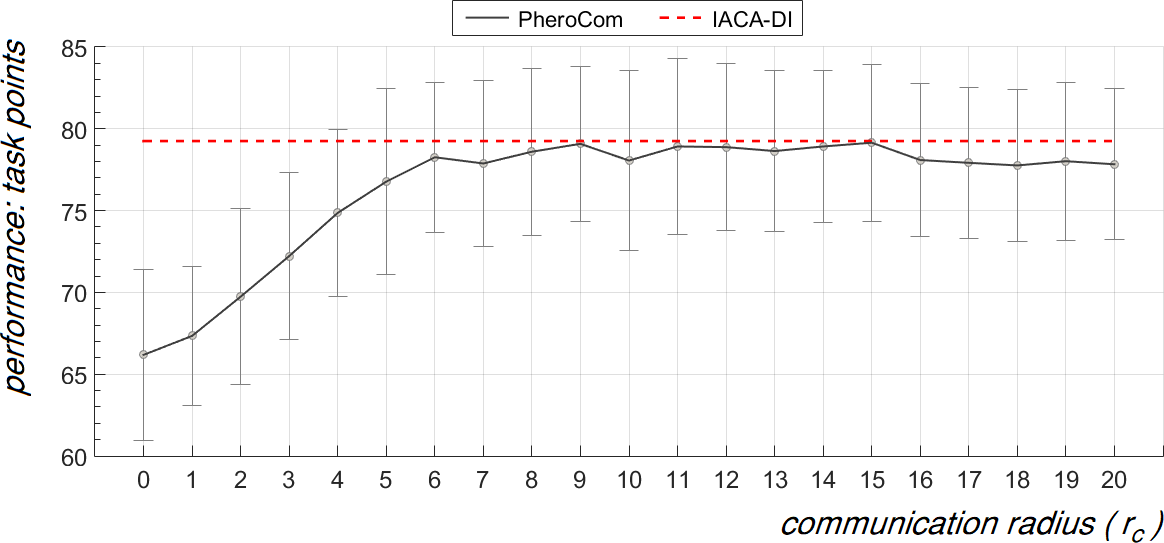}
		\label{pherocom_pt1}}
	\hfil
	\subfloat[Environment E1' (3 robots).]{
		\includegraphics[width=2.7in]{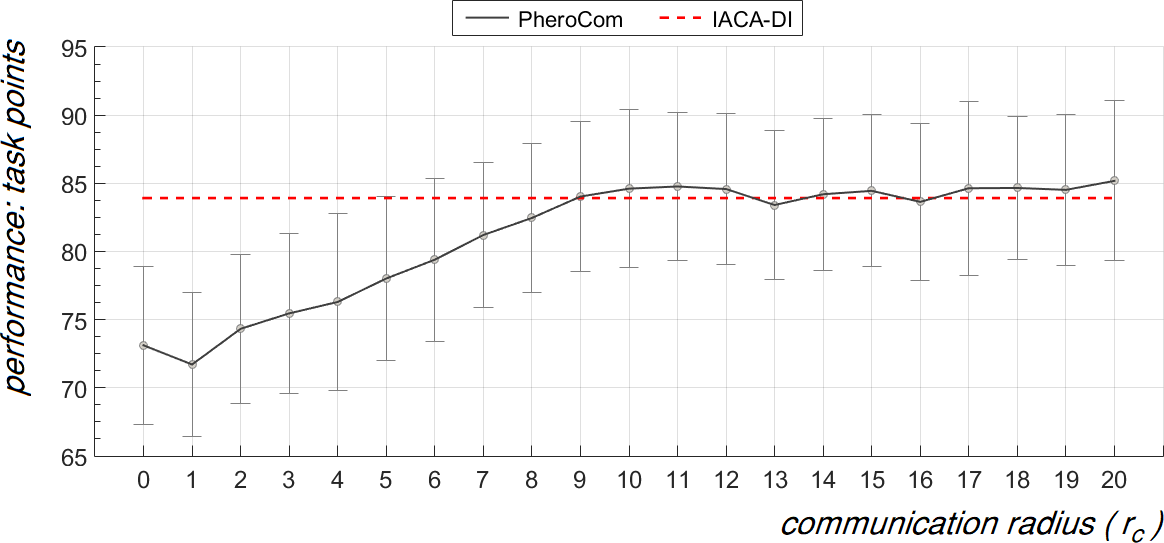}%
		\label{pherocom_pt2}}
	
	\subfloat[Environment E2 (3 robots).]{
		\includegraphics[width=2.7in]{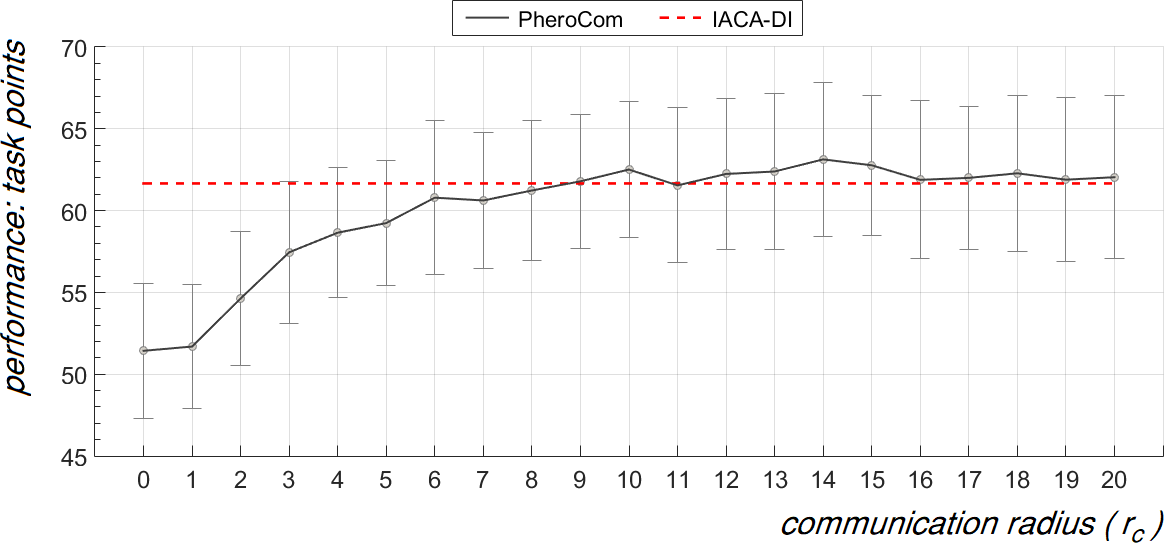}%
		\label{pherocom_pt3}}
	\hfil
	\subfloat[Environment E2' (3 robots).]{
		\includegraphics[width=2.7in]{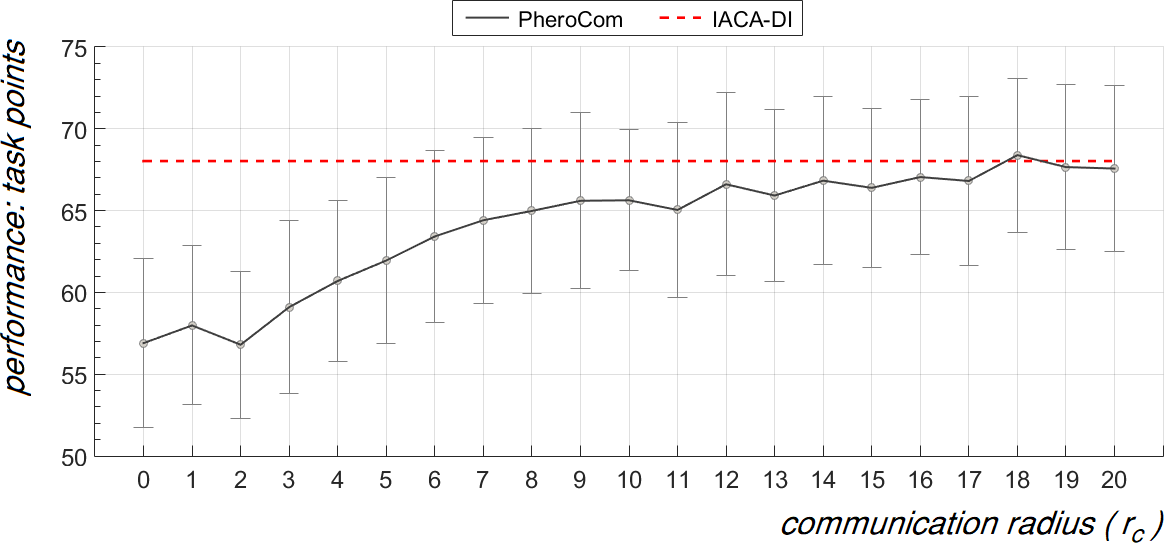}%
		\label{pherocom_pt4}}
    	
	\subfloat[Environment E3 (3 robots).]{
		\includegraphics[width=2.7in]{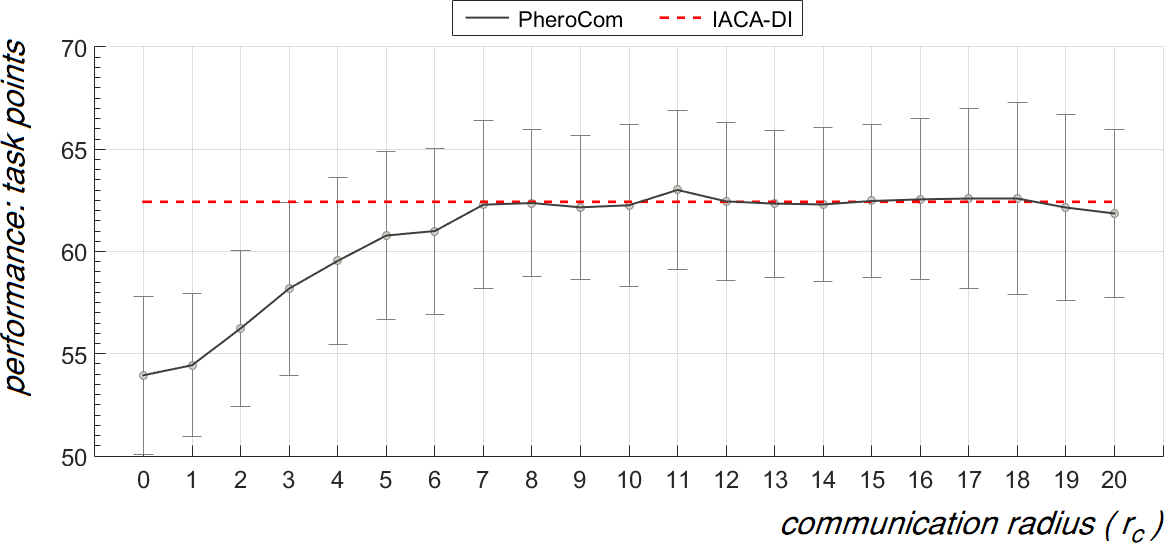}%
		\label{pherocom_pt5}}
	\hfil
	\subfloat[Environment E3' (3 robots).]{
		\includegraphics[width=2.7in]{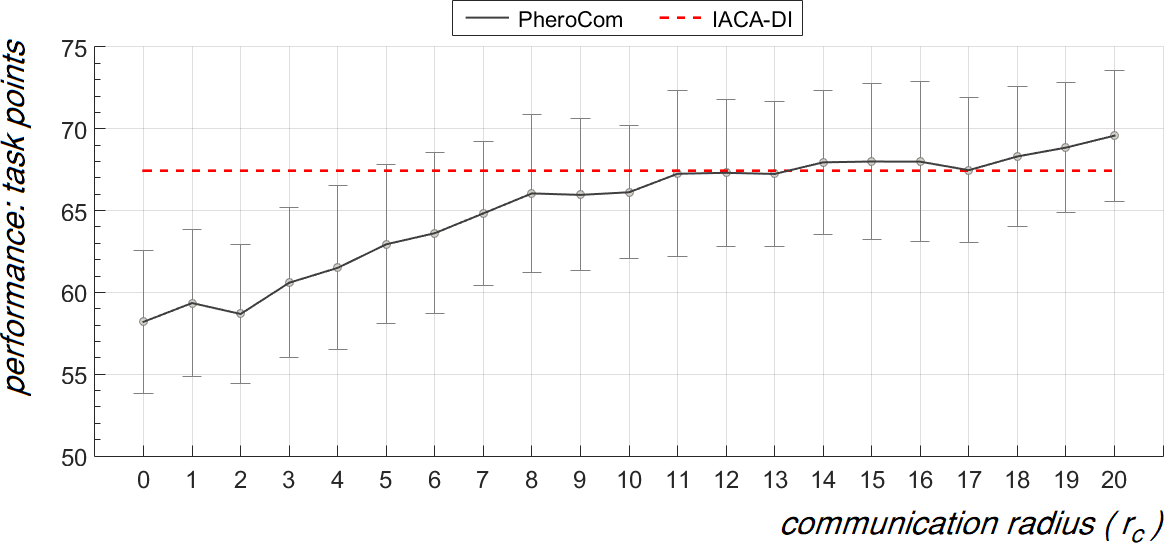}%
		\label{pherocom_pt6}}
	
	\subfloat[Environment E4 (12 robots).]{
		\includegraphics[width=2.7in]{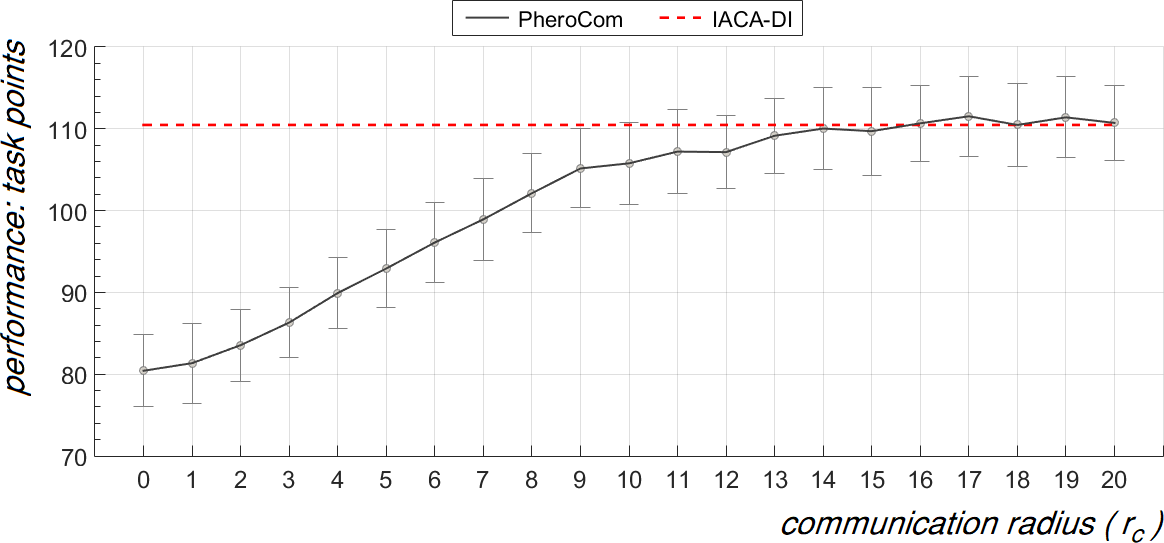}
		\label{pherocom_pt7}}
	\hfil
	\subfloat[Environment E4 (36 robots).]{
		\includegraphics[width=2.7in]{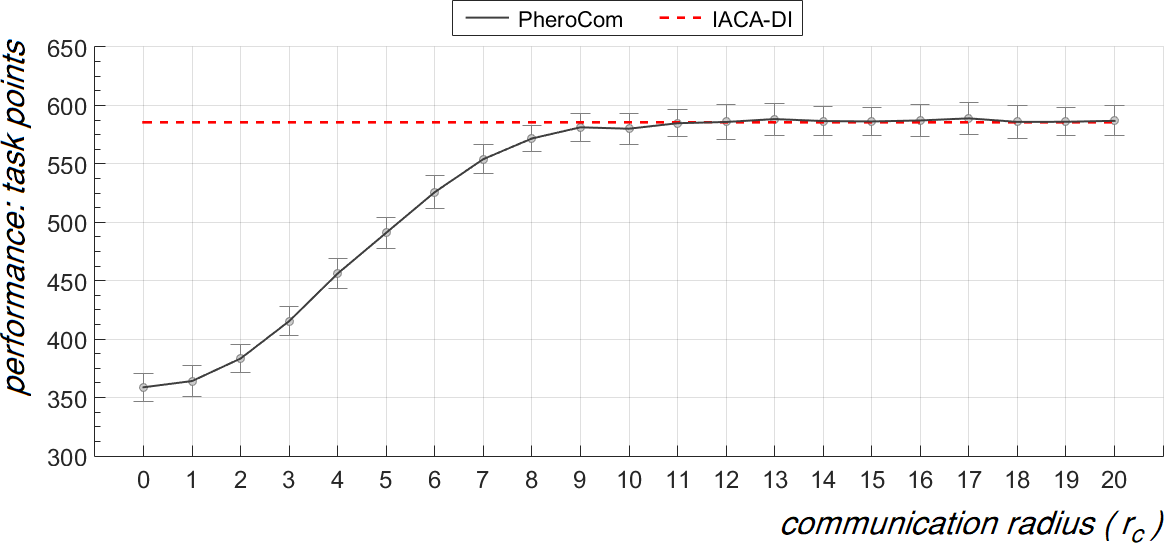}%
		\label{pherocom_pt8}}
		
	\caption{PheroCom model performance with task-points. The charts show the results of the variation of the model's communication radius, in different environments and with different swarm sizes. The line in red represents the performance of our previous model IACA-DI \cite{tinoco2017improved}.}
	\label{pherocom_pttask1}
\end{figure}

Figure~\ref{pherocom_pttask1} presents the results of task-points experiments for the seven described environments. In addition, it was applied a variation in the size of the transmission radius $r_t$, which consequently changes the size of the data disseminated. The vertical axis describes the number of task-points reached. In turn, the horizontal axis represents the variations in the size of the transmission radius $r_t$, which in this case ranged from $r_t = 0$ (no information is exchanged between the robots) to $r_t = 20$ (the information exchanged is made up of all cells that are less than or equal to radius 20). The red dotted line indicates the ground truth, in which the centralised model IACA-DI~\cite{tinoco2017improved} is applied. This centralisation can be represented by a specialised agent, internal or external (e.g., a robot of the swarm itself, an external system or an intelligent environment), which performs the centralisation and control of pheromone information. Given this, it is possible to notice that changes in the transmission radius affects the swarm performance. Therefore, the main objective of this experiment is to analyse the evolution of the swarm efficiency with the gradual increase of the amount of information exchanged by the robots, trying to verify if this efficiency approaches the ground truth.

Analysing the set of environments with dimensions $(20 \times 30)$ (Figs.~\ref{pherocom_pt1}, \ref{pherocom_pt3} and \ref{pherocom_pt5}), it can be seen that the PheroCom model has achieved a performance equivalent to ground truth averaging around a transmission radius equal to six $(r_t = 6)$. A radius equal to six represents a coverage of approximately $28.16\%$ of the size of an environment $(20 \times 30)$, i.e., within this radius, the transmitted message can contain up to $28.16\%$ of the total cells that compose this environment (given that the dissemination area may exceed the boundaries of the environment and the number of cells covered may be smaller). Employing the set of environments with dimensions $(40 \times 60)$ (Figs.~\ref{pherocom_pt2}, \ref{pherocom_pt4} and \ref{pherocom_pt6}), the ground truth was hit approximately at radius nine ($r_t = 9 \rightarrow 15.04\%$).

In order to analyse the scalability of the PheroCom model, experiments were also carried out with bigger swarms in a larger environment. These experiments are illustrated in Figures~\ref{pherocom_pt7} and \ref{pherocom_pt8}, in which the environment E4 and a swarm with $12$ and $36$ robots were applied, respectively. The goal of the PheroCom model, i.e., the ground truth, is $\approx111$ task-points in the first chart (Fig.~\ref{pherocom_pt7}) and $\approx585$ in the second (Fig.~\ref{pherocom_pt8}). In the experiments of the Figure~\ref{pherocom_pt7}, the PheroCom model has achieved its goal with a transmission radius equal to thirteen $(r_t = 13  \rightarrow 7.59\%$ of the size of the environment$)$. In turn, as expected due to the size of the swarm, in the experiments in Figure~\ref{pherocom_pt8}, it was achieved with a radius equal to nine $(r_t = 9 \rightarrow 3.76\%)$, four units smaller than the previous one. Although these experiments were carried out in the same environment, the number of robots is different. The greater the number of robots, the less is the chance of the system's outcomes fluctuate. More than that, the smaller is the transmission radius needed to achieve a satisfactory result. This explains the faster convergence and stability on the curve in the chart of Figure~\ref{pherocom_pt8}, in which the swarm was composed of 36 robots.

These outcomes have allowed to conclude that, in all analysed configurations, the PheroCom model was able to achieve a performance similar to that of the IACA-DI model, considering the task-points metric. It is noteworthy that this similarity in performance occurs despite the fact that the PheroCom model does not have access to the same quality of information as the IACA-DI model: the former applies a decentralised and asynchronous technique, sharing only regions of interest on the pheromone map, while the latter shares the entire pheromone map in a centralised and synchronous way.

\subsubsection{Data transmission experiment}
In the PheroCom model, there is no synchronisation of the information exchanged between the robots, no centralising agent and, moreover, it does not need to transmit the entire pheromone map to achieve a performance similar to that achieved by the IACA-DI model (as can be seen in the experiments shown in Figure~\ref{pherocom_pttask1}). Therefore, when the PheroCom model is applied, it is expected a lower number of transmissions and data exchanges, which is an extremely desirable feature in robotics, since data exchange is very costly.

\begin{figure}[ht]
	\centering
	\includegraphics[width=3.3in]{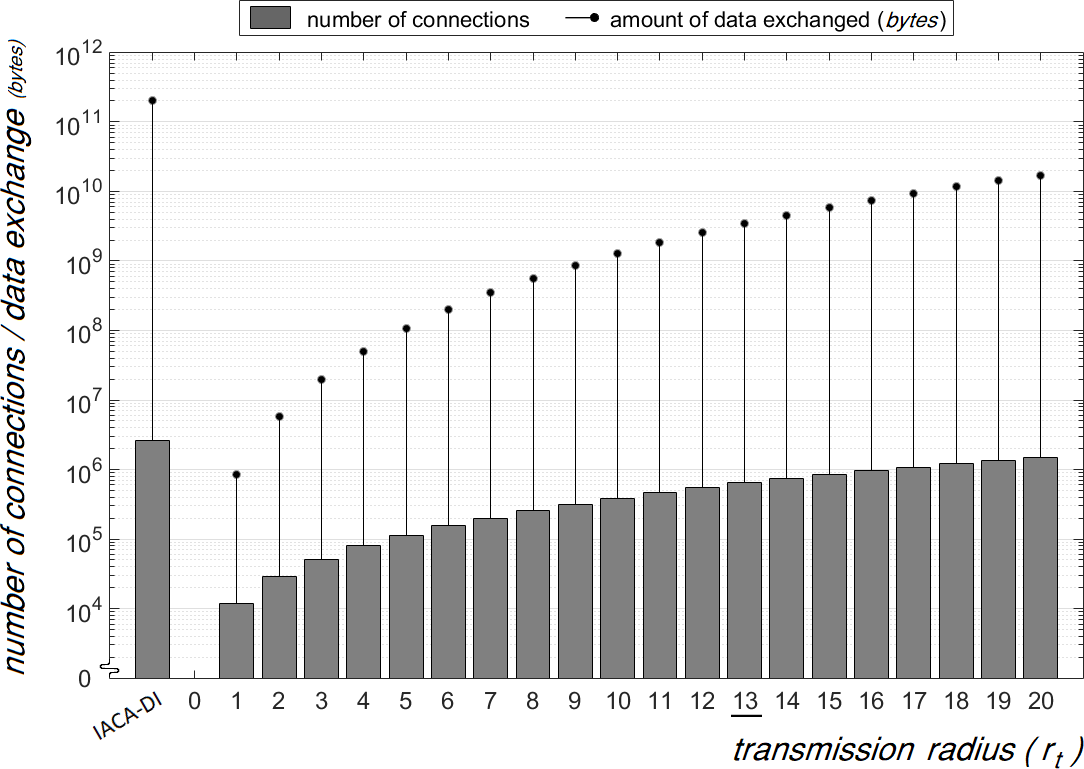}%
	\caption{Measurement of intra-swarm data dissemination, given different transmission radius. The figure shows the average number of transmissions (bars chart) and amount of data exchanged (stems chart).}
	\label{fig:exp_comm}
\end{figure}

Figure~\ref{fig:exp_comm} shows a comparative chart containing information on the mean number of intra-swarm transmissions and data exchange. Looking at the bar chart (mean of transmissions), it is possible to notice that the centralised model made $2.64 \times 10^{6}$ transmissions in each simulation. In turn, the PheroCom model, with $(r_t = 13)$, has made around $\approx{6.42} \times 10^{5}$ transmissions. These values show that the number of transmissions made by the PheroCom model is $\approx{4.11}$ times smaller to that made by the IACA-DI model. Considering the stem chart (amount of data exchanged - unit of measure in \textit{bytes}), it is also possible to verify an extensive difference between the PheroCom and the IACA-DI model. In the former, where the entire map is transferred due to centralisation and synchronisation, the amount of data exchanged was $\approx{2.03}\times{10^{11}}\ bytes$ throughout the simulation. The latter, this value decreases to $\approx{3.43}\times{10^{9}}\ bytes \rightarrow\ \approx{59}$ times smaller.

\begin{figure}[h]
	\centering
	\subfloat[Dissemination rate $(r_t = 13)$]{
		\includegraphics[width=2.5in]{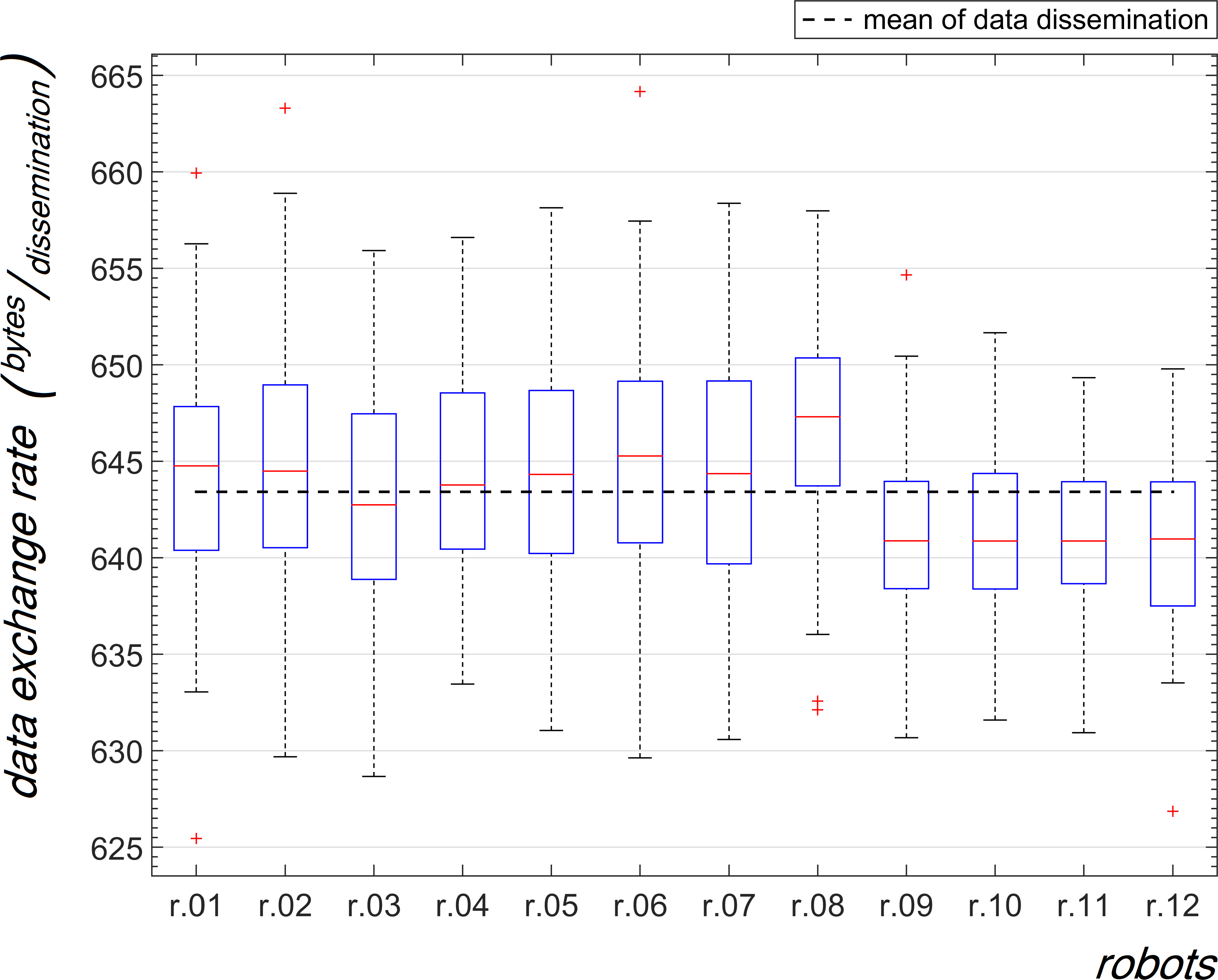}%
		\label{fig:bandwidth_box1}}
	\hfil
	\subfloat[Aggregation rate $(r_t = 13)$]{
		\includegraphics[width=2.5in]{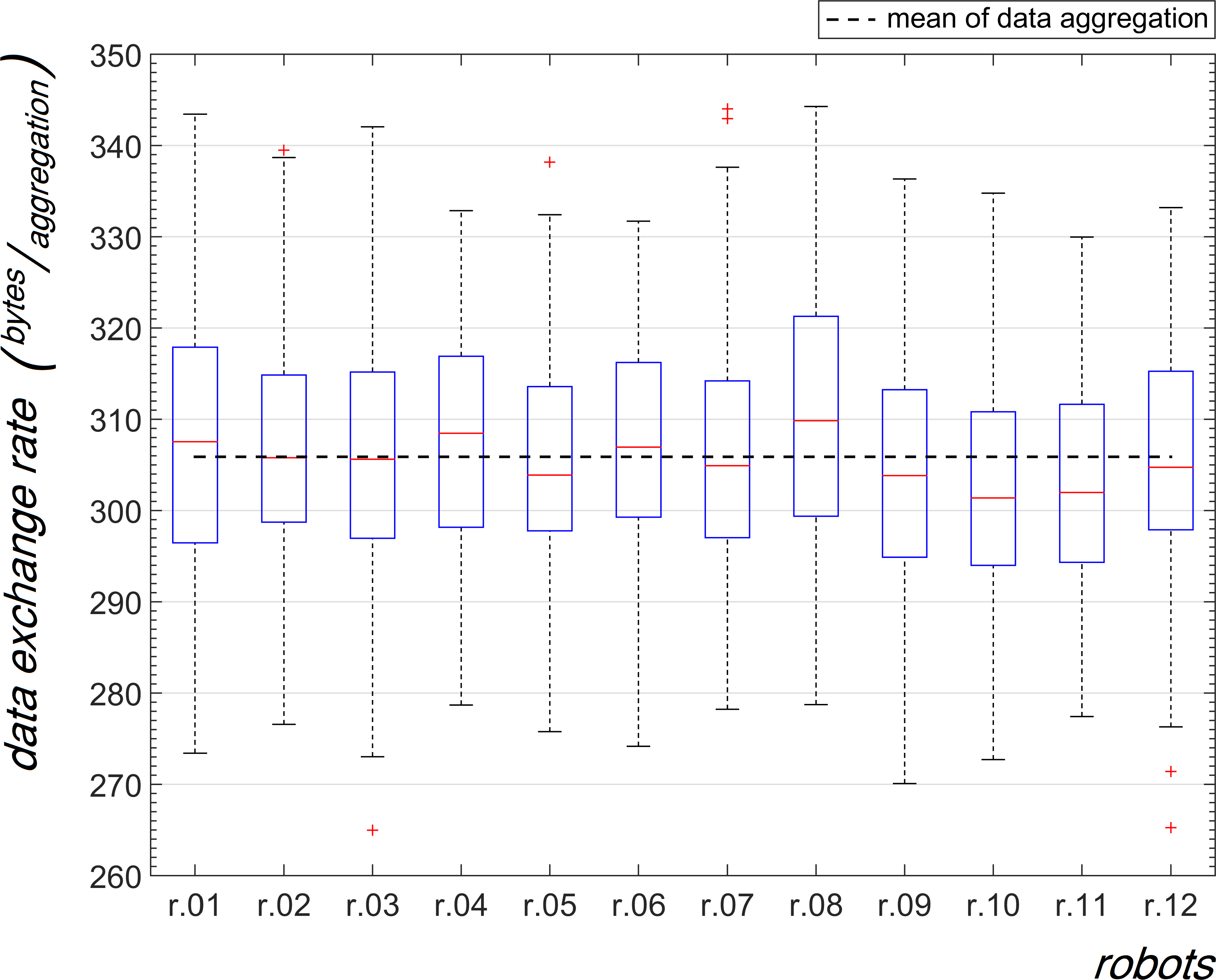}%
		\label{fig:bandwidth_box2}}
	\caption{Measurement of the bandwidth required for implementation on real robots. Figure~\ref{fig:bandwidth_box1} shows the discretised and the mean rate of information dissemination, in turn, Figure~\ref{fig:bandwidth_box2} presents the rates of aggregation.}
	\label{fig:bandwidth}
\end{figure}

In a simplified way, considering the number of transmissions carried out successfully, the PheroCom model has performed $\approx{24.31\%}$ of the number of transmissions performed by the IACA-DI model. Regarding the amount of data transmitted, this value is even more expressive, since the PheroCom model has transmitted only $\approx{1.69\%}$ of the amount of data transmitted by the IACA-DI model. Bearing in mind that, even with this low amount of data exchanged, the two models have achieved similar outcomes in the evaluated task.

In order establish whether the bandwidth required would dictate the communication medium (e.g., infra-red, Bluetooth, Wi-Fi) required for implementation on real robots, a quantitative study was carried out to define how much pheromone data needs to be transmitted by each robot. Figure~\ref{fig:bandwidth} presents data related to this bandwidth analysis. Figure~\ref{fig:bandwidth_box1} shows the discretised and the mean rate of information dissemination, in turn, Figure~\ref{fig:bandwidth_box2} presents the rates of aggregation based on bytes. The x-axis (horizontal) describes the 12 robots that form the swarm, while on the y-axis (vertical) is the data exchange rate. Each box-plot contains data from 100 model runs. The dotted black line represents the overall mean, whether in dissemination and aggregation. According to the data presented, it is possible to observe that, in the evaluated scenario, each robot uses an mean of 644 bytes per dissemination and 306 bytes per aggregation. These values already include the message version controls (see Section~\ref{sec:state6_dissemination} for more details).

\subsubsection{Pheromone heatmaps analysis}
Pheromone heatmaps represent the concentration of pheromone throughout the environment at a given time step, i.e., a screenshot that shows the amount of pheromone spread in each analysed area. The concentration of pheromone is represented by different colour temperatures: areas with warmer colours (tending to red) have higher concentration, on the other side, areas with colder colours (tending to blue) have lower or no concentration.

\begin{figure}[!ht]
	\centering
	\subfloat[PheroCom - robot.01]{
		\includegraphics[width=1.8in]{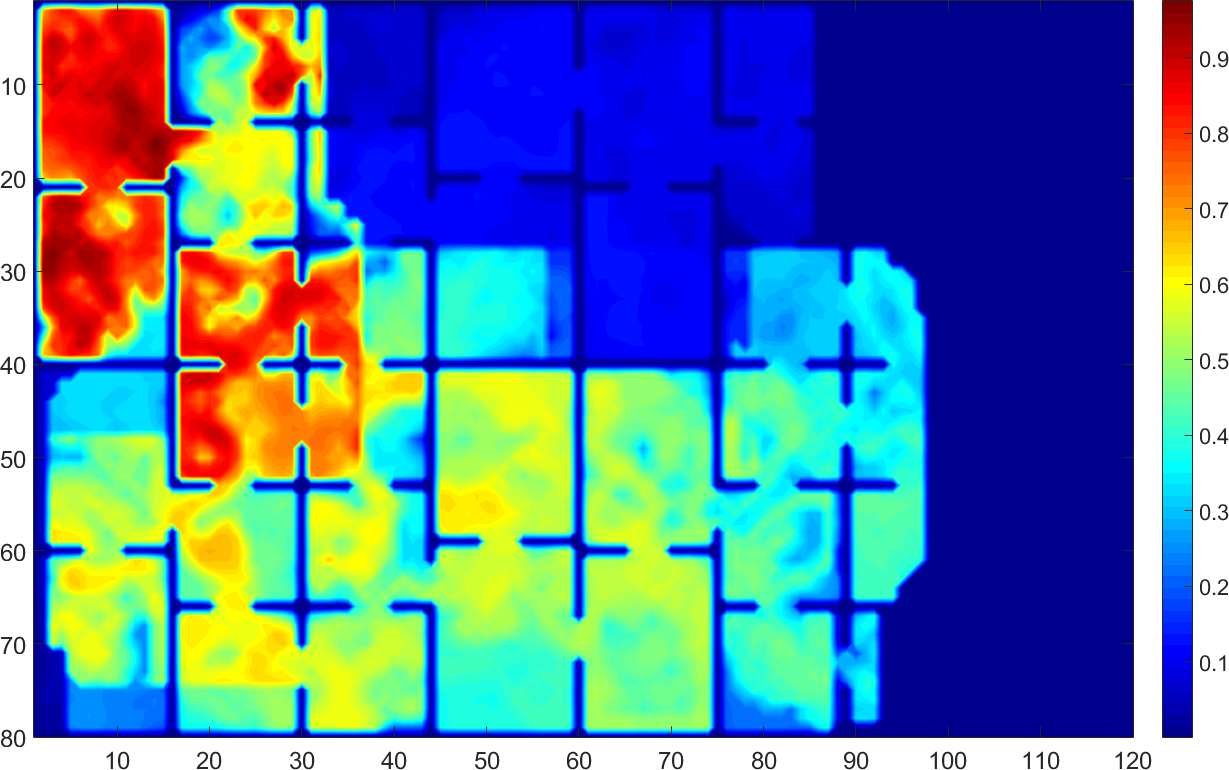}%
		\label{fig:exp_heatmap1}}
	\hfil
	\subfloat[PheroCom - robot.02]{
		\includegraphics[width=1.8in]{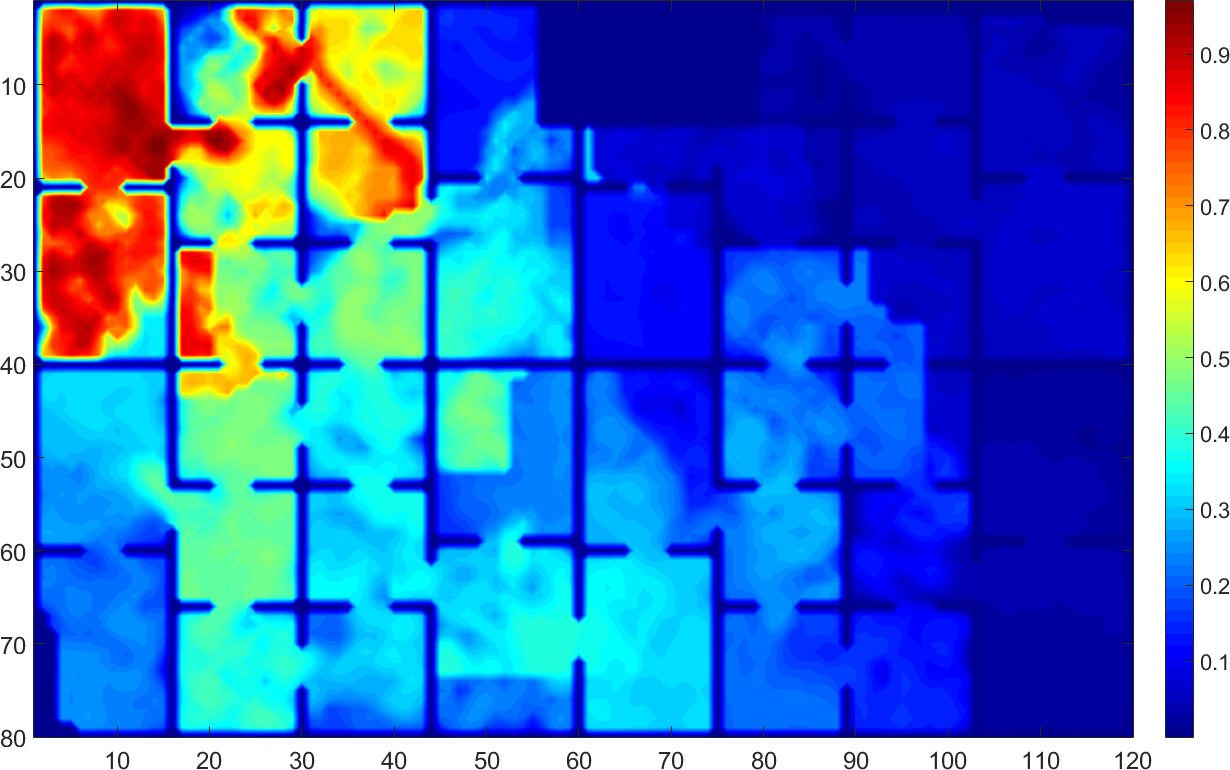}%
		\label{fig:exp_heatmap2}}
	\hfil
	\subfloat[PheroCom - robot.03]{
		\includegraphics[width=1.8in]{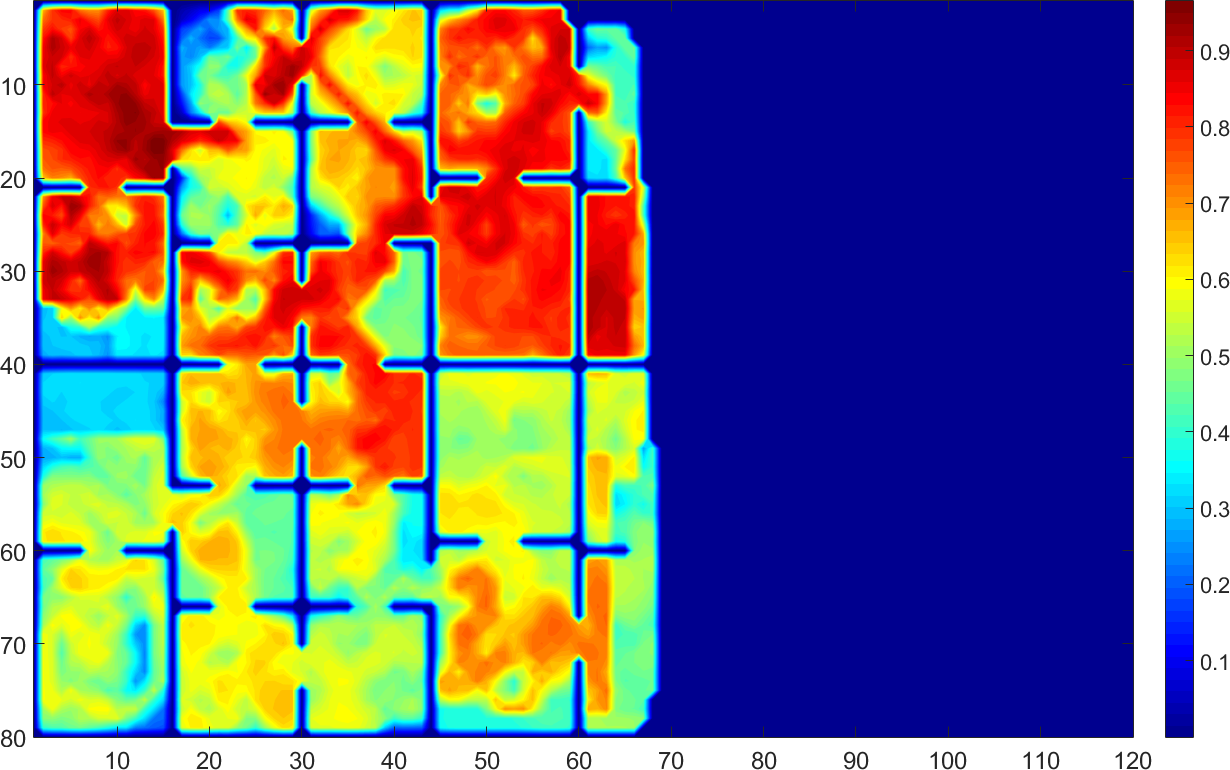}%
		\label{fig:exp_heatmap3}}
	
	\subfloat[PheroCom - robot.04]{
		\includegraphics[width=1.8in]{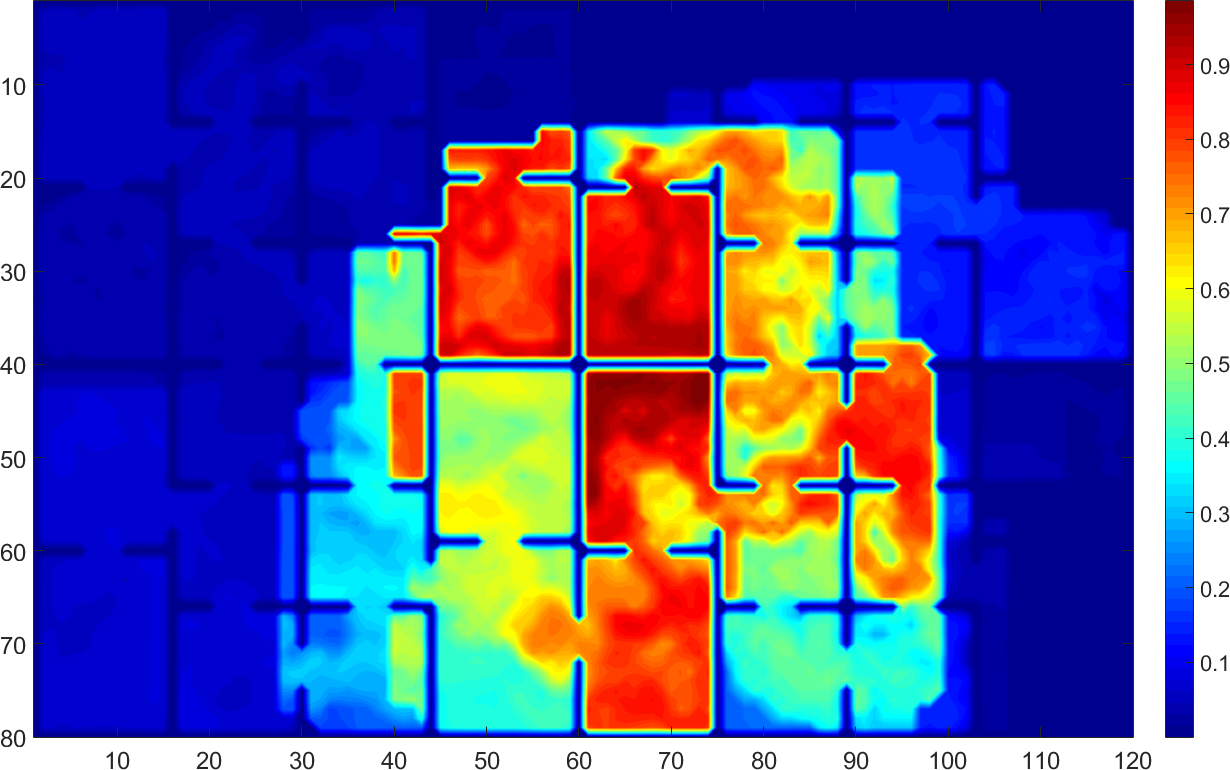}%
		\label{fig:exp_heatmap4}}
	\hfil
	\subfloat[PheroCom - robot.05]{
		\includegraphics[width=1.8in]{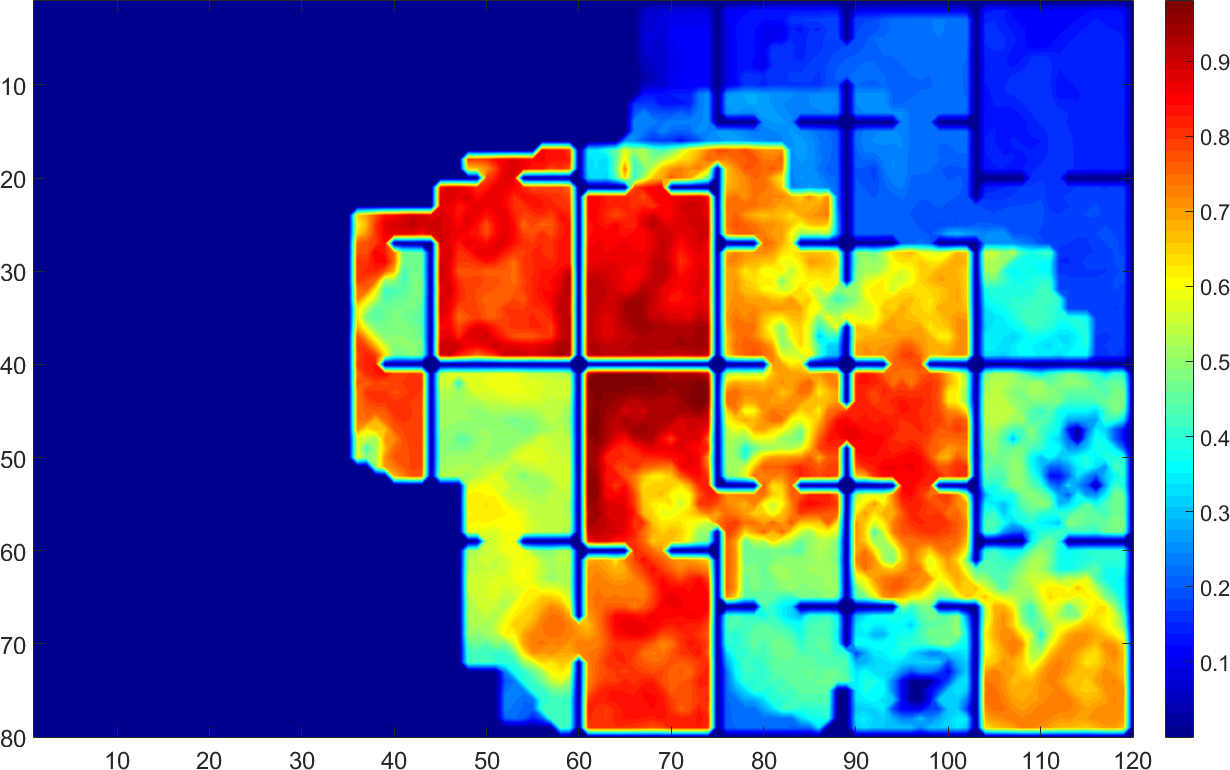}%
		\label{fig:exp_heatmap5}}
	\hfil
	\subfloat[PheroCom - robot.06]{
		\includegraphics[width=1.8in]{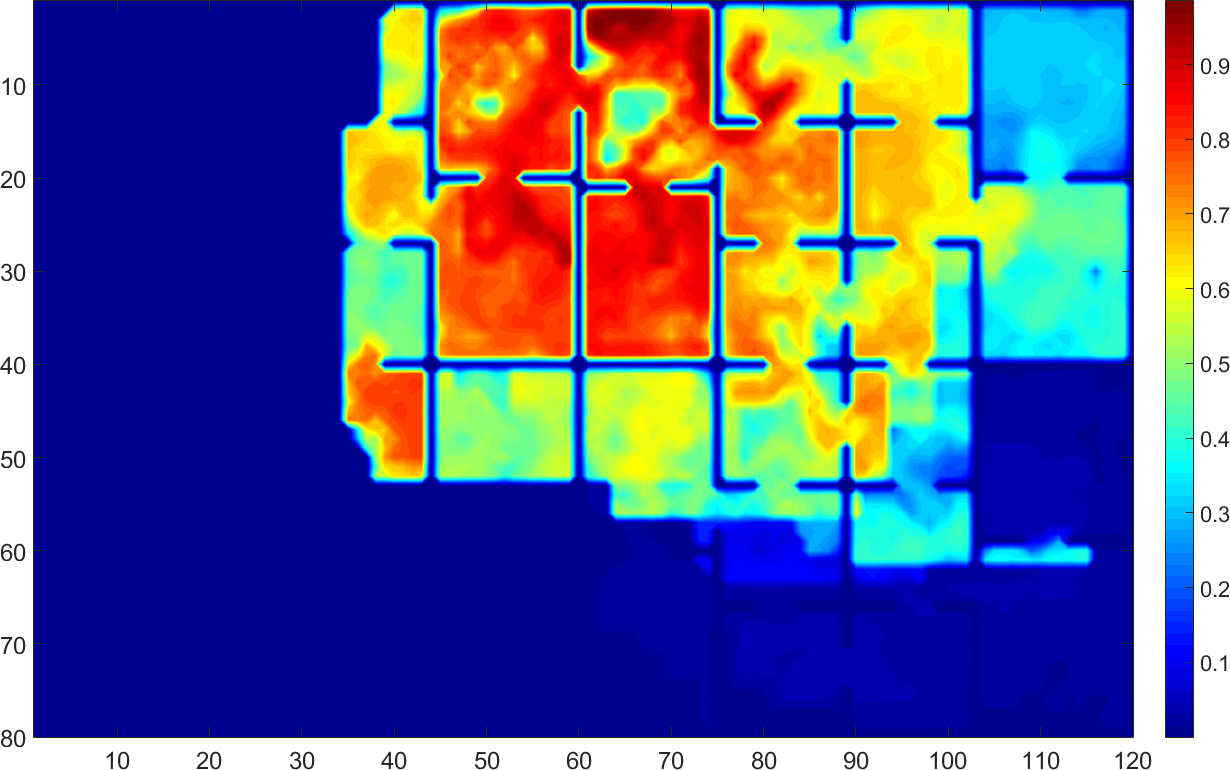}%
		\label{fig:exp_heatmap6}}
		
	\subfloat[PheroCom - robot.07]{
		\includegraphics[width=1.8in]{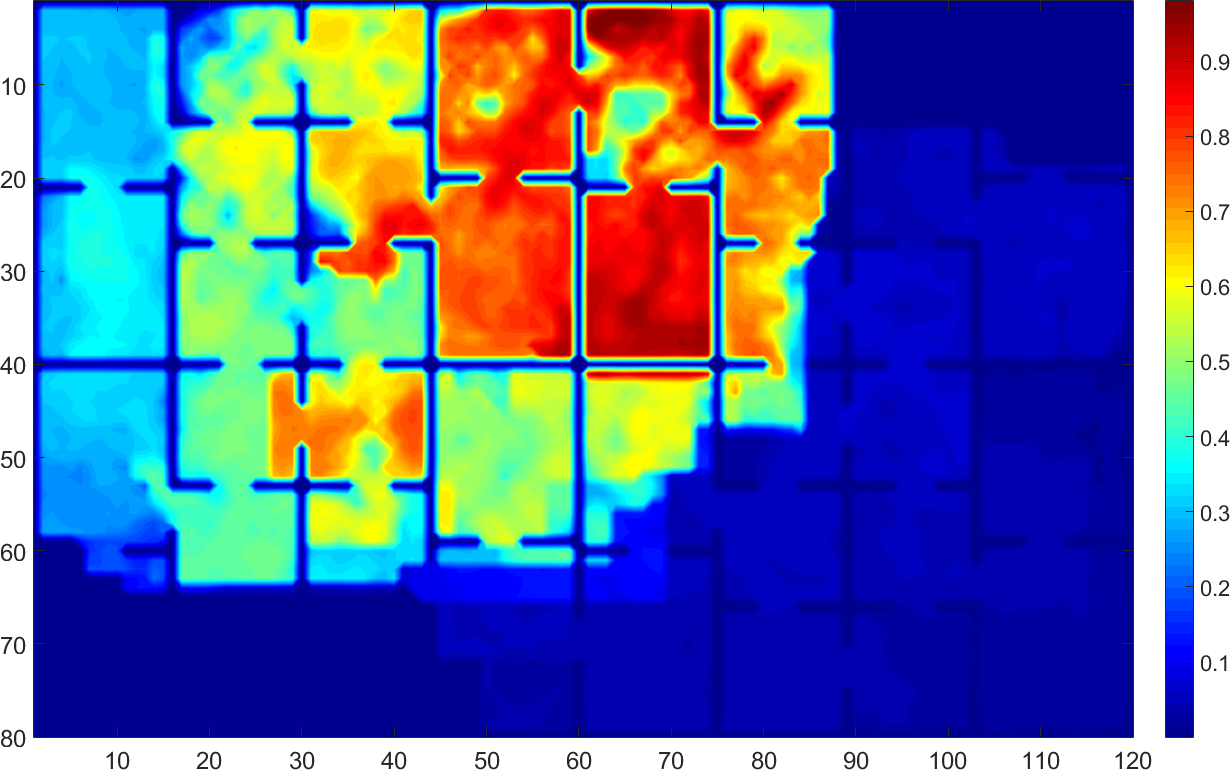}%
		\label{fig:exp_heatmap7}}
	\hfil
	\subfloat[PheroCom - robot.08]{
		\includegraphics[width=1.8in]{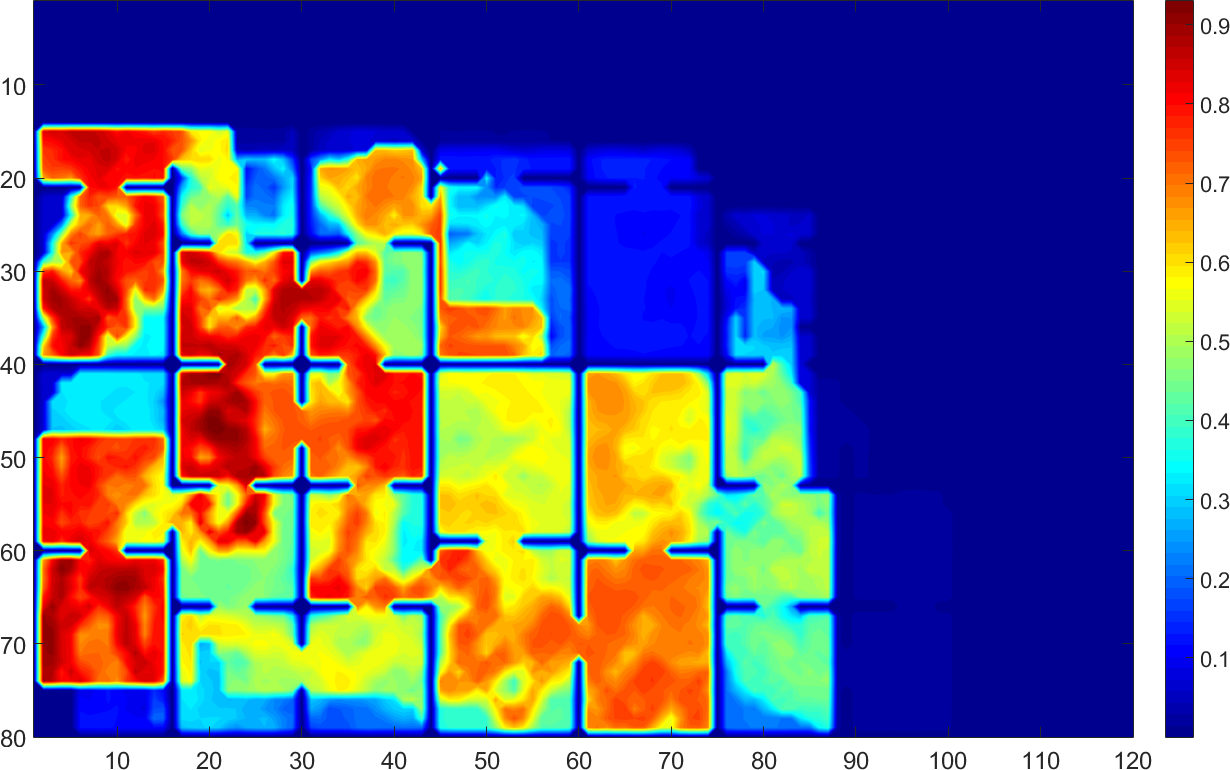}%
		\label{fig:exp_heatmap8}}
	\hfil
	\subfloat[PheroCom - robot.09]{
		\includegraphics[width=1.8in]{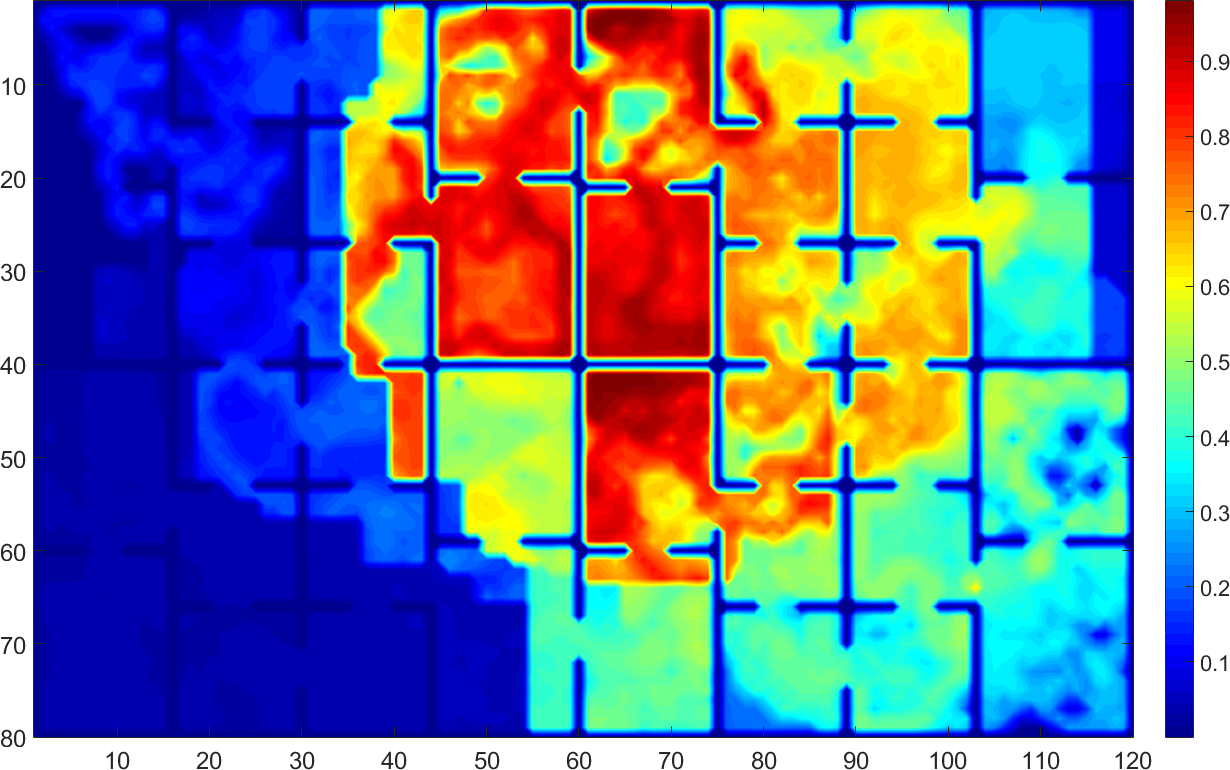}%
		\label{fig:exp_heatmap9}}
		
	\subfloat[PheroCom - robot.10]{
		\includegraphics[width=1.8in]{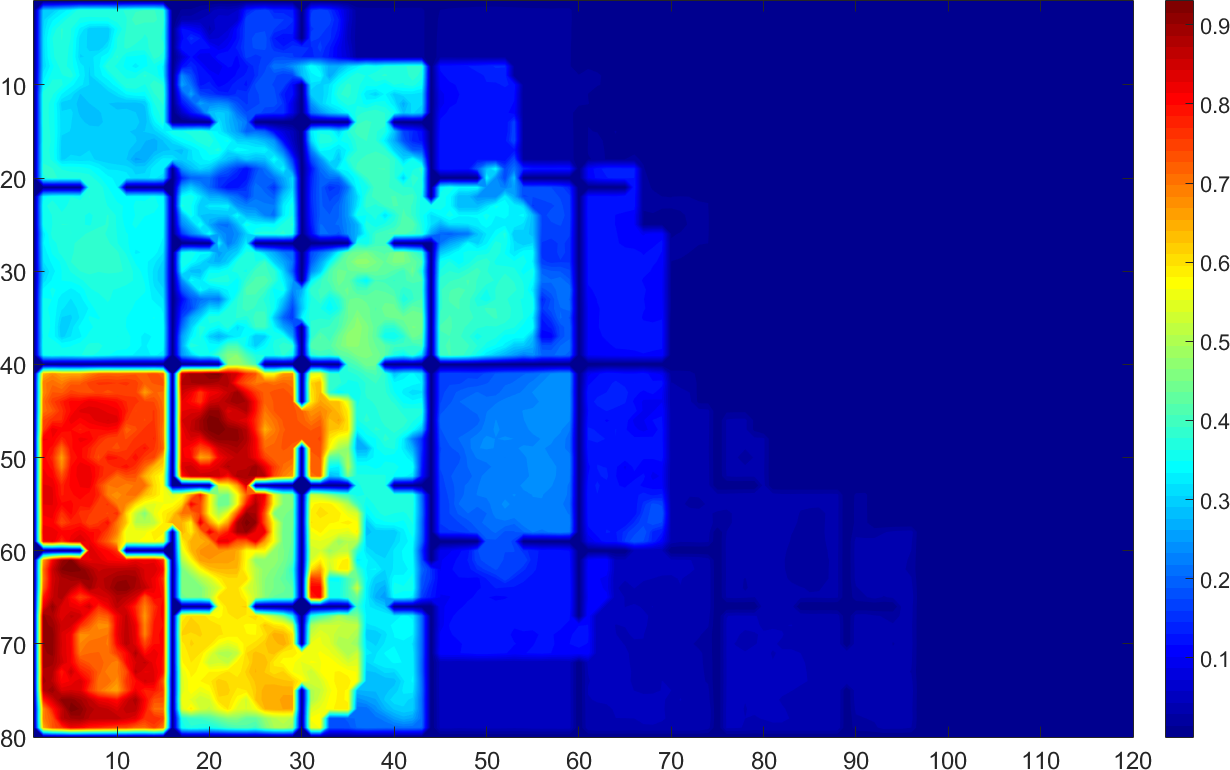}%
		\label{fig:exp_heatmap10}}
	\hfil
	\subfloat[PheroCom - robot.11]{
		\includegraphics[width=1.8in]{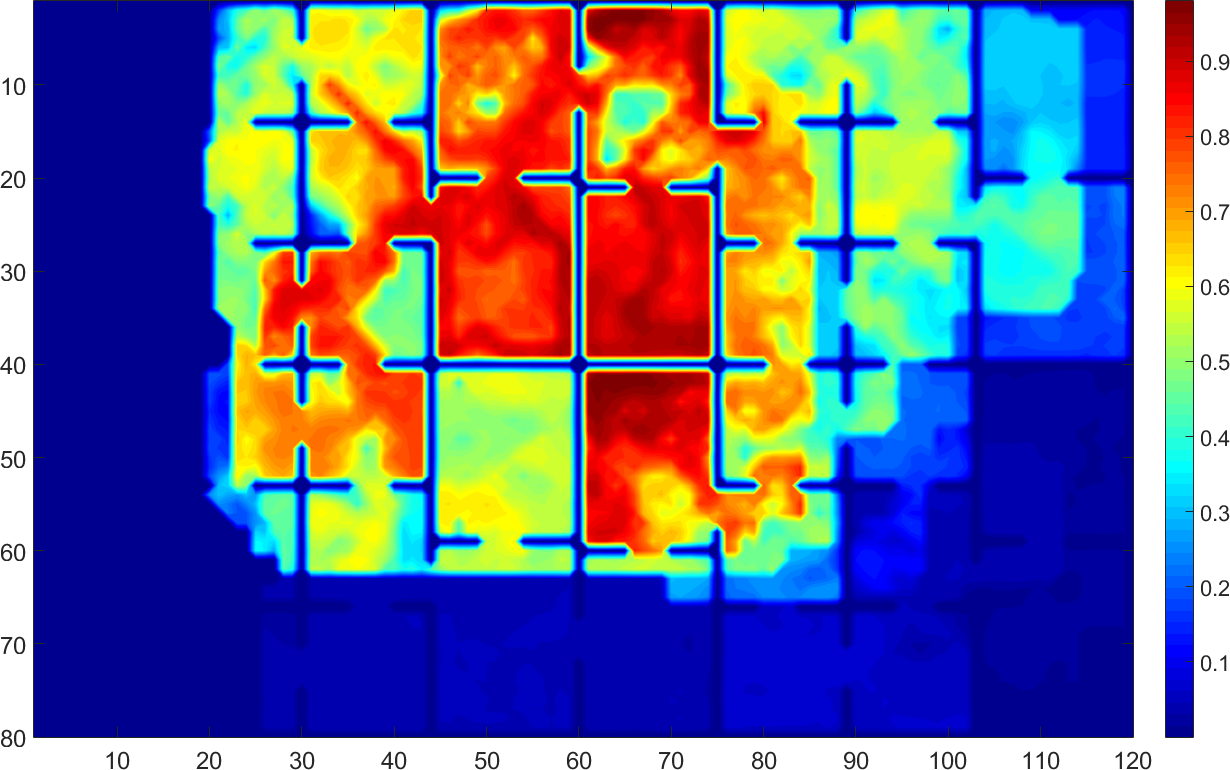}%
		\label{fig:exp_heatmap11}}
	\hfil
	\subfloat[PheroCom - robot.12]{
		\includegraphics[width=1.8in]{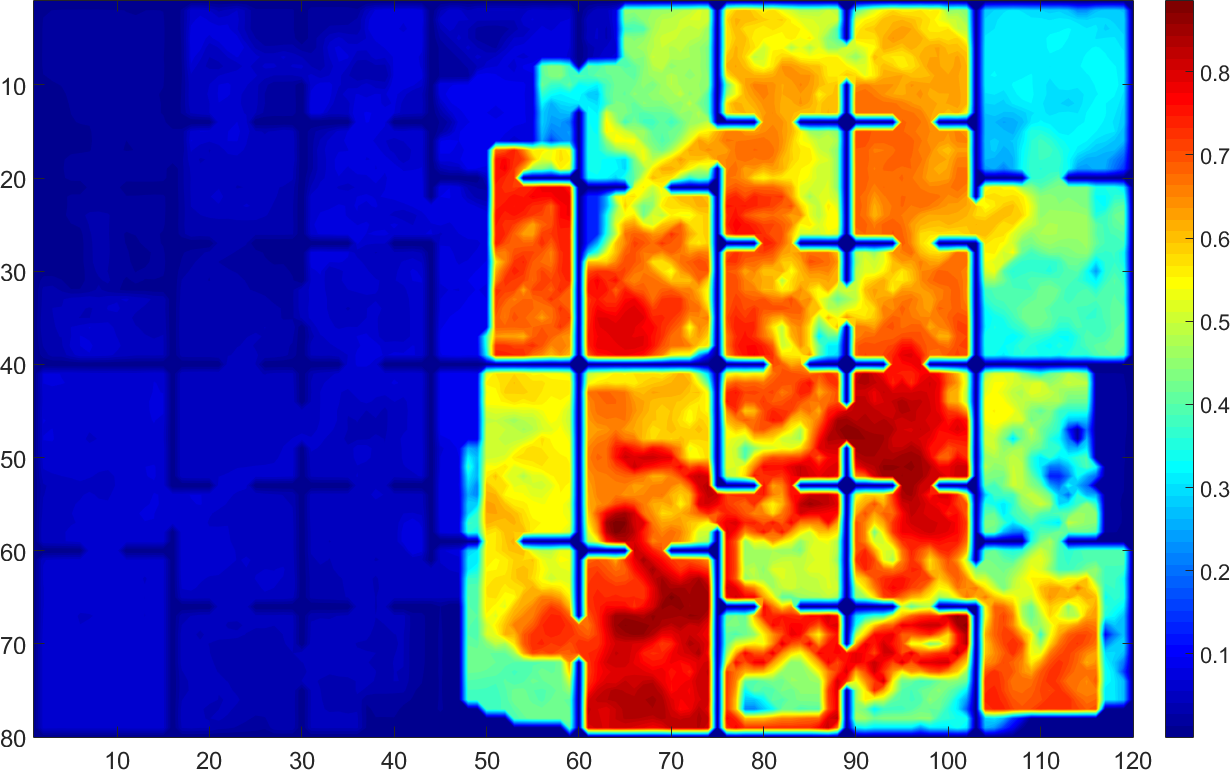}%
		\label{fig:exp_heatmap12}}
	
	\subfloat[IACA-DI (centralised).]{
		\includegraphics[width=1.8in]{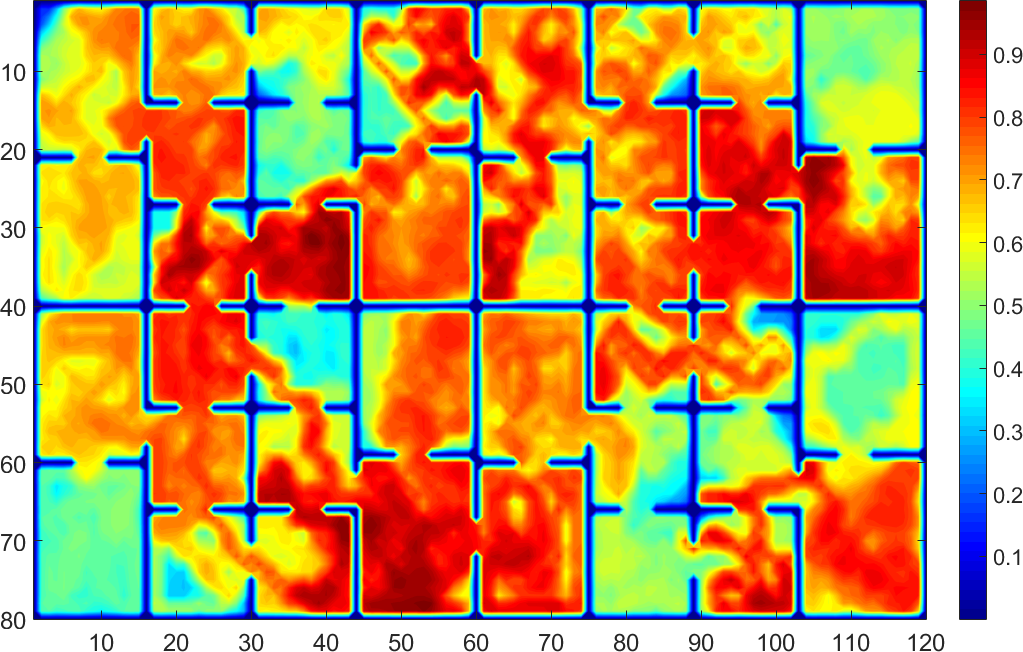}%
		\label{fig:exp_heatmap13}}
	\caption{Divergence between the pheromone heatmaps of the PheroCom model (decentralised and asynchronous) and the IACA-DI model (centralised and synchronous). Scenarios in which both models have similar performance.}
	\label{fig:heatmaps}
\end{figure}

Figure~\ref{fig:heatmaps} illustrates pheromone heatmaps from environment E4. The first figures (Figs.~\ref{fig:exp_heatmap1}~-~\ref{fig:exp_heatmap12}) are the heatmaps of the local memory of a swarm with twelve robots ($N = 12$) when the PheroCom model is applied, and the last one (Fig.~\ref{fig:exp_heatmap13}) the IACA-DI model. The heatmaps were extracted exactly at the last time step $(t = 120,000)$ of the simulations. It is noteworthy that the heatmaps of the PheroCom and IACA-DI models, represent equivalent outcomes since the amount of task-points achieved were similar, i.e., although different characteristics can be observed between the heatmaps, they were derived from swarms that have achieved similar performance.

In the figures of the PheroCom's heatmaps (Figs.~\ref{fig:exp_heatmap1}~-~\ref{fig:exp_heatmap12}), it can be seen that the robot does not have information about all areas of the environment, but only of a macro-region nearby. Although the local maps of each robot are not complete, since the swarm is spread throughout the environment, a possible combination of these maps would result in a map with characteristics equivalent to the heatmap of the IACA-DI model.

In our previous work~\cite{tinoco2018pheromone}, experiments involving pheromone heatmaps sought to verify whether the pheromone scattered around the environment allowed robots to properly perform their decision-making process. More specifically, it was observed whether the pheromone was spread evenly, without the presence of areas with zero concentration (never visited) and areas with full concentration (stagnant). Through empirical analysis, it was possible to establish the evaporation rate and swarm size for each environment. However, the pheromone heatmaps of the PheroCom model have different characteristics, as can be seen from Figure~\ref{fig:exp_heatmap1} to Figure~\ref{fig:exp_heatmap12}. Considering that both have similar efficiency regarding task-points, one can conclude that the robots do not need to simulate the entire map in their local memories, but only a region of interest. Consequently, condensing the simulation into a specific region entails a considerable drop in processing costs.

\subsubsection{Cellsteps maps analysis}
Cellsteps maps aim to graphically show the scattering of the swarm throughout the environment. Each time a robot pass through a cell, a cellstep is counted. In these maps, cells with warmer colours (tending to red) show a higher mean of cellsteps and cells with colder colours (tending to blue) have lower or no cellsteps. Thus, it is possible to observe whether the spread is homogeneous or whether there are areas not being monitored at an adequate frequency. Unlike pheromone heatmaps, which illustrate a specific moment in the simulation, cellsteps maps accumulate the entire evolution of the model. The main contribution of this experiment is to verify the similarities of the swarm behaviour. It is expected that the cellsteps maps of the PheroCom model have characteristics similar to those of the IACA-DI model.

\begin{figure}[b]
	\centering
	\subfloat[PheroCom (decentralised)]{
		\includegraphics[width=1.85in]{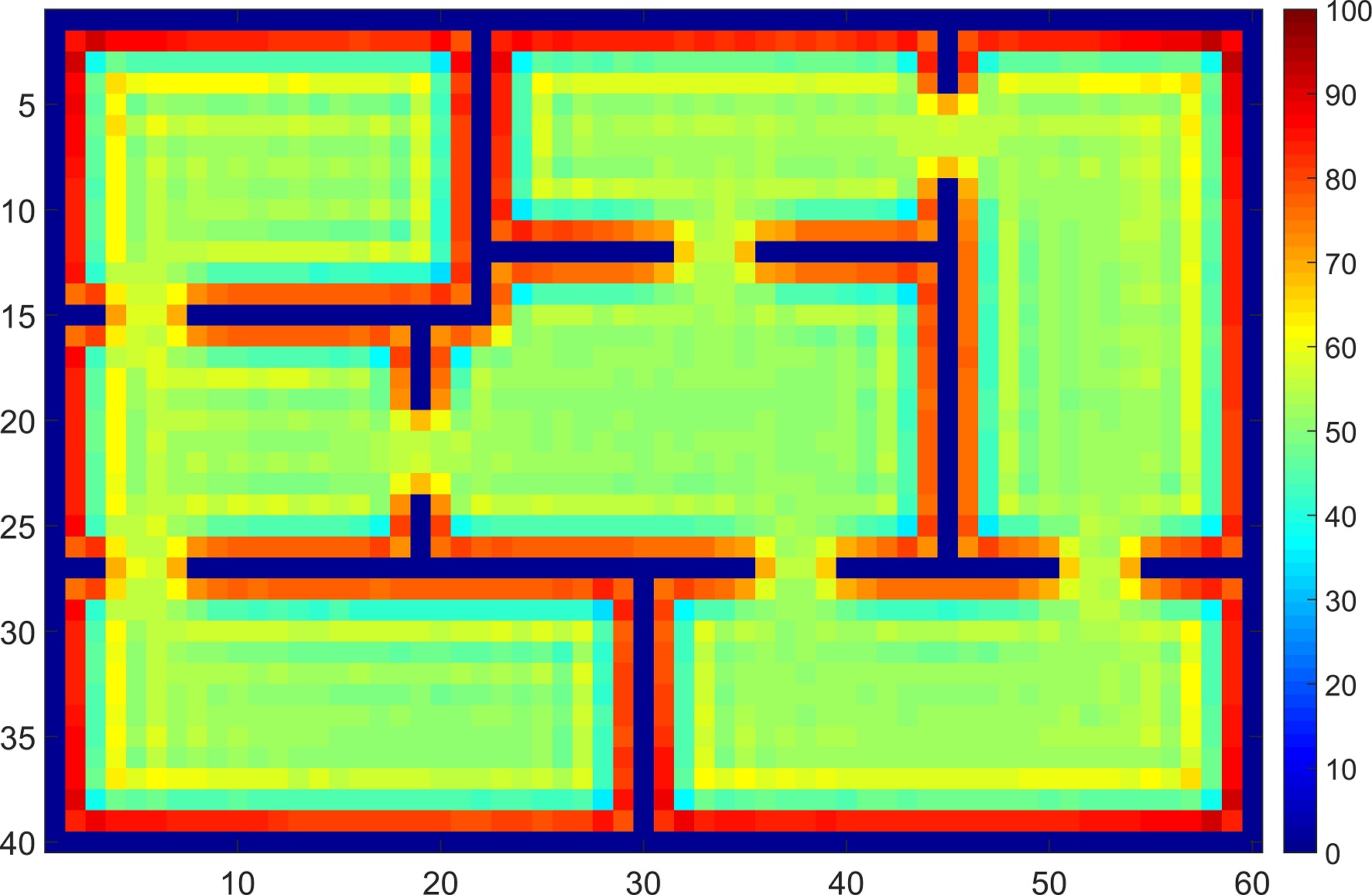}%
		\label{fig:exp_cellsteps1}}
	\hfil
	\subfloat[IACA-DI (centralised)]{
		\includegraphics[width=1.85in]{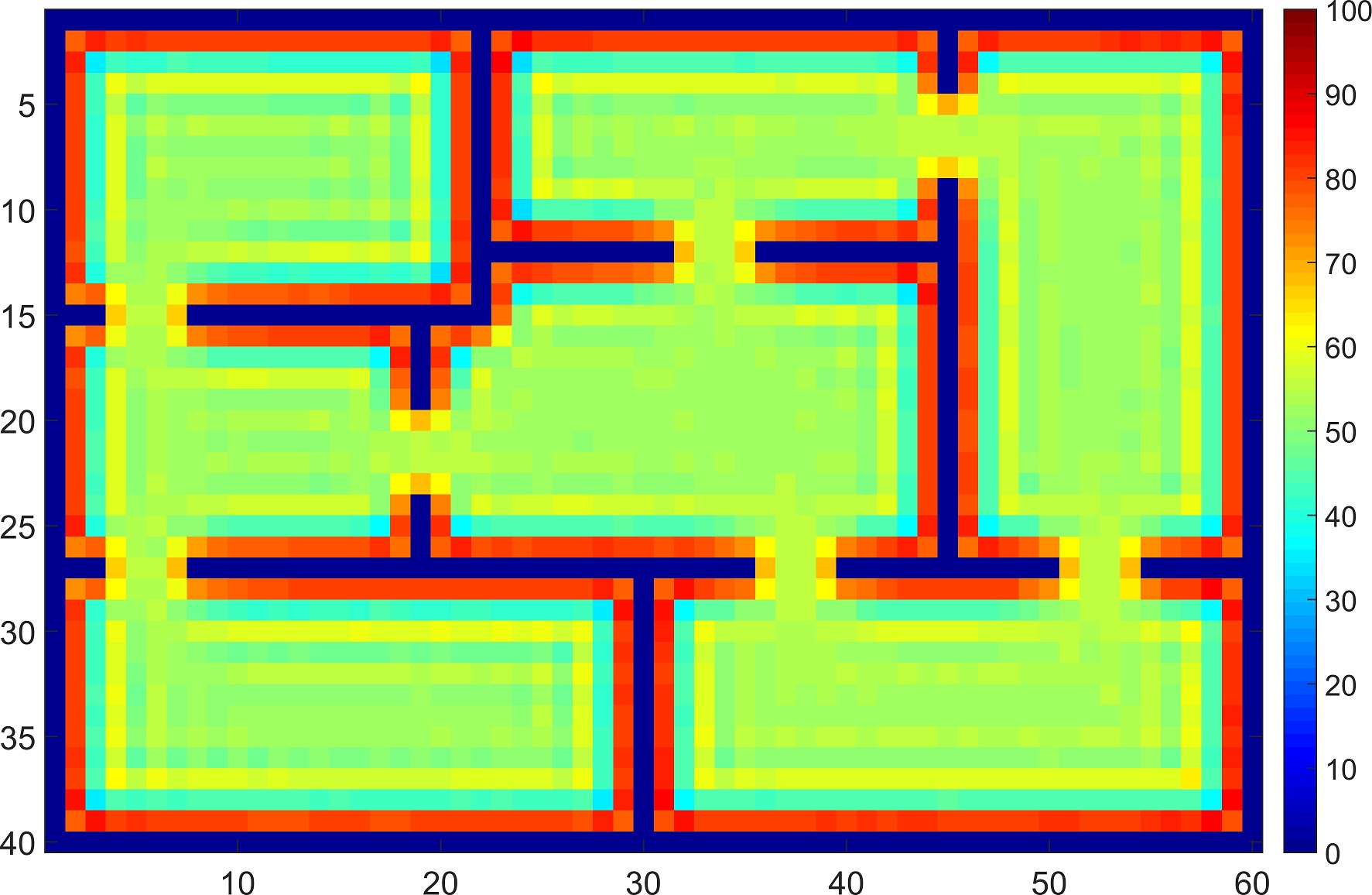}%
		\label{fig:exp_cellsteps2}}
	\hfil	
	\subfloat[Cells subtr. \ref{fig:exp_cellsteps1} and \ref{fig:exp_cellsteps2}]{
		\includegraphics[width=1.65in]{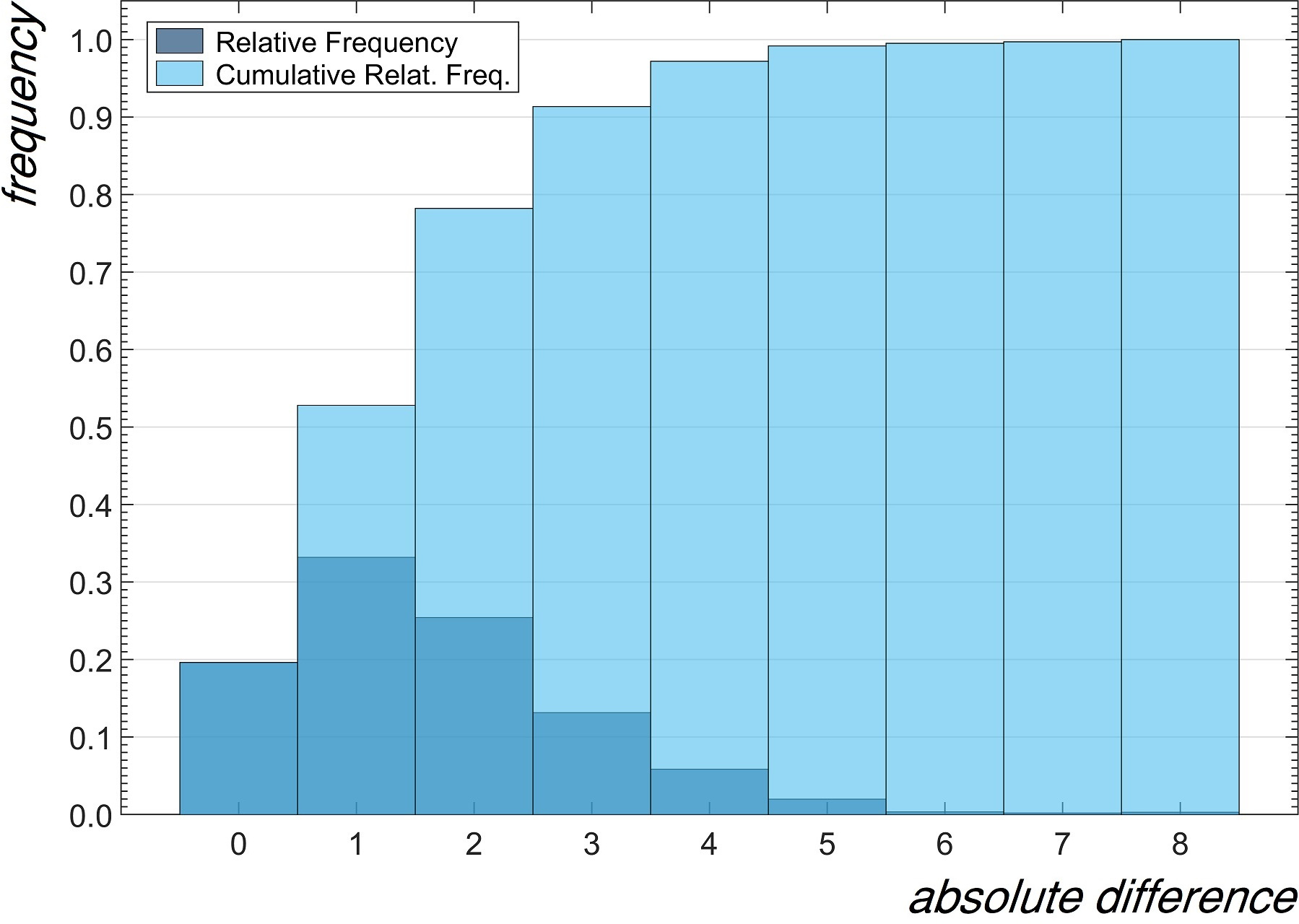}%
		\label{fig:exp_cellsteps3}}
		
	\subfloat[PheroCom (decentralised)]{
		\includegraphics[width=1.85in]{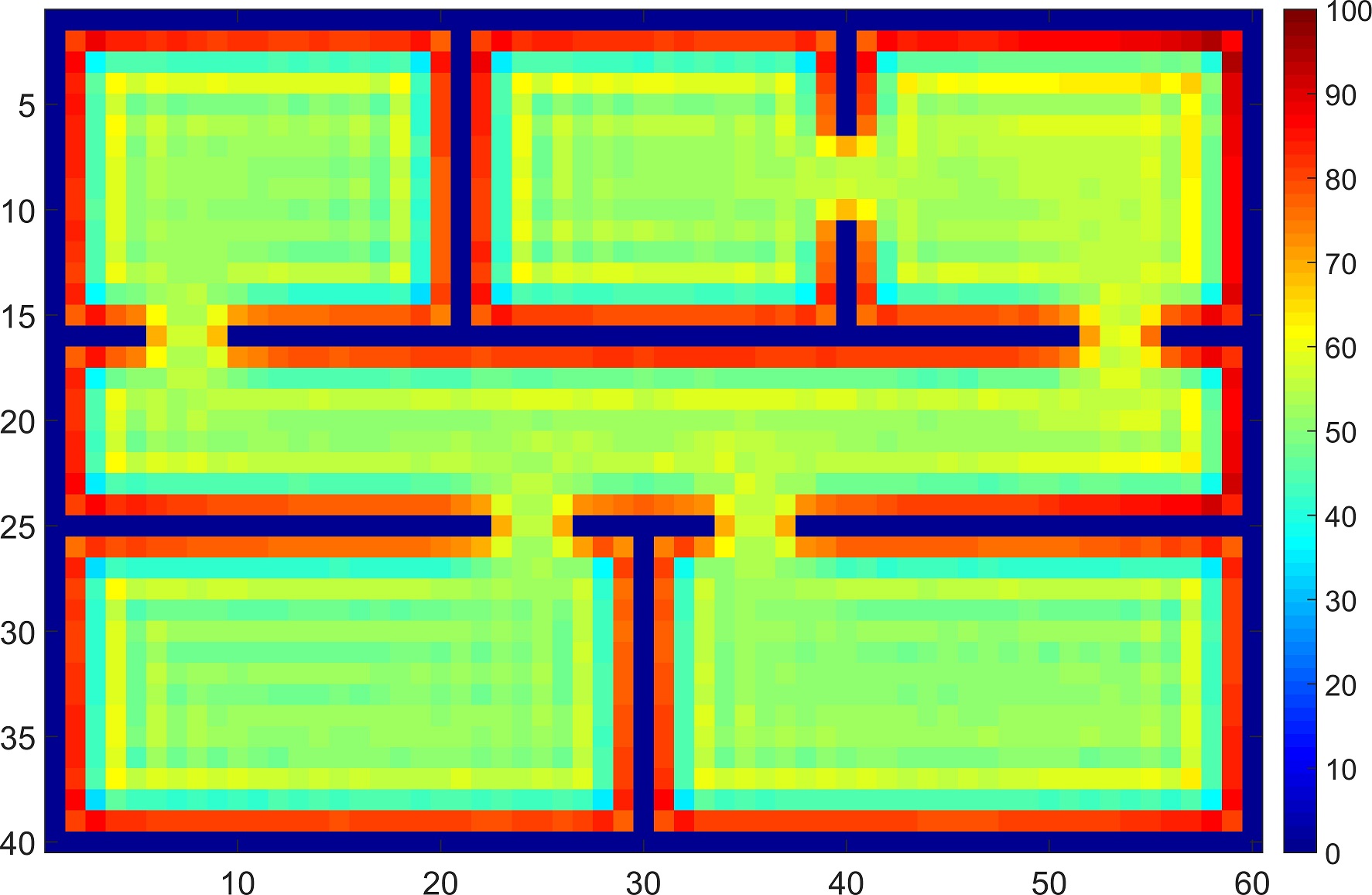}%
		\label{fig:exp_cellsteps4}}
	\hfil
	\subfloat[IACA-DI (centralised)]{
		\includegraphics[width=1.85in]{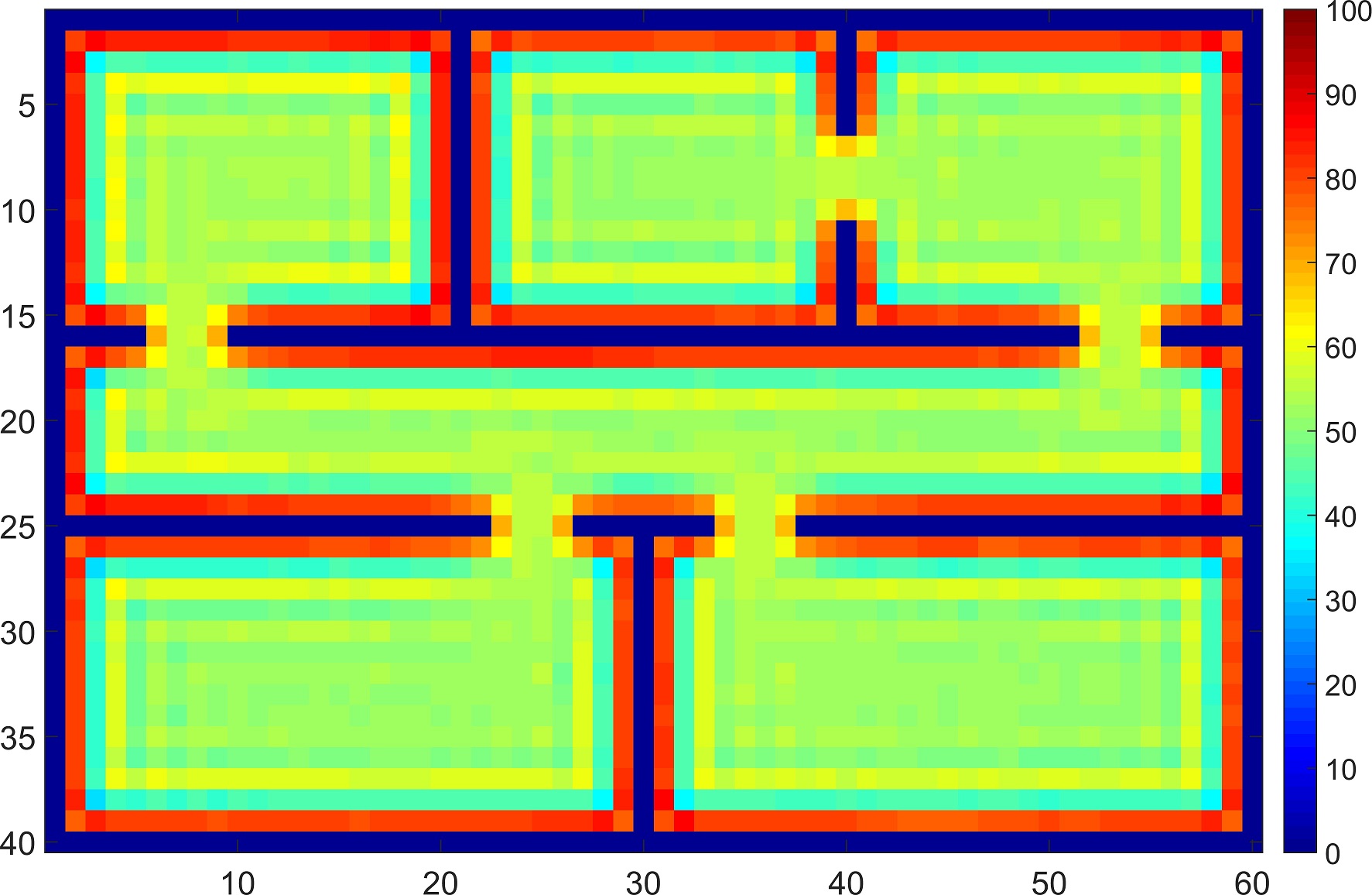}%
		\label{fig:exp_cellsteps5}}
	\hfil	
	\subfloat[Cells subtr. \ref{fig:exp_cellsteps4} and \ref{fig:exp_cellsteps5}]{
		\includegraphics[width=1.65in]{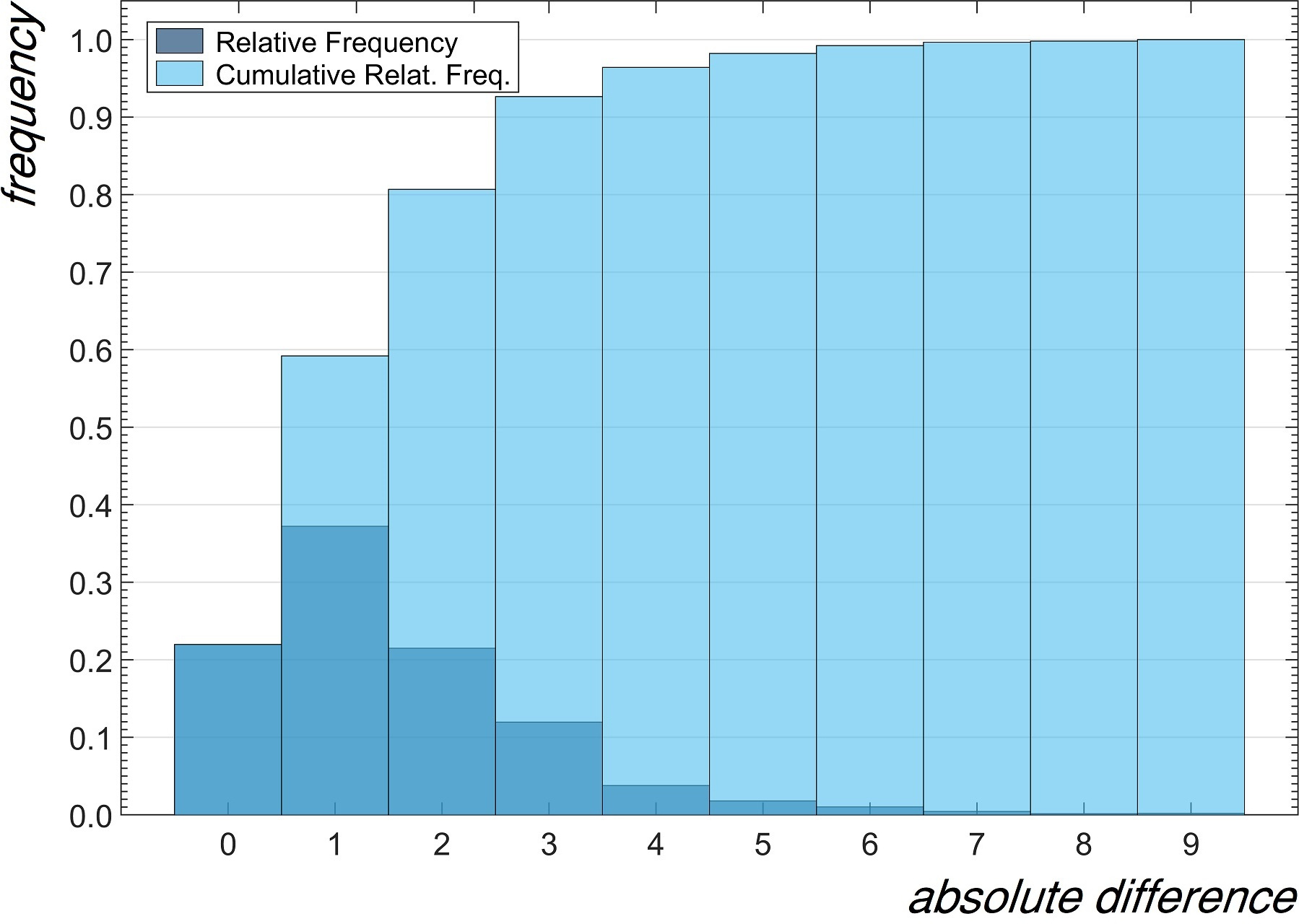}%
		\label{fig:exp_cellsteps6}}
		
	\subfloat[PheroCom (decentralised)]{
		\includegraphics[width=1.85in]{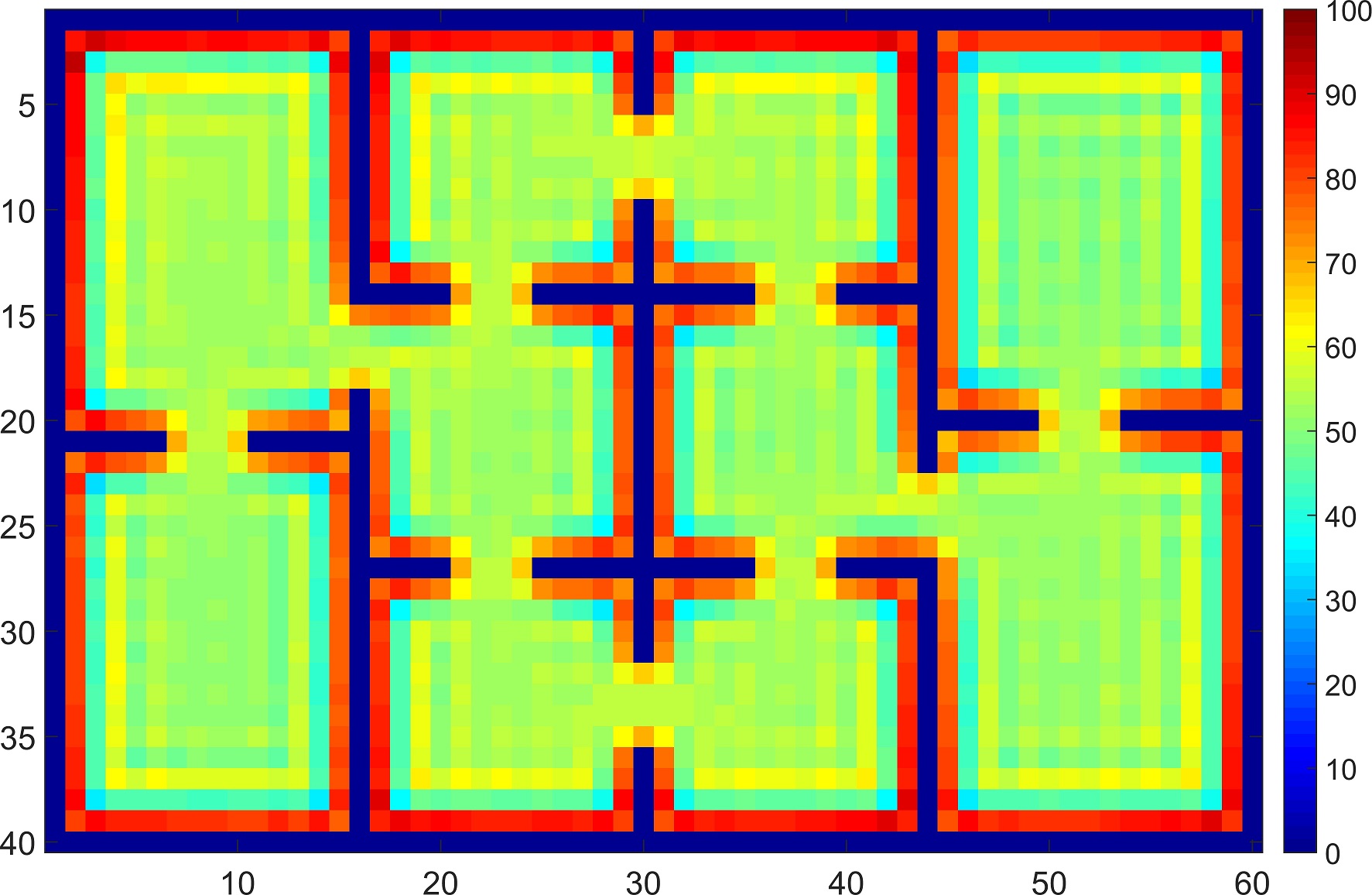}%
		\label{fig:exp_cellsteps7}}
	\hfil
	\subfloat[IACA-DI (centralised)]{
		\includegraphics[width=1.85in]{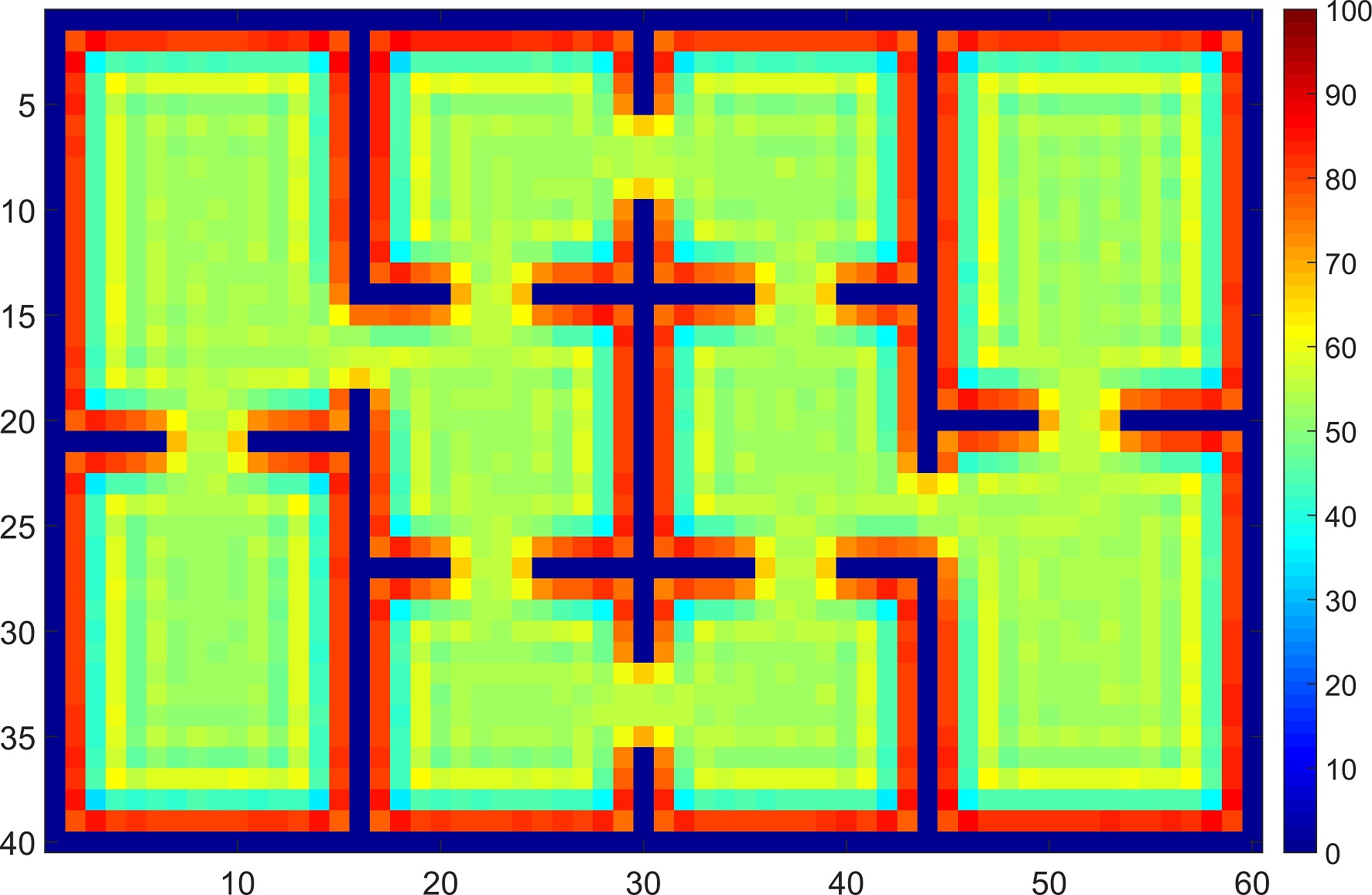}%
		\label{fig:exp_cellsteps8}}
	\hfil	
	\subfloat[Cells subtr. \ref{fig:exp_cellsteps7} and \ref{fig:exp_cellsteps8}]{
		\includegraphics[width=1.65in]{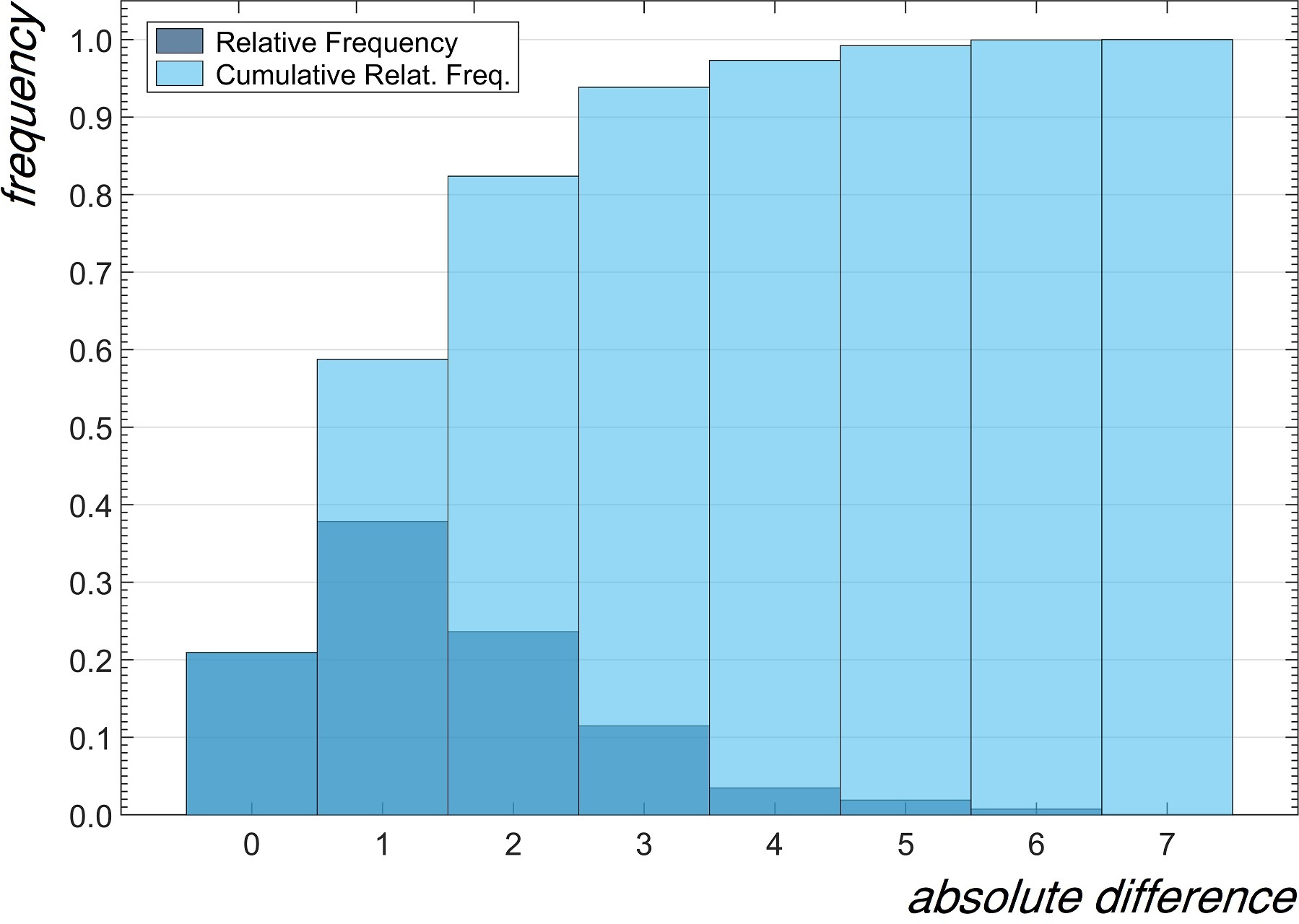}%
		\label{fig:exp_cellsteps9}}
		
	\subfloat[PheroCom (decentralised)]{
		\includegraphics[width=1.85in]{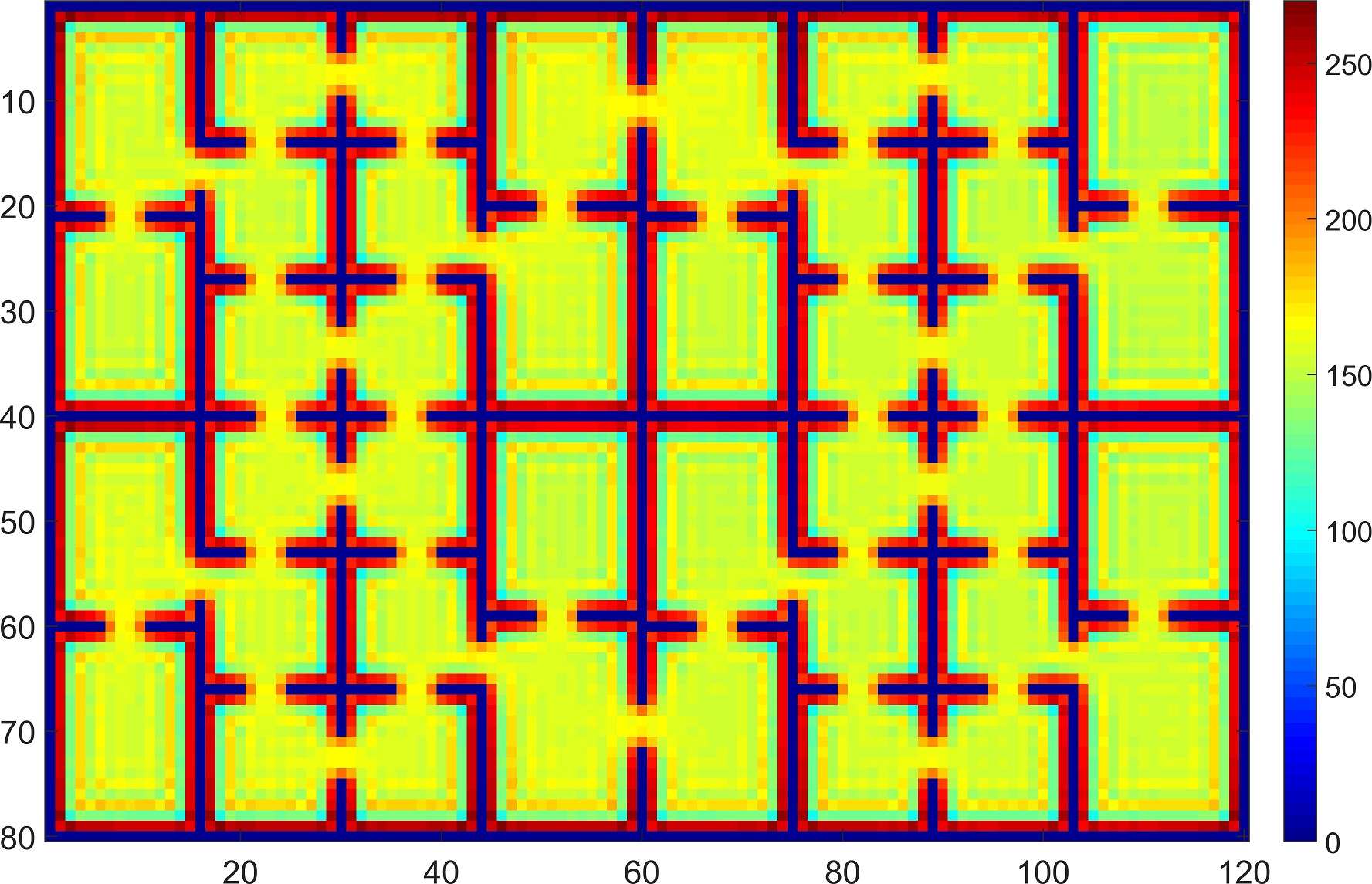}%
		\label{fig:exp_cellsteps10}}
	\hfil
	\subfloat[IACA-DI (centralised)]{
		\includegraphics[width=1.85in]{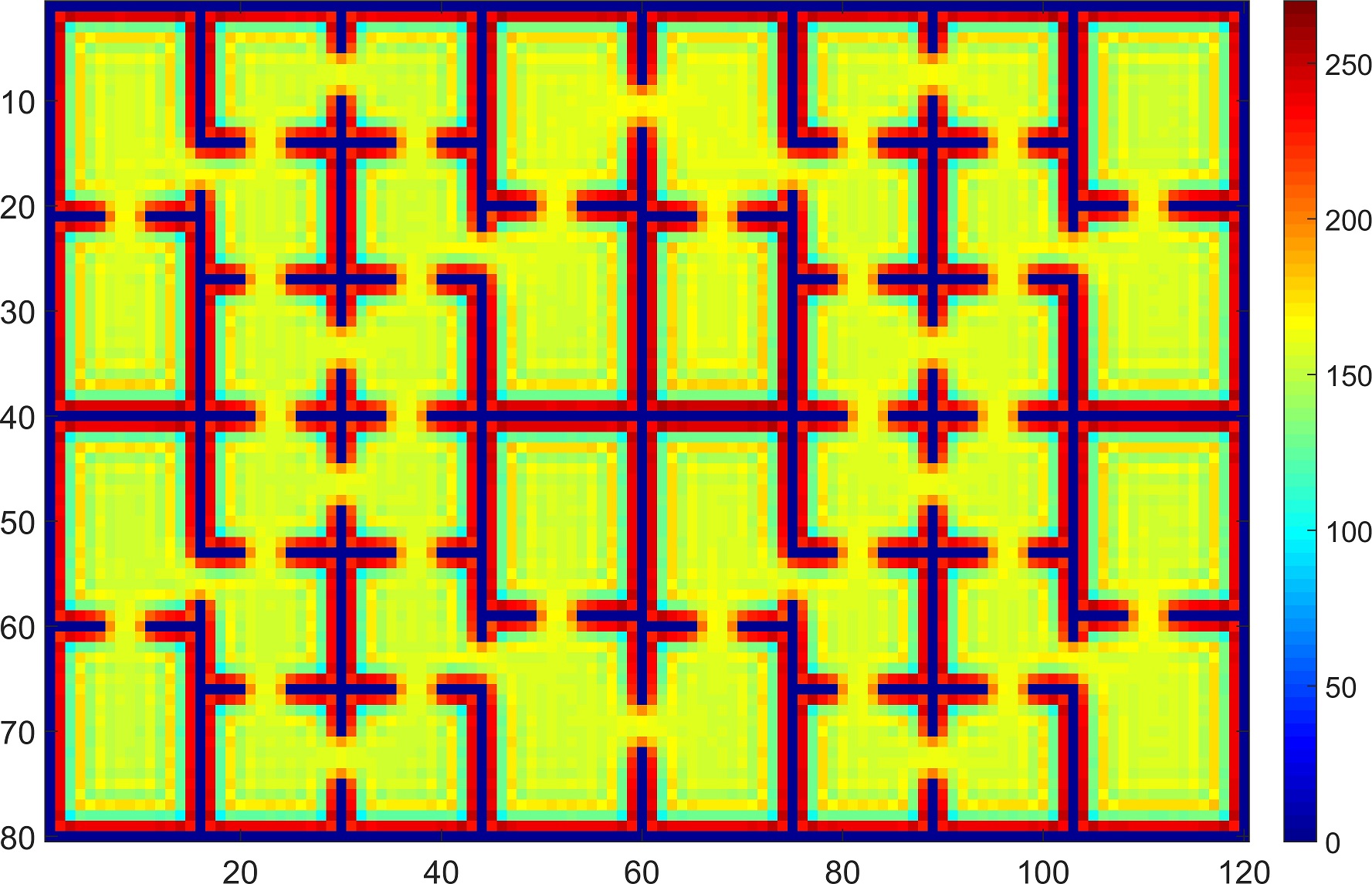}%
		\label{fig:exp_cellsteps11}}
	\hfil	
	\subfloat[Cells subtr. \ref{fig:exp_cellsteps10} and \ref{fig:exp_cellsteps11}]{
		\includegraphics[width=1.65in]{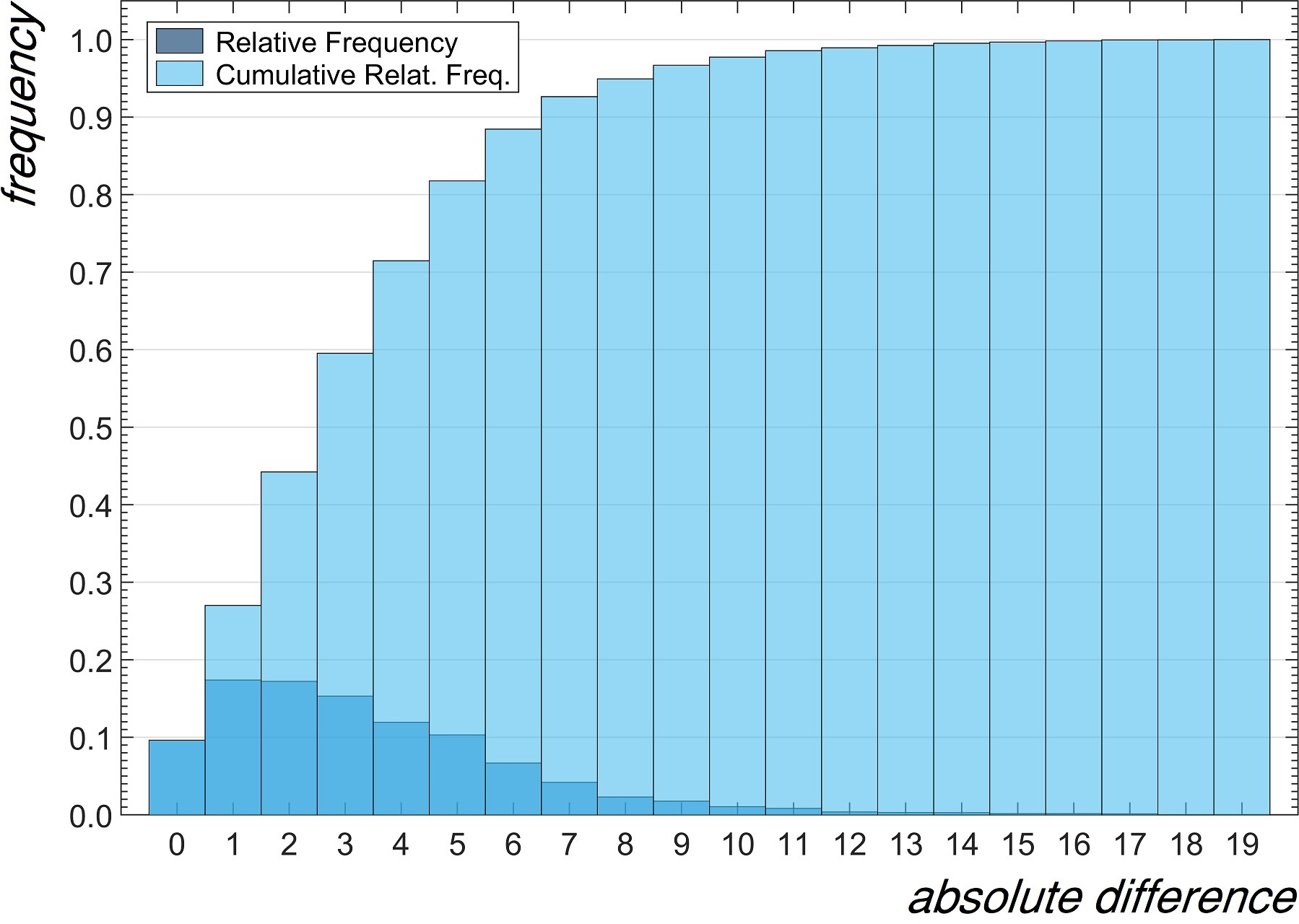}%
		\label{fig:exp_cellsteps12}}
	\caption{Divergence between the cellsteps maps of the PheroCom model (decentralised and asynchronous) and the IACA-DI model (centralised and synchronous). Scenarios in which both models have similar performance.}
	\label{fig:exp_cellsteps}
\end{figure}

Figure \ref{fig:exp_cellsteps} illustrates different cellstep maps along with histogram charts for a quantitative comparison. For each pair of cellstep maps (one of the PheroCom model and the other of the IACA-DI model), a histogram with frequencies was added to facilitate and formalise the similarity analysis between the models. These histograms are constructed employing the \textit{relative frequency} and the \textit{cumulative relative frequency} of the absolute differences of each cell in the cellstep maps of the models, i.e., it is a similarity measure of cells that are in the same position in the cellstep maps of both models.

According to the results, it can be seen that the swarm behaviour using the PheroCom model is similar to the IACA-DI model. This conclusion is based on the fact that the distribution of cellsteps in the environment are extremely similar. Considering the charts of relative and accumulated frequencies, it is possible to observe that $\approx90\%$ of the cells, in environments E1 (Figs.~\ref{fig:exp_cellsteps1} and \ref{fig:exp_cellsteps2}), E2 (Figs.~\ref{fig:exp_cellsteps4} and \ref{fig:exp_cellsteps5}) and E3 (Figs.~\ref{fig:exp_cellsteps7} and \ref{fig:exp_cellsteps8}), have a difference from $1$ to $3$ cellsteps. Furthermore, these three environments have $\approx20\%$ of identical cells (same number of cellsteps). In the E4 environment (Figs.~\ref{fig:exp_cellsteps10} and \ref{fig:exp_cellsteps11}), which have larger dimensions, this difference is from $1$ to $7$ cellsteps, in which $\approx10\%$ of the cells have identical values and $\approx50\%$ have a difference from $1$ to $3$ cellsteps. This quantitative analysis confirms the degree of similarity between the two models. Therefore, since the IACA-DI model has already demonstrated its effectiveness and efficiency \cite{tinoco2017improved}, this similarity analysis demonstrates that the swarm scattering capacity using the PheroCom model is equivalent to that of the IACA-DI model.

\subsection{Webots Simulation Platform}
\label{exp:sec_webots_platform}
The experiments presented in Section \ref{exp:sec_simple_platform} were performed using the MaSS, which, in turn, does not have a graphical interface. The main objective of using this environment is to allow the execution of mass experiments, evaluating different scenarios over many time steps. Although these results are an approximation of the real behaviour, mainly because the model uses a CA, it is necessary to perform experiments on a simulation platform that takes into account more realistic aspects, like physics and hardware. 

\begin{figure}[b]
	\centering
	\subfloat[PheroCom (5min)]{
		\includegraphics[width=1.5in]{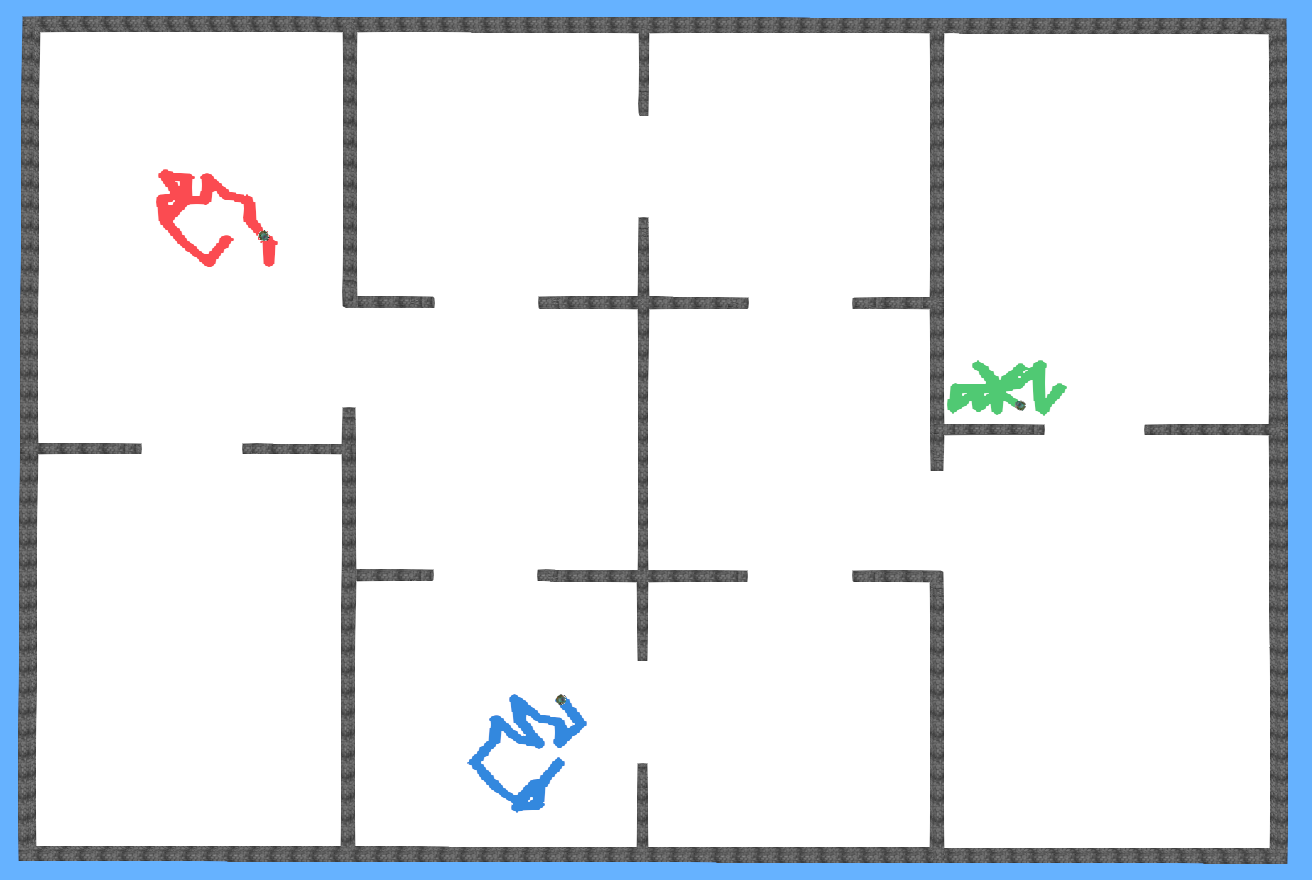}}
	\hfil
	\subfloat[PheroCom (1hr)]{
		\includegraphics[width=1.5in]{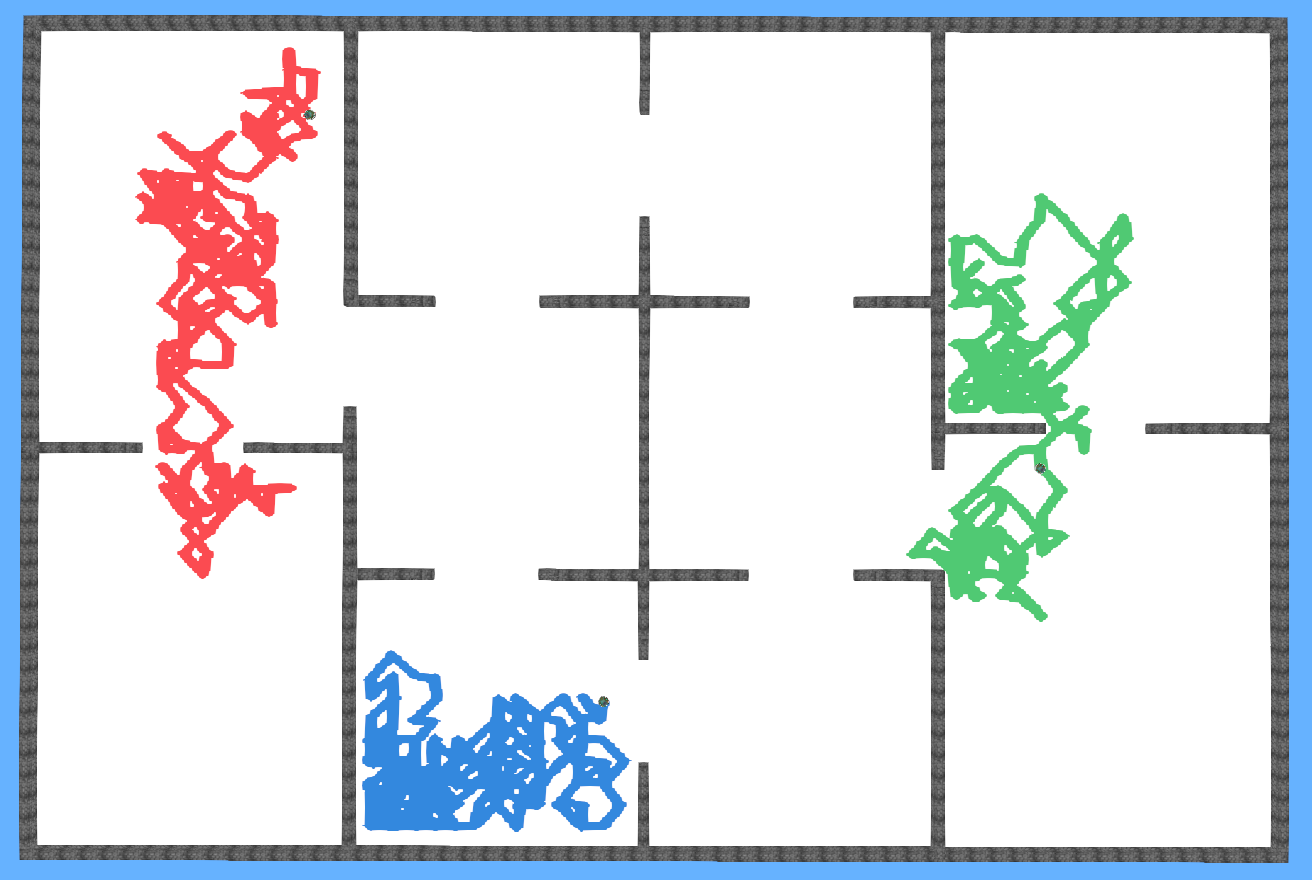}}
	\hfil
	\subfloat[PheroCom (5hrs)]{
		\includegraphics[width=1.5in]{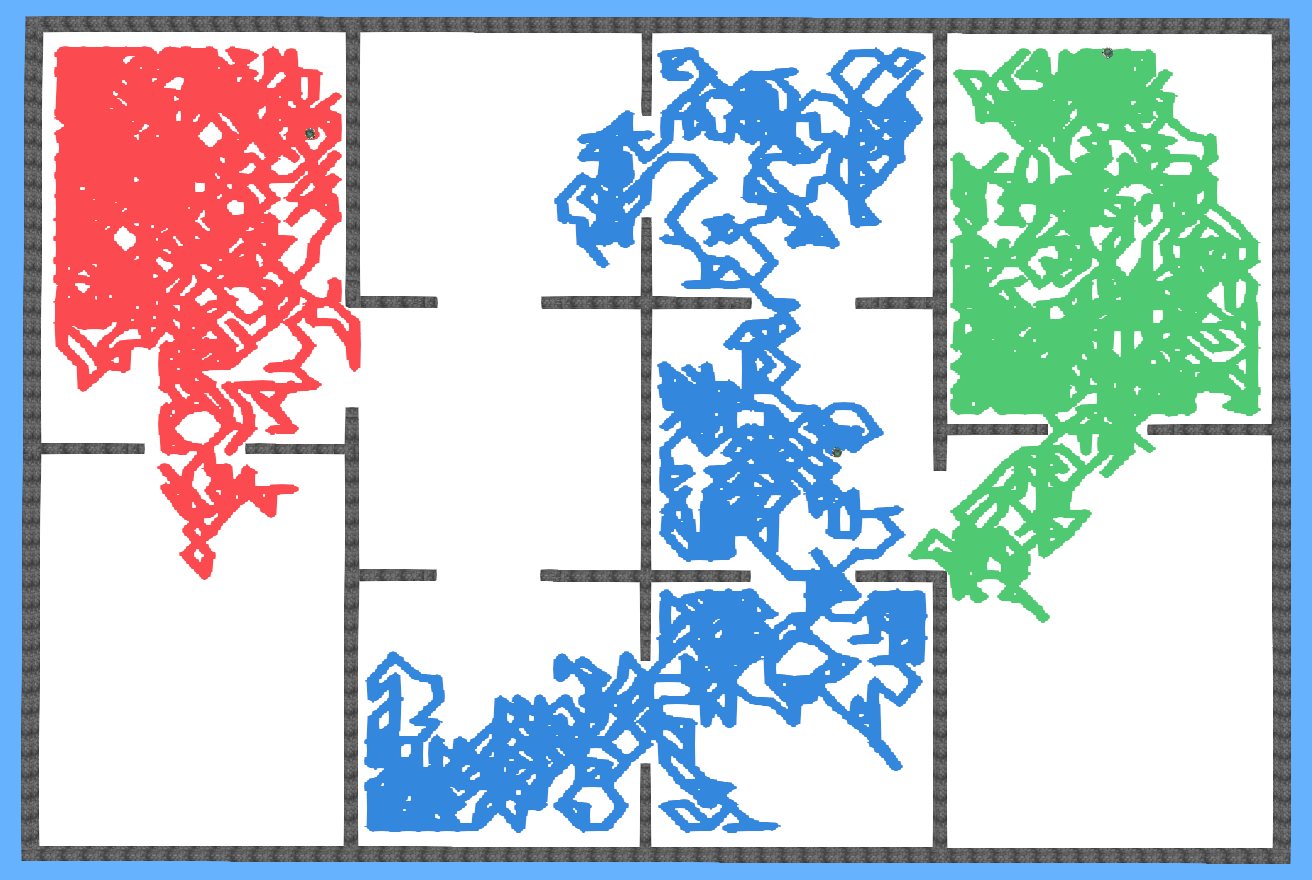}}
	\hfil
	\subfloat[PheroCom (10hrs)]{
		\includegraphics[width=1.5in]{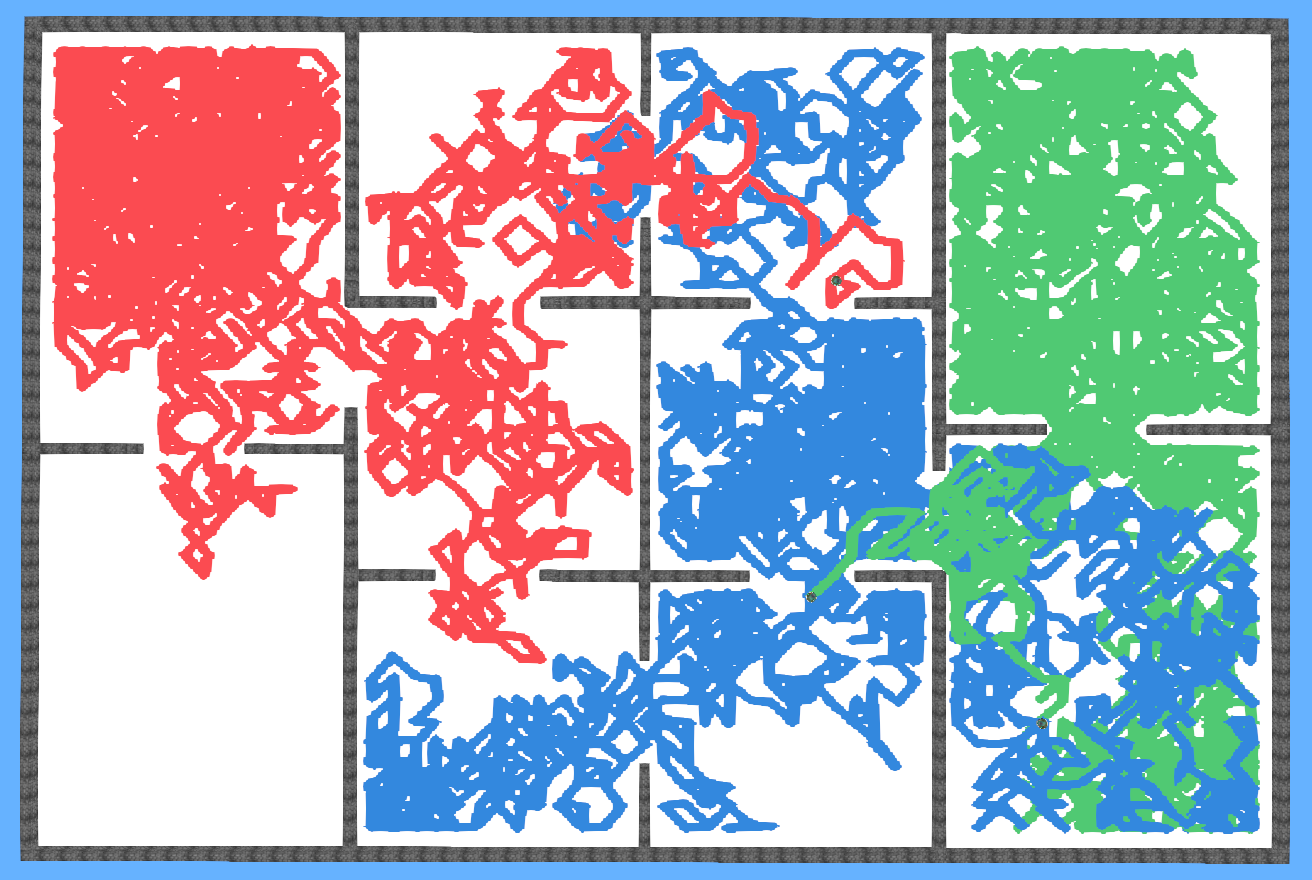}}
	\caption{Experiment with trails: environment E3' and 3 robots (each robot is represented by a RGB colour) with the Random movement strategy.}
	\label{random_trails}
\end{figure}

\begin{figure}[t]
	\centering
	\subfloat[PheroCom (5min)]{
		\includegraphics[width=1.5in]{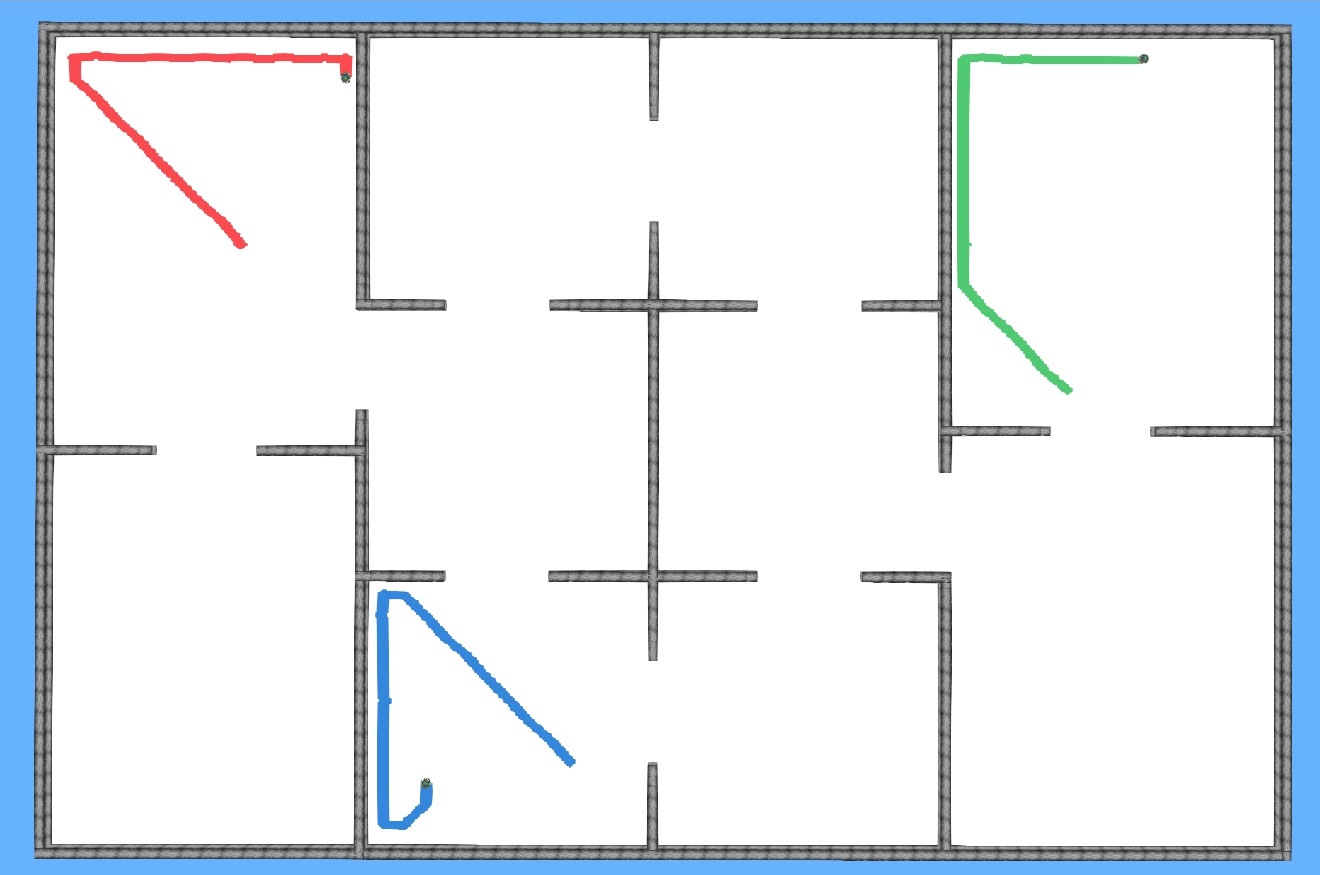}}
	\hfil
	\subfloat[PheroCom (1hr)]{
		\includegraphics[width=1.5in]{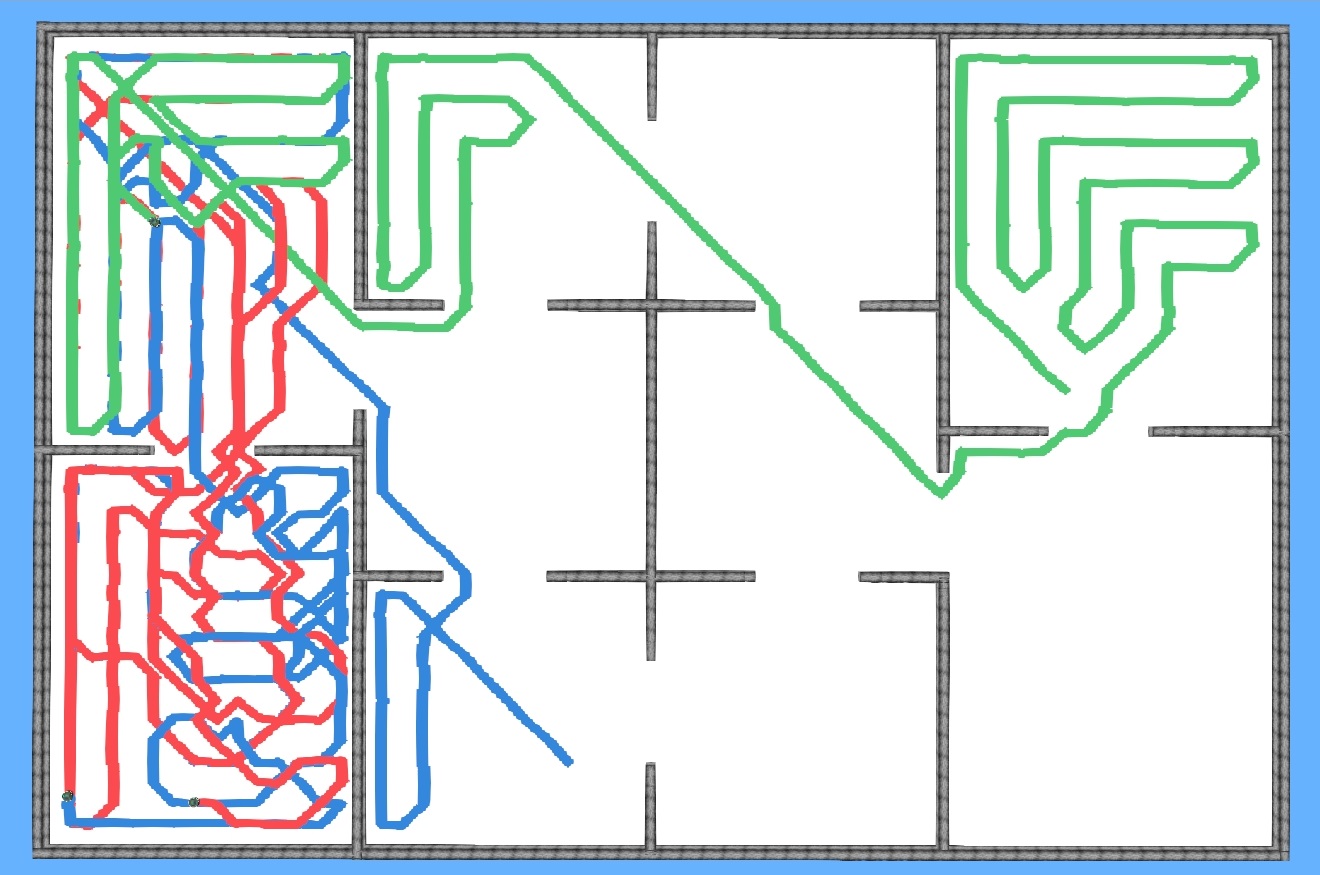}}
	\hfil
	\subfloat[PheroCom (5hrs)]{
		\includegraphics[width=1.5in]{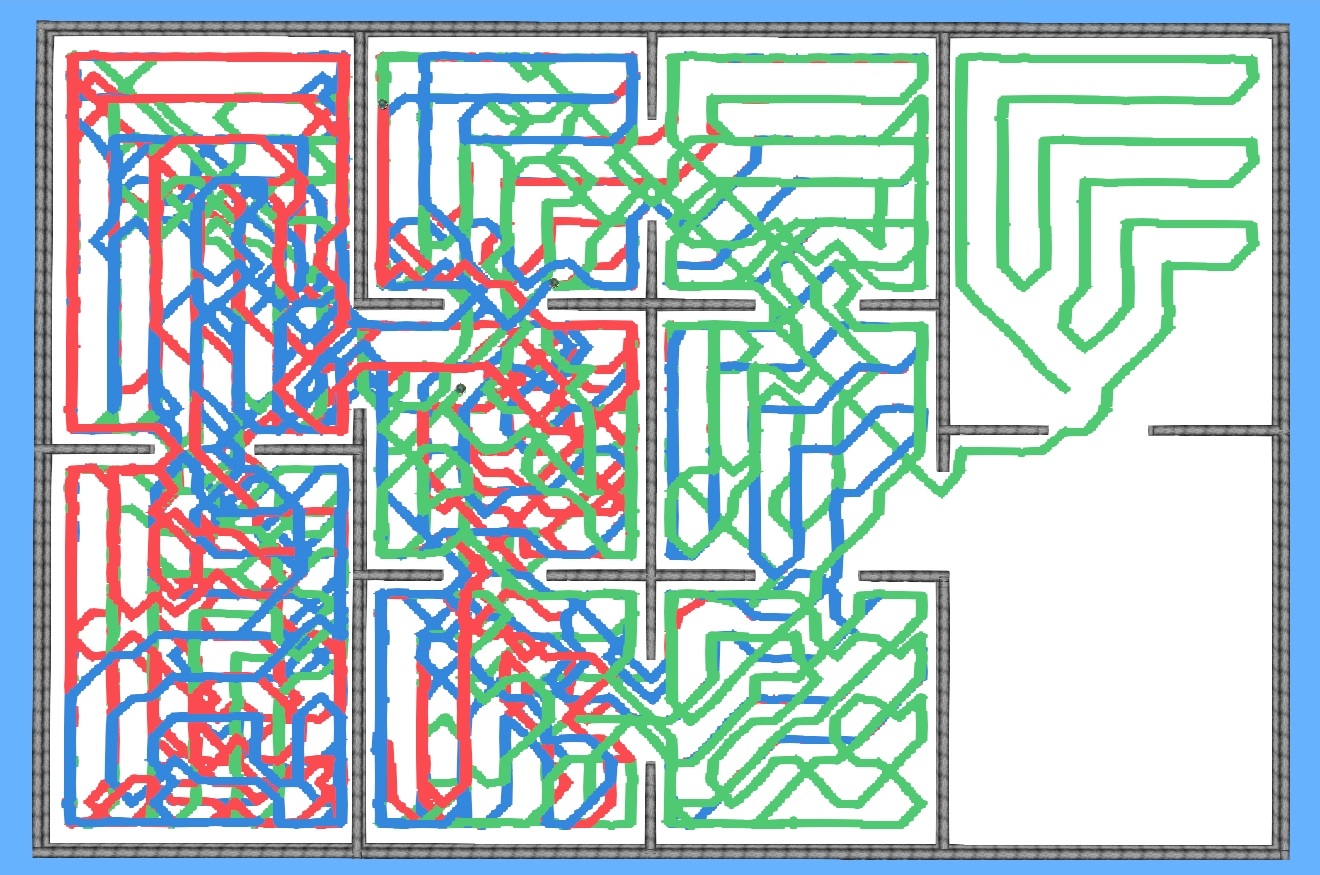}}
	\hfil
	\subfloat[PheroCom (10hrs)]{
		\includegraphics[width=1.5in]{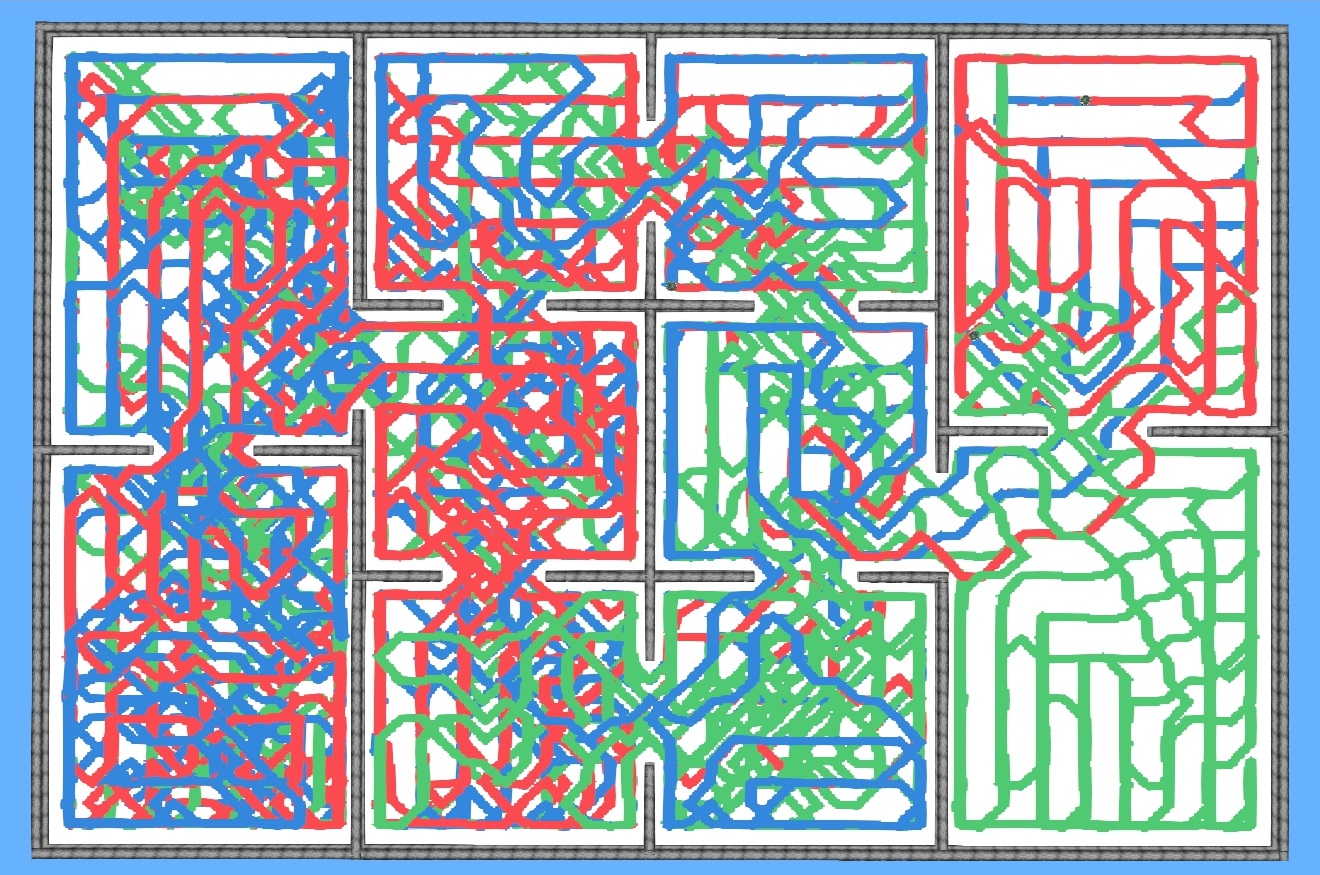}}
		
	\subfloat[IACA-DI (5min)]{
		\includegraphics[width=1.5in]{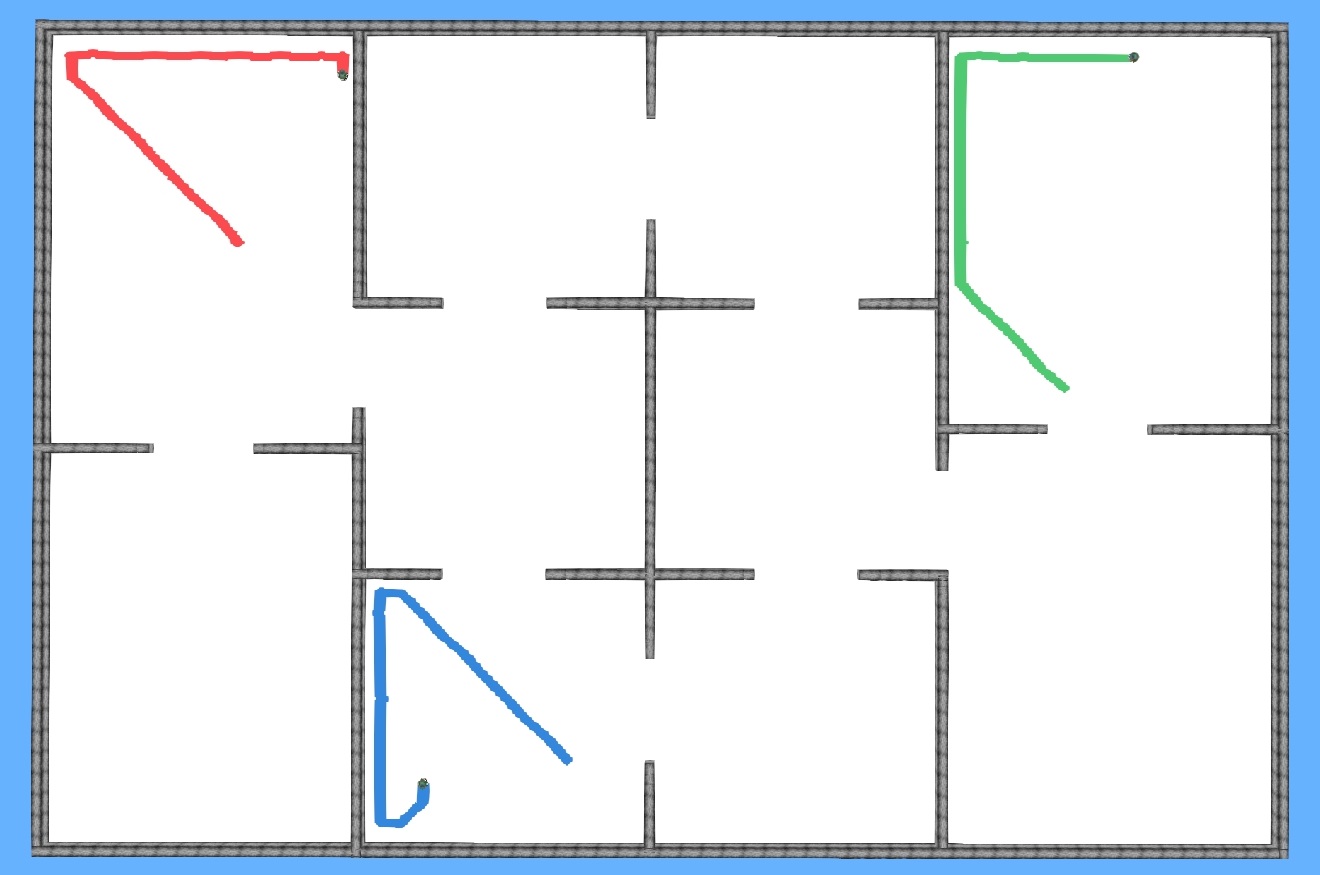}}
	\hfil
	\subfloat[IACA-DI (1hr)]{
		\includegraphics[width=1.5in]{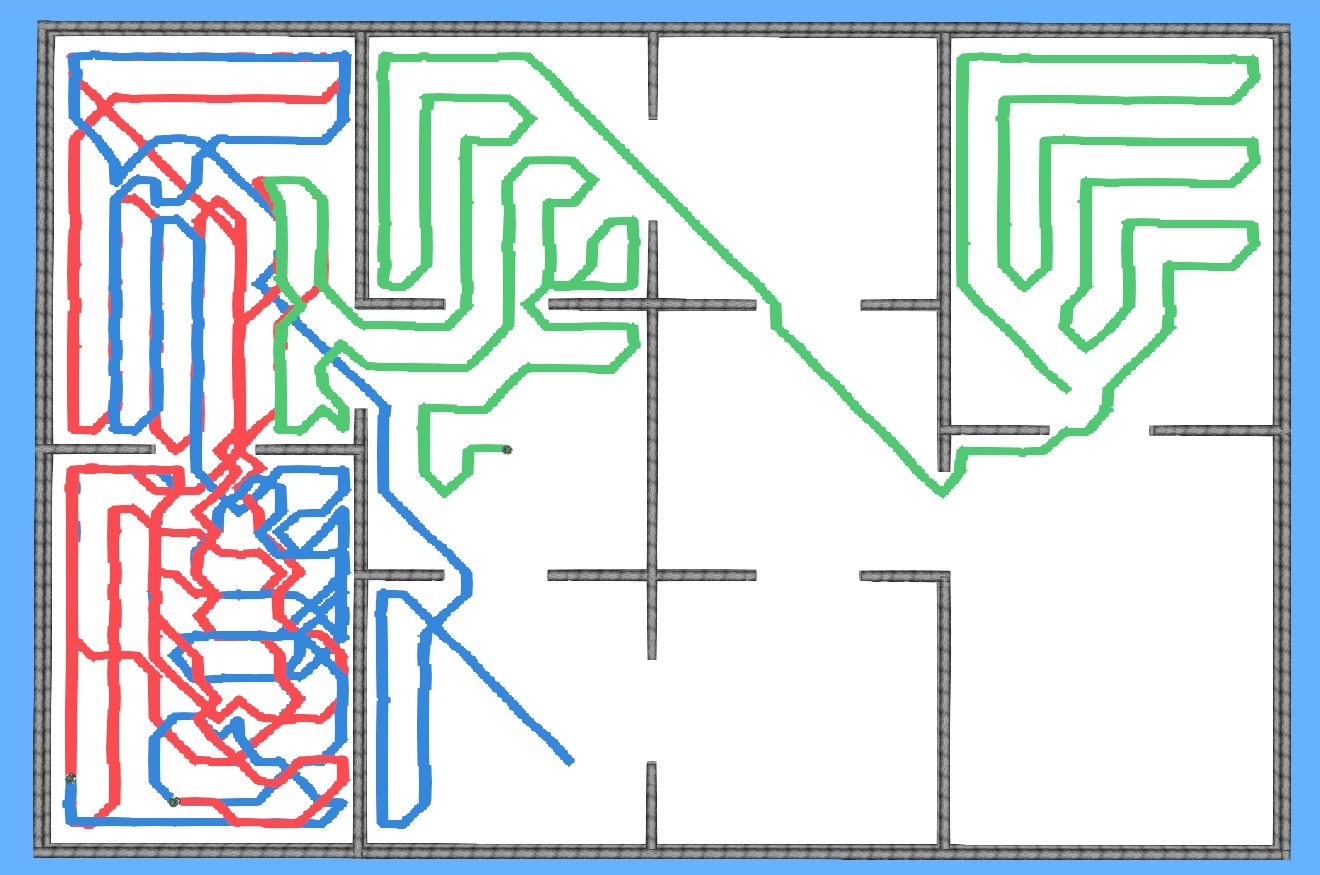}}
	\hfil
	\subfloat[IACA-DI (5hrs)]{
		\includegraphics[width=1.5in]{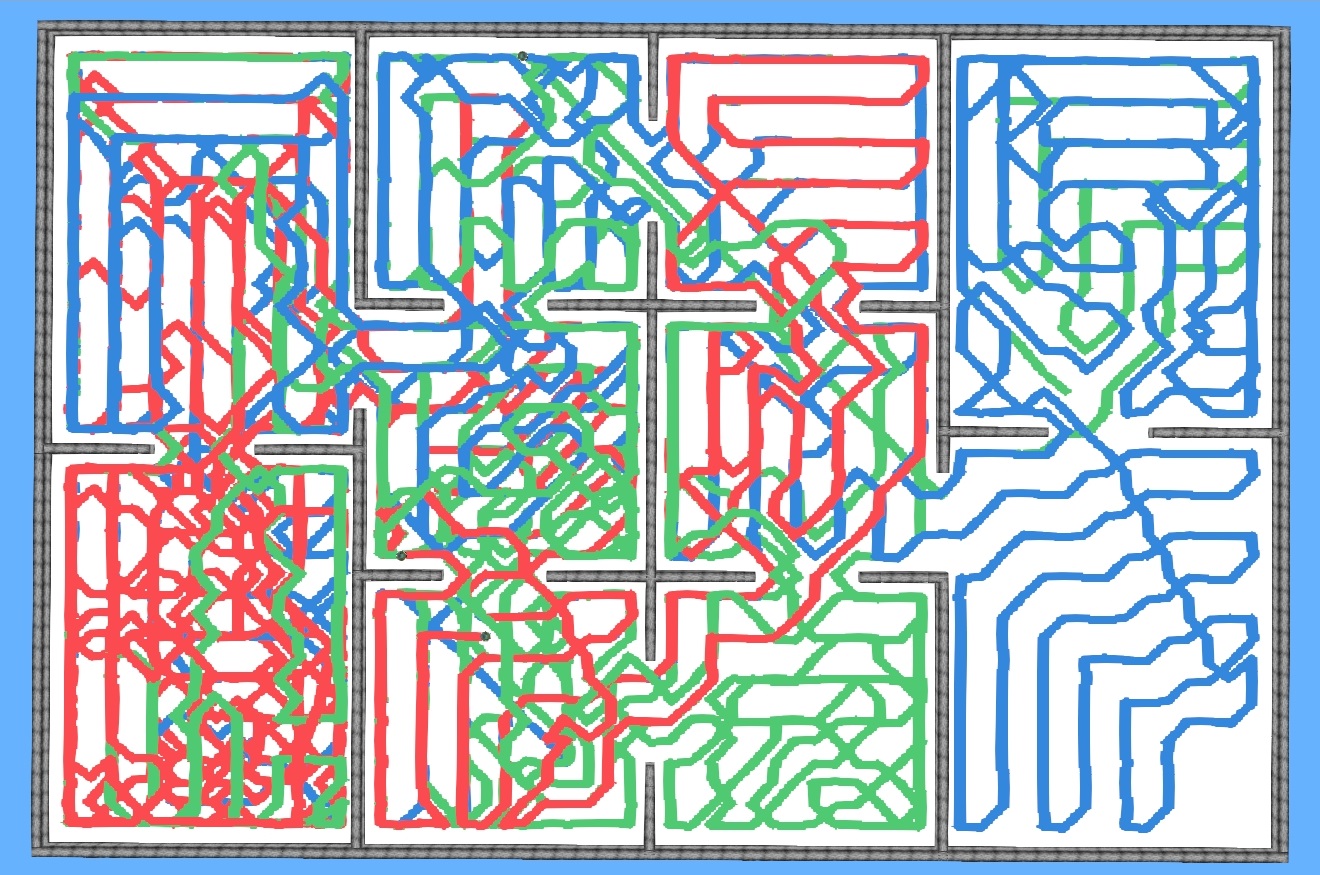}}	
	\hfil
	\subfloat[IACA-DI (10hrs)]{
		\includegraphics[width=1.5in]{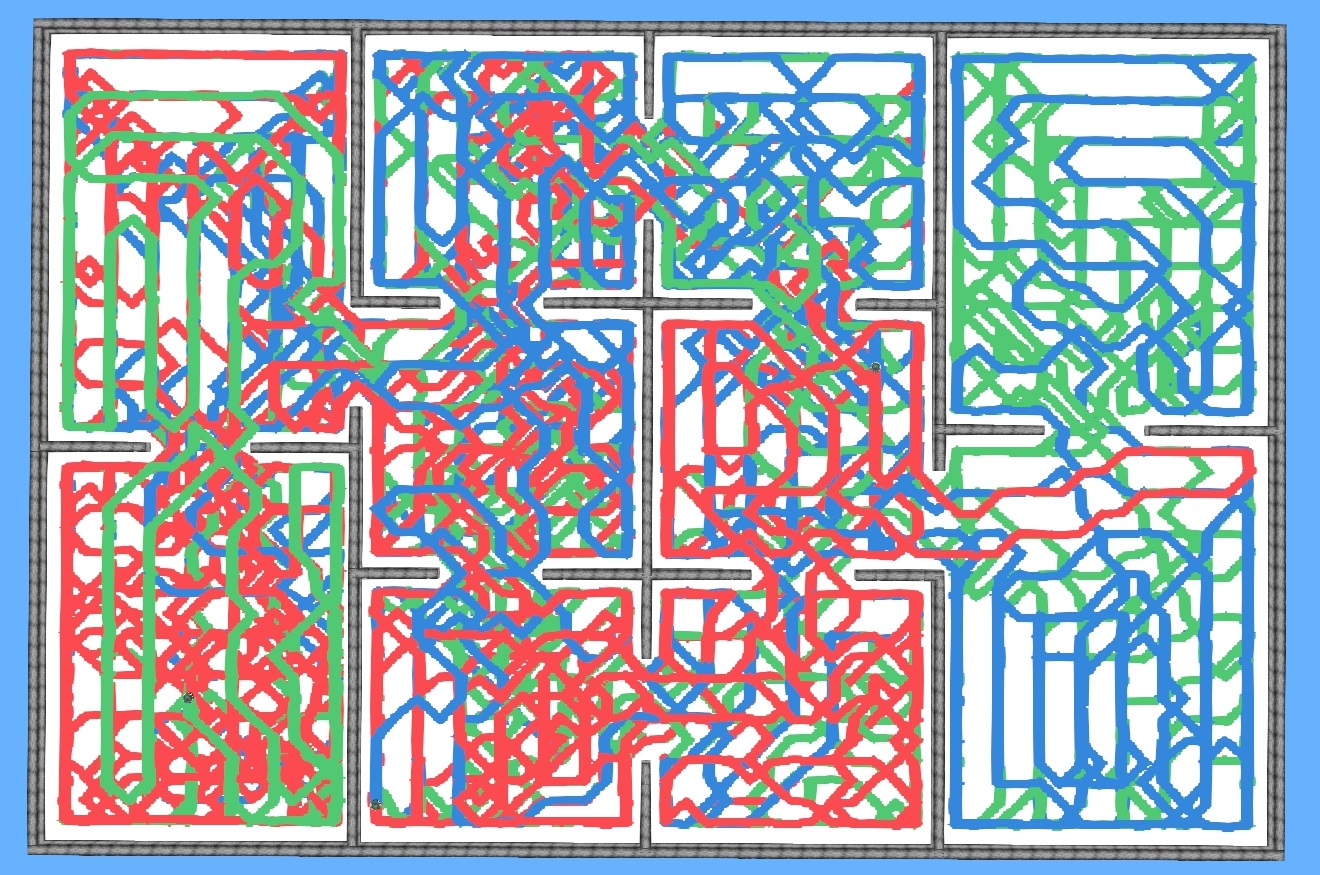}}	
	\caption{Experiment with trails: environment E3' and 3 robots (each robot is represented by a RGB colour) with a homogeneous movement strategy (Determinisitc). Figures (a), (b), (c) and (d) illustrates the performance of the PheroCom model, and figures (e), (f), (g) and (h) the IACA-DI model.}
	\label{deterministic_trails}
\end{figure}

\begin{figure}[t]
	\centering
	\subfloat[PheroCom (5min)]{
		\includegraphics[width=1.5in]{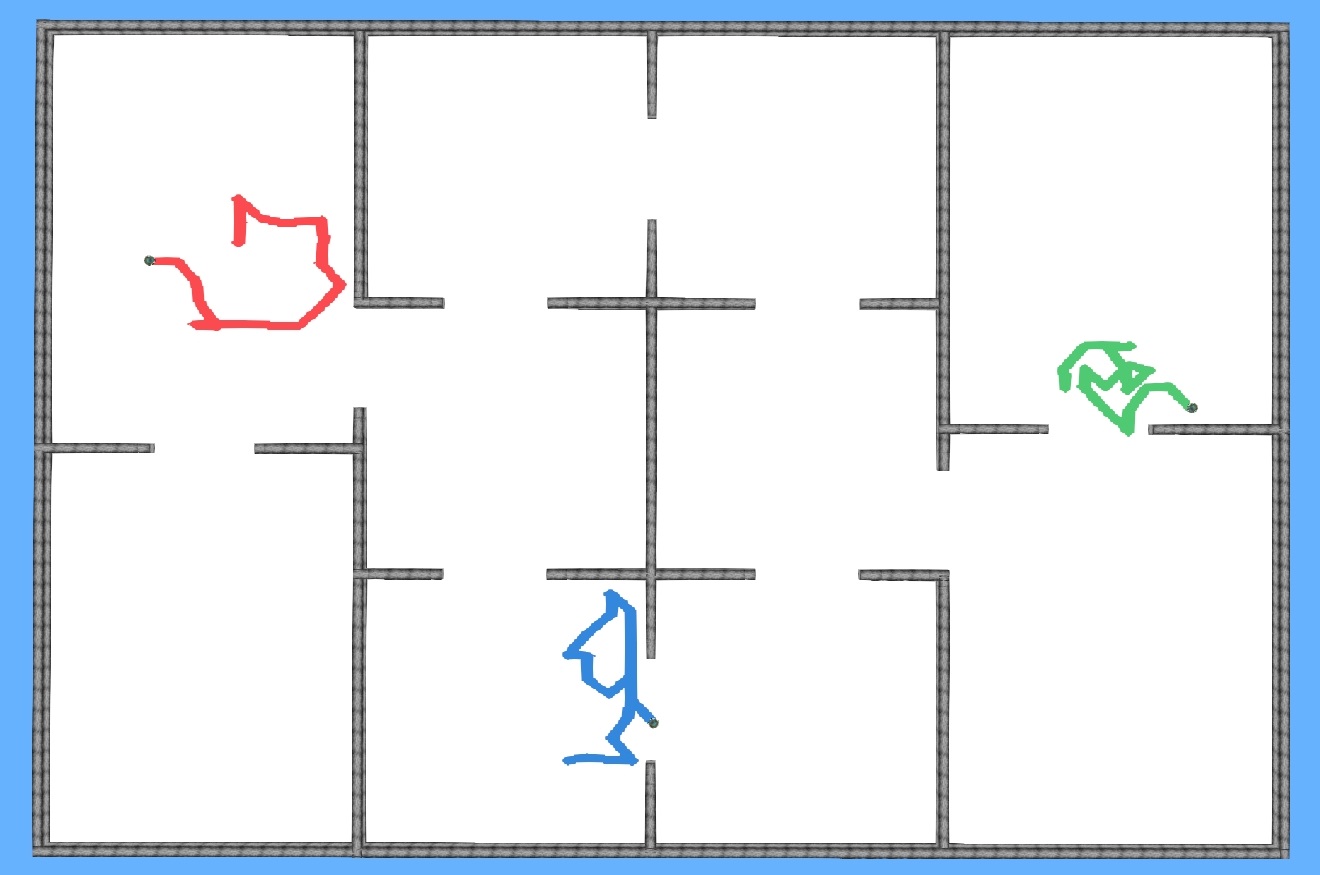}}
	\hfil
	\subfloat[PheroCom (1hr)]{
		\includegraphics[width=1.5in]{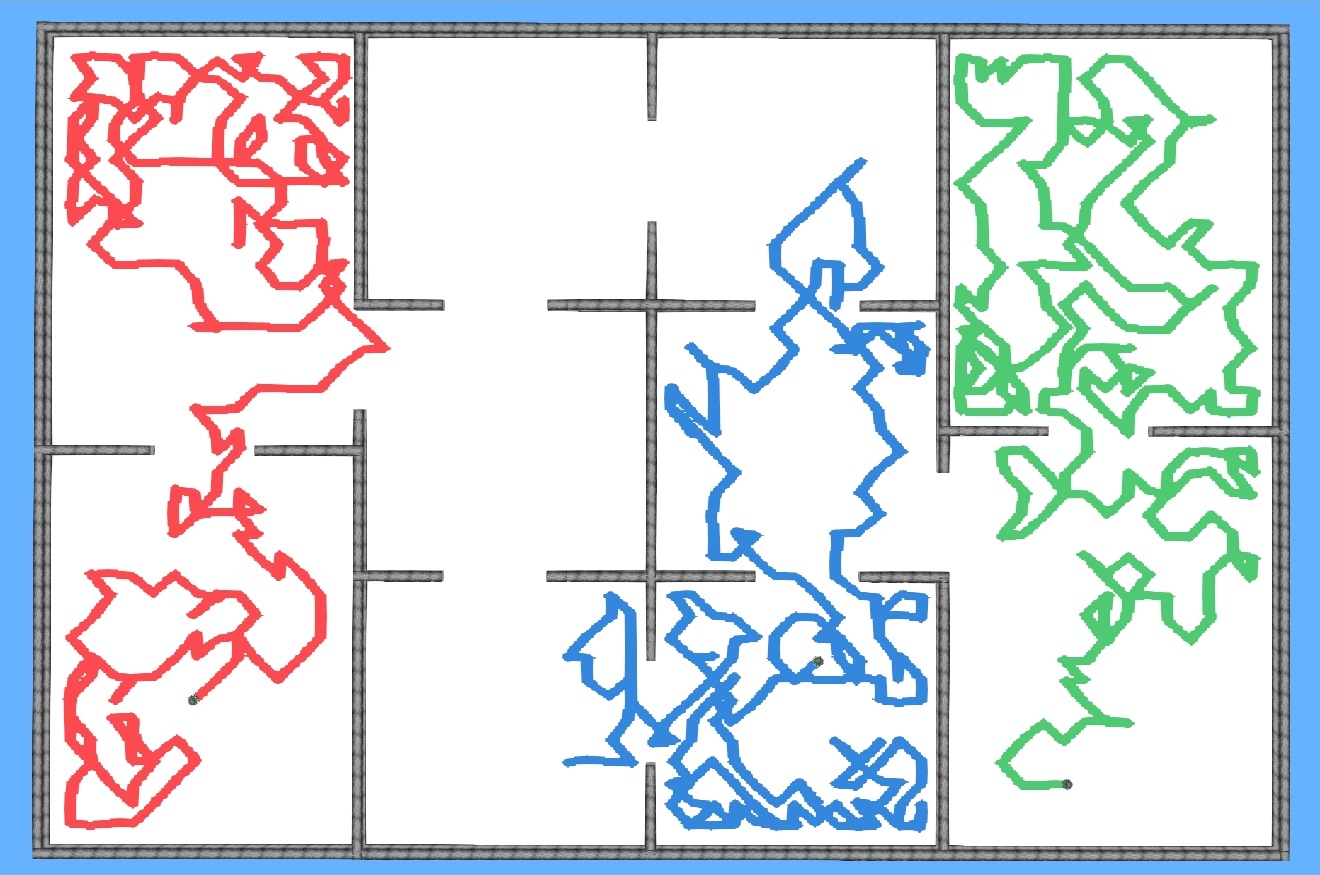}}
	\hfil
	\subfloat[PheroCom (5hrs)]{
		\includegraphics[width=1.5in]{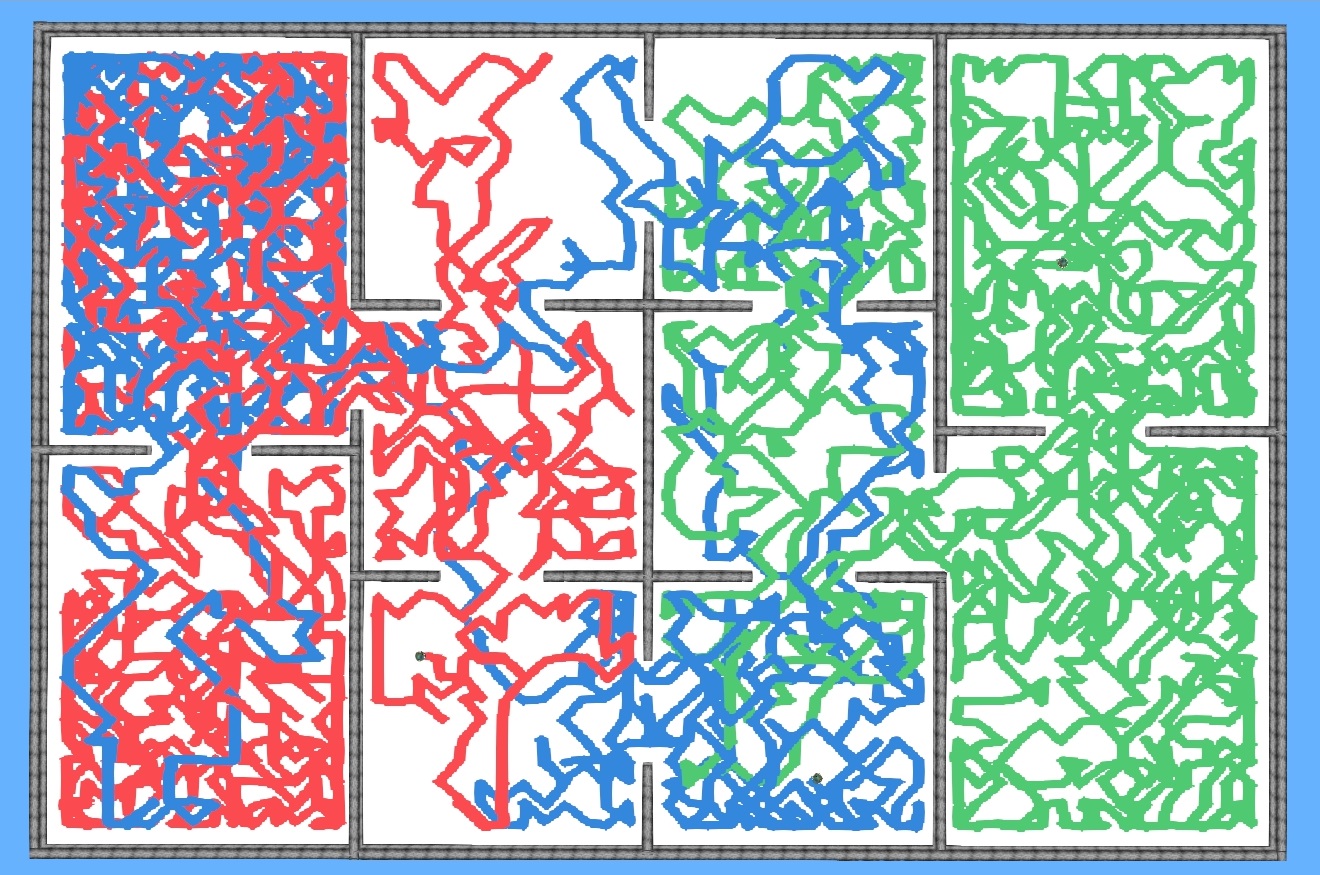}}
	\hfil
	\subfloat[PheroCom (10hrs)]{
		\includegraphics[width=1.5in]{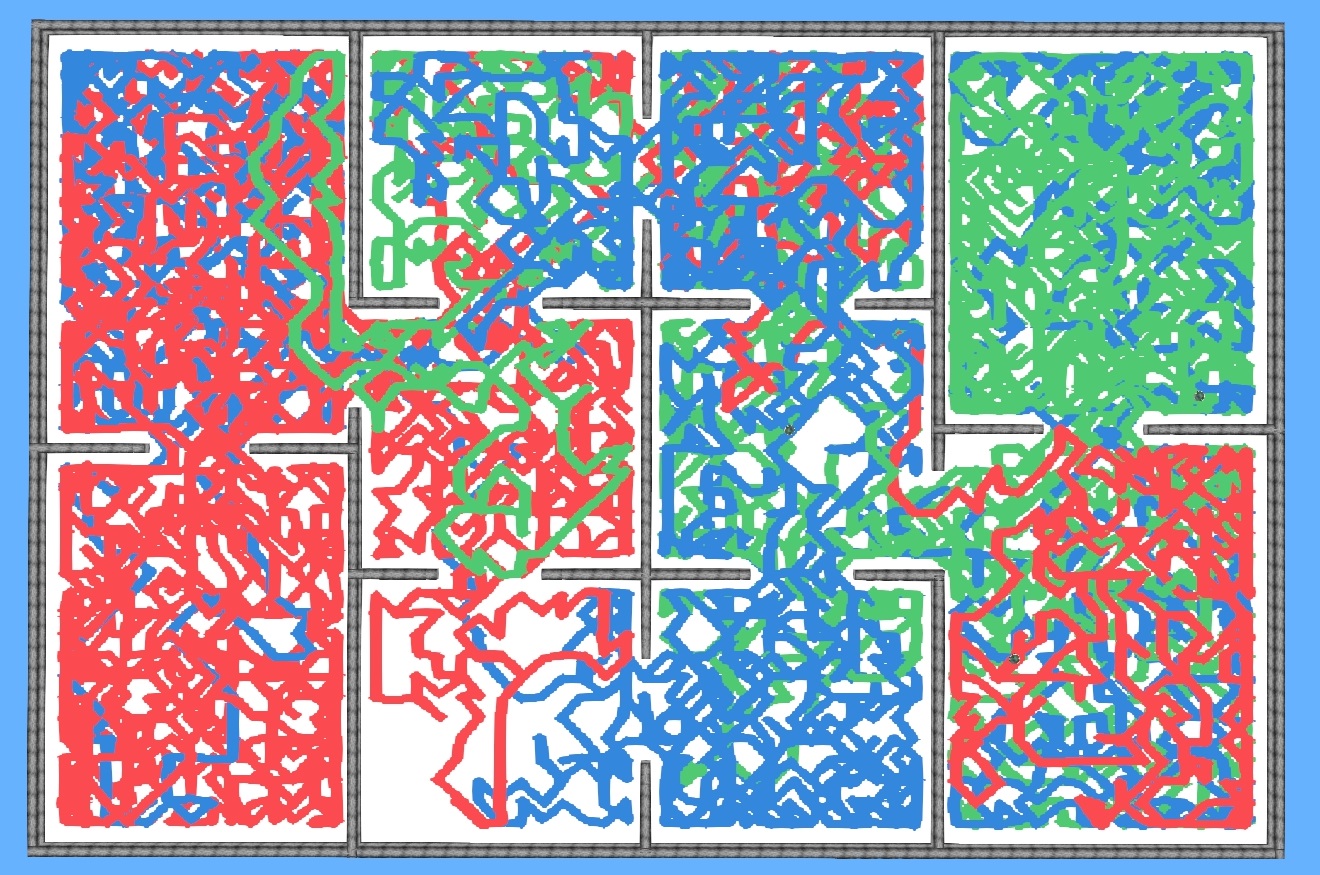}}
		
	\subfloat[IACA-DI (5min)]{
		\includegraphics[width=1.5in]{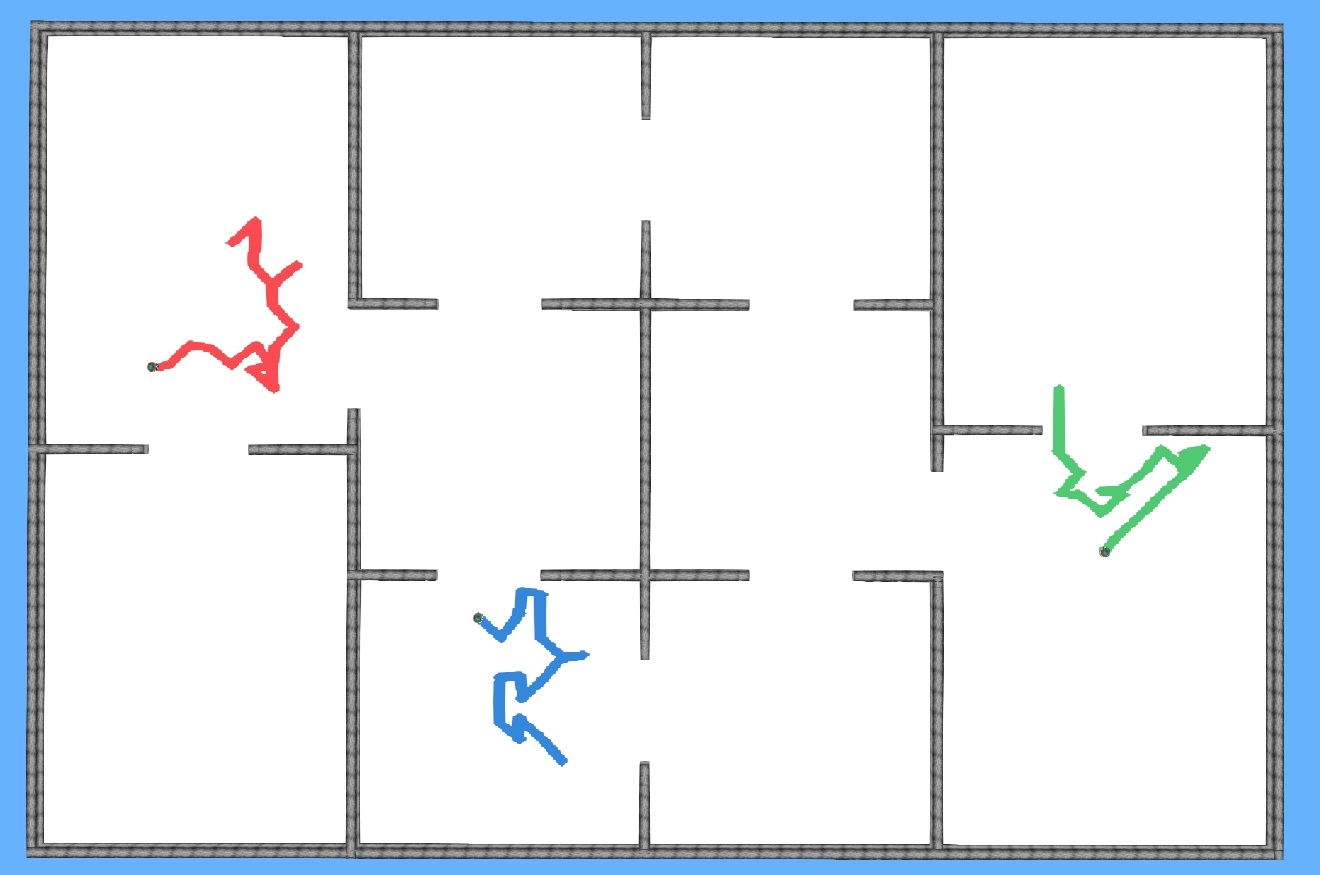}}
	\hfil
	\subfloat[IACA-DI (1hr)]{
		\includegraphics[width=1.5in]{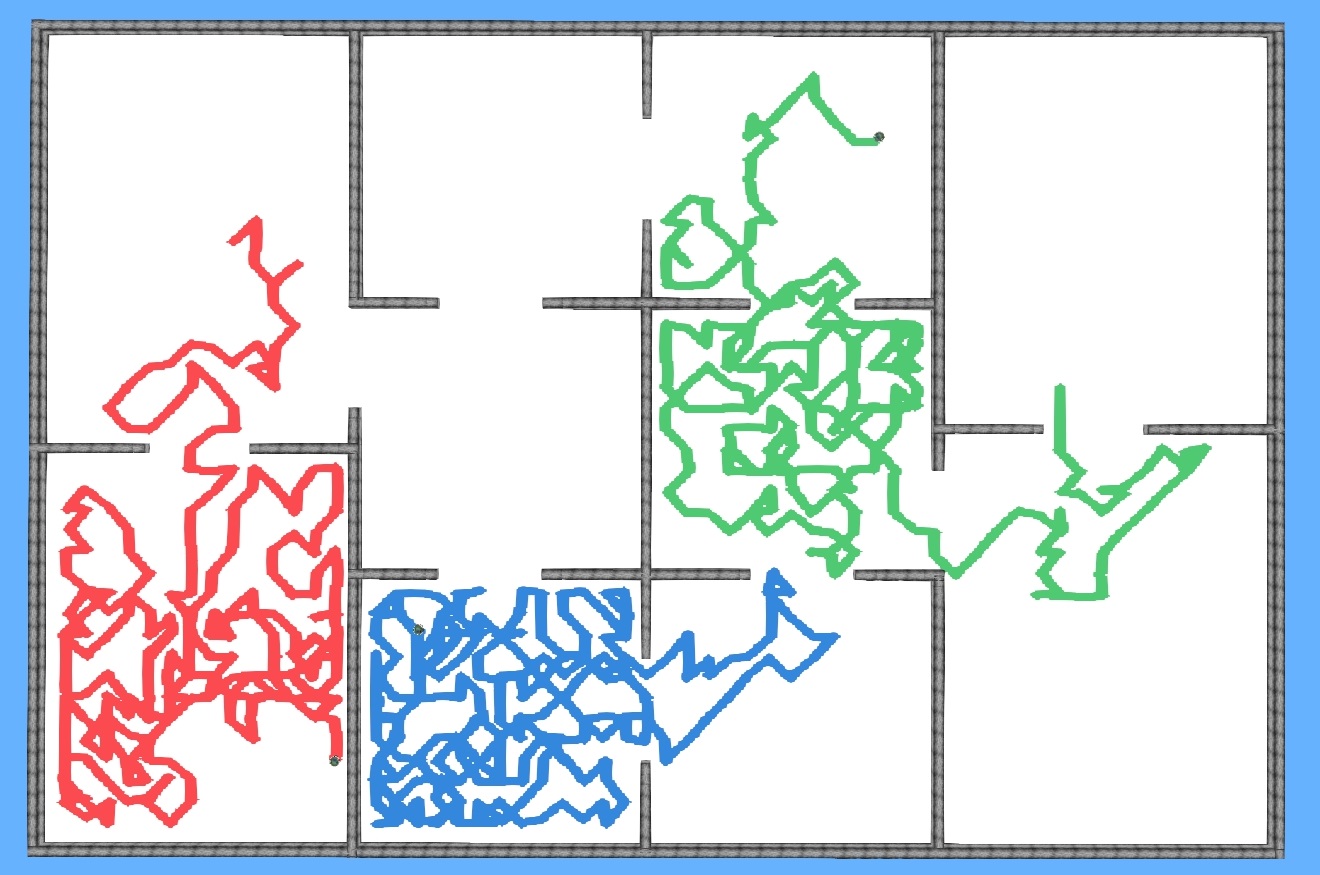}}
	\hfil
	\subfloat[IACA-DI (5hrs)]{
		\includegraphics[width=1.5in]{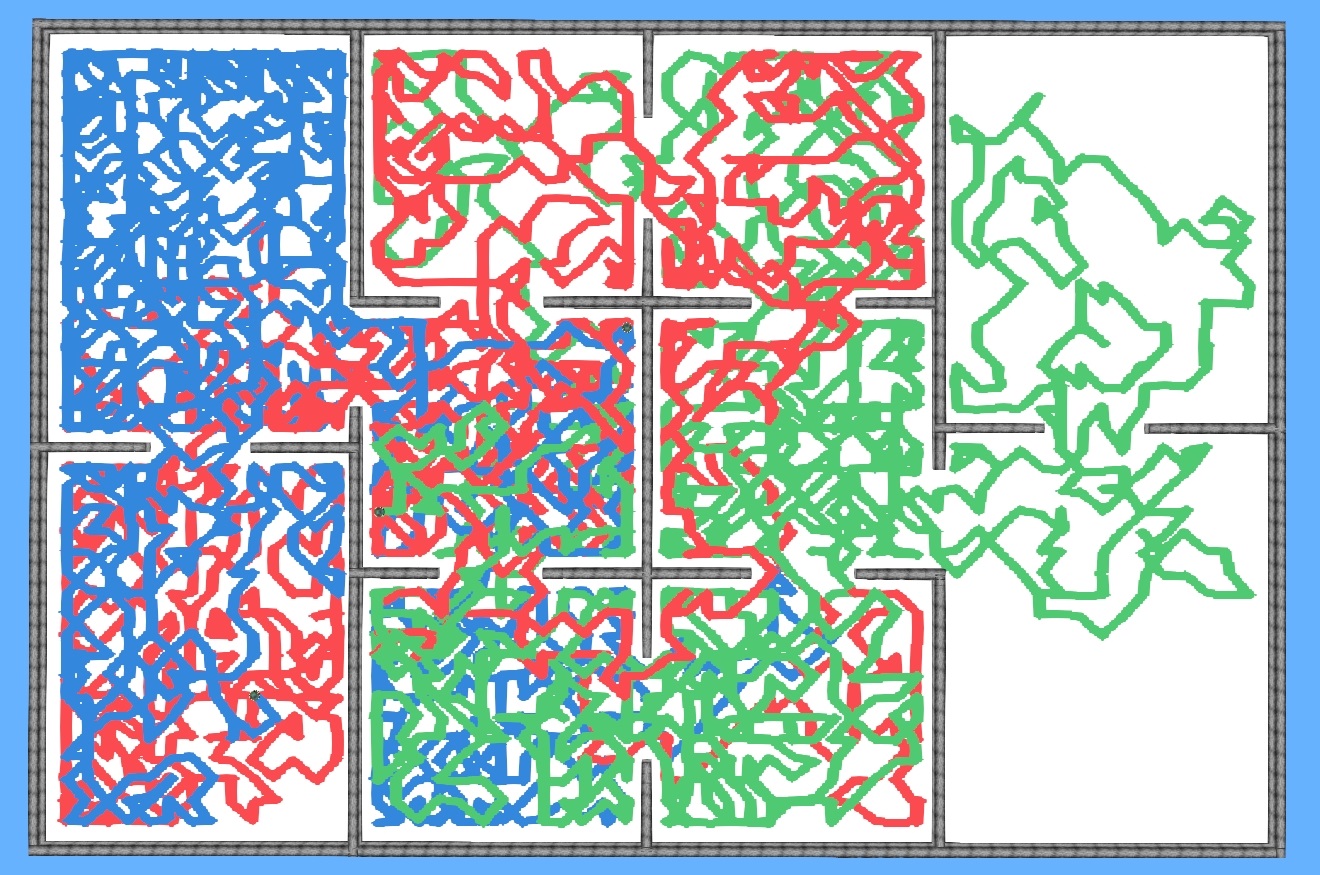}}
	\hfil
	\subfloat[IACA-DI (10hrs)]{
		\includegraphics[width=1.5in]{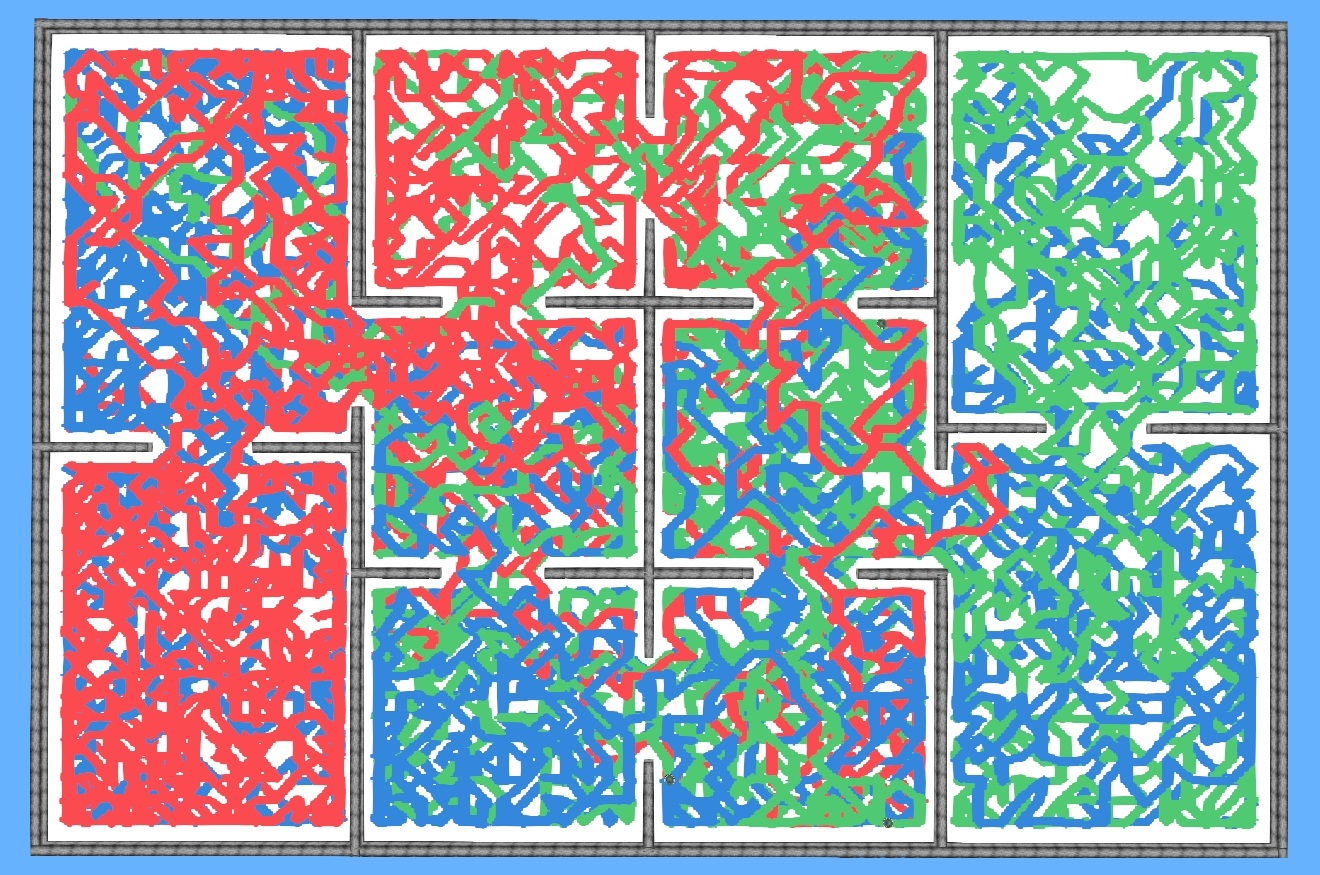}}
	\caption{Experiment with trails: environment E3' and 3 robots (each robot is represented by a RGB colour) with a homogeneous movement strategy (Inertial Probabilistic). Figures (a), (b), (c) and (d) illustrates the performance of the PheroCom model, and figures (e), (f), (g) and (h) the IACA-DI model.}
	\label{inertial_trails}
\end{figure}

\begin{figure}[t]
	\centering
	\subfloat[PheroCom (5min)]{
		\includegraphics[width=1.5in]{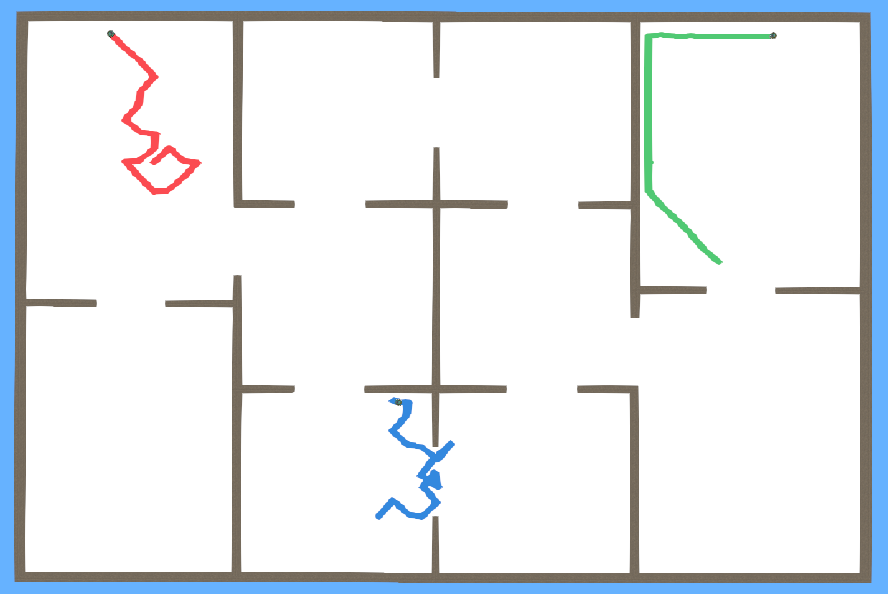}}
	\hfil
	\subfloat[PheroCom (1hr)]{
		\includegraphics[width=1.5in]{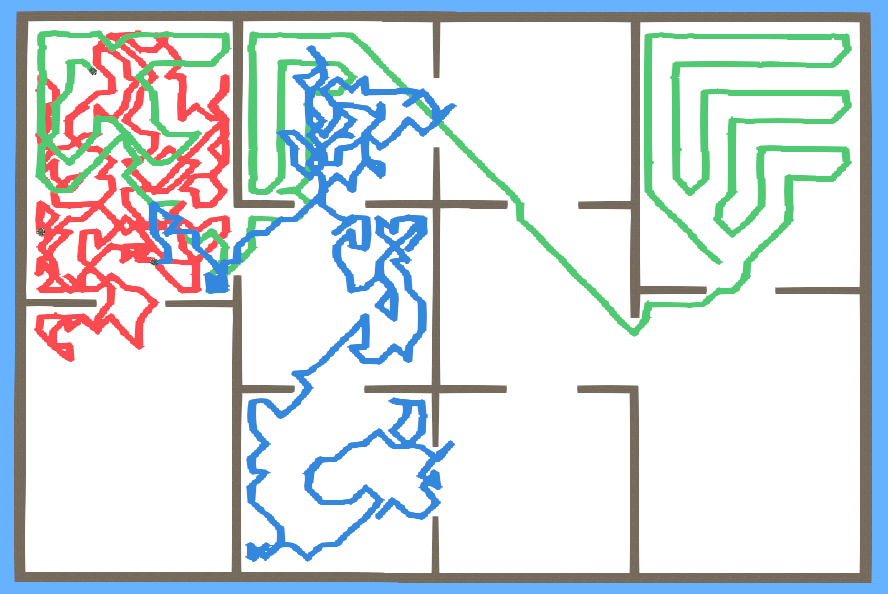}}
	\hfil
	\subfloat[PheroCom (5hrs)]{
		\includegraphics[width=1.5in]{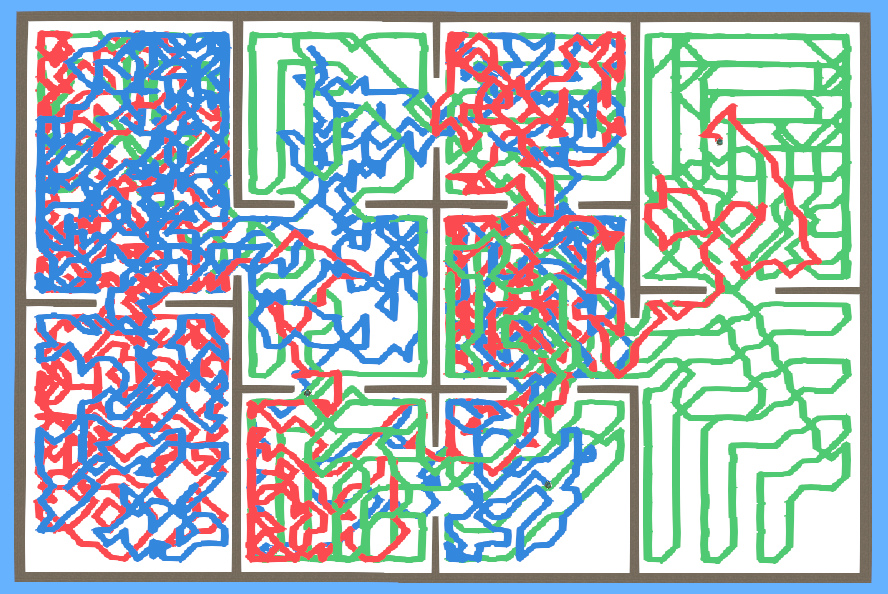}}
	\hfil
	\subfloat[PheroCom (10hrs)]{
		\includegraphics[width=1.5in]{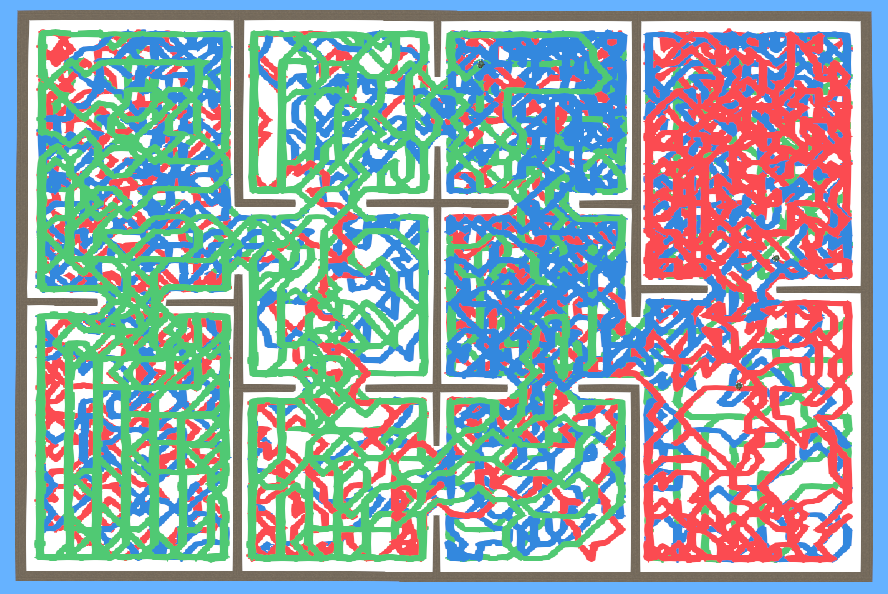}}
		
	\subfloat[IACA-DI (5min)]{
		\includegraphics[width=1.5in]{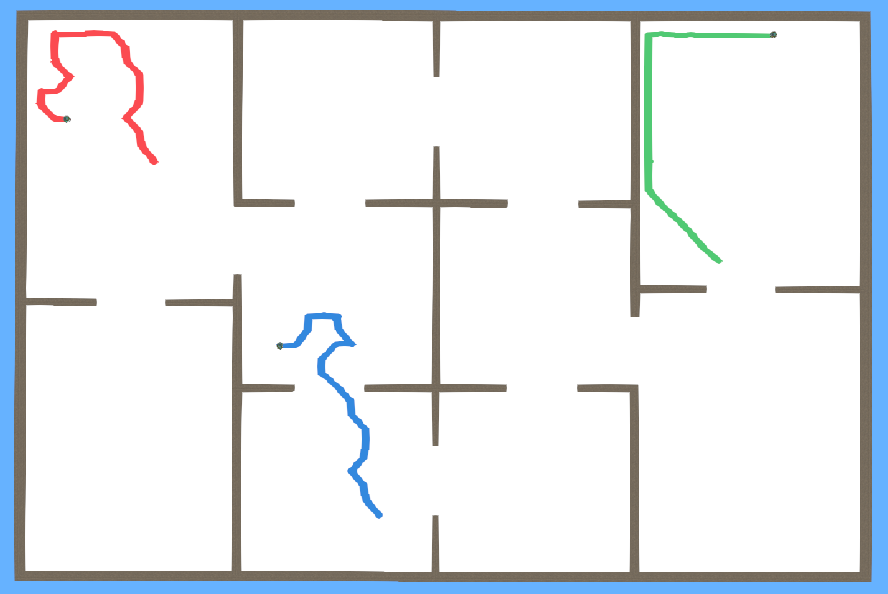}}
	\hfil
	\subfloat[IACA-DI (1hr)]{
		\includegraphics[width=1.5in]{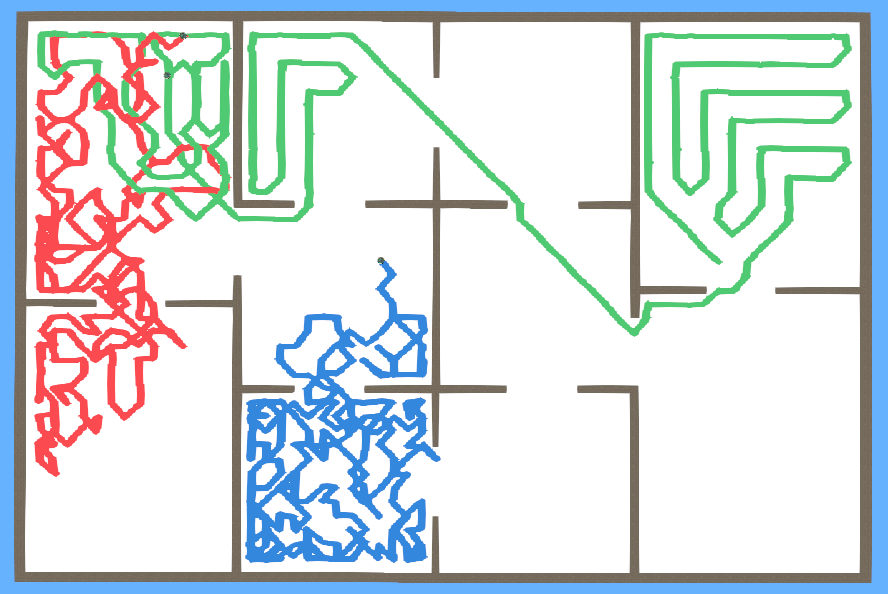}}
	\hfil
	\subfloat[IACA-DI (5hrs)]{
		\includegraphics[width=1.5in]{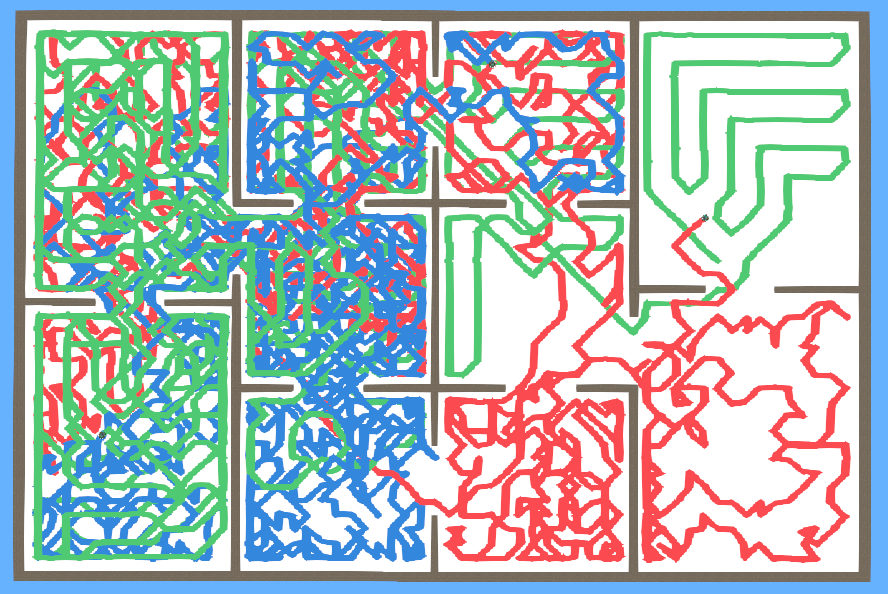}}
	\hfil
	\subfloat[IACA-DI (10hrs)]{
		\includegraphics[width=1.5in]{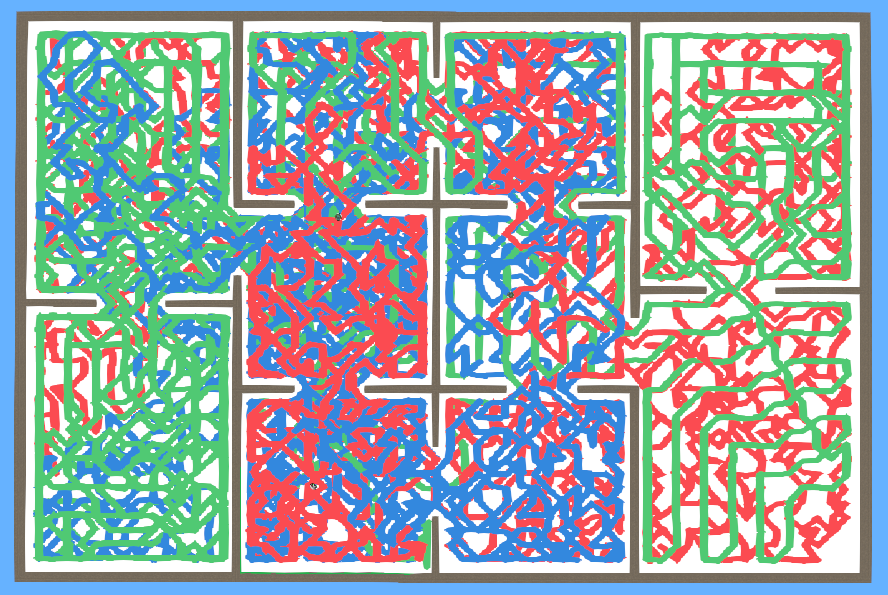}}
	\caption{Experiment with trails: environment E3' and 3 robots (each robot is represented by a RGB colour) with a heterogeneous movement strategy (\nicefrac{2}{3} of the robots with inertial strategy and \nicefrac{1}{3} with deterministic strategy). Figures (a), (b), (c) and (d) illustrates the performance of the PheroCom model, and figures (e), (f), (g) and (h) the IACA-DI model.}
	\label{heterogeneous_trails}
\end{figure}

Figures~\ref{random_trails}, \ref{deterministic_trails}, \ref{inertial_trails} and \ref{heterogeneous_trails} illustrate the experiments with trails on the Webots platform (simulations videos online\footnote{\url{https://youtube.com/playlist?list=PLhS2nTblfgy54mwSvWamQf36juvMpYZRu}}). These experiments make it possible to analyse the model considering: (i) continuous displacements throughout the environment; (ii) the possibility of collisions; and (iii) to allow an evaluation during the execution of the task and not just observing final results (since previous experiments do not have a graphical representation, precisely to speed up the simulations, the evaluation of a model with evolutionary characteristics during execution becomes quite complex).

Applying the random strategy (Fig.~\ref{random_trails}), one can observe the struggle of the robots to spread across the environment. This is due to the concentration of just one colour in some areas, depicting the presence of local repetitive circular motions. Since the random strategy allows movements in any direction with the same probability, even in those directions that lead to recently visited areas, the robots no longer monitor areas with higher priorities. With five minutes of simulation $(t = 5min)$, it can be noted the initial positions of the robots. With one hour $(t = 1h)$ two robots only changed rooms once (red and green). Around five hours $(t = 5h)$, it is still possible to observe that robots rarely changed rooms. At the end of the simulation $(t = 10h)$, there is a territory demarcation built by each robot, which remains around their initial coordinates. This behaviour highlights the low number of task-points achieved by the random strategy in our previous work~\cite{tinoco2019heterogeneous}.

In turn, applying the deterministic strategy (Fig.~\ref{deterministic_trails}), has produced the best results considering environmental coverage. It can be seen that robots form linear trails and hardly overlap paths when visiting a room, unless an interference occurs. Nevertheless, this behaviour makes the movements of the robots highly predictable, and predictability is not a desirable feature whether the task being performed is surveillance. On the other hand, if the task is, for example, exploration or foraging, a predictable behaviour is not an issue. With an hour of simulation $(t = 1h)$, only one room was not visited by the swarm. At five hours $(t = 5h)$, all rooms had already been visited, although there is slightly higher coverage when the IACA-DI model is applied. This is mainly due to the sensitivity of the deterministic strategy to the pheromone concentration in the environment, as it was observed in the experiments of our previous work~\cite{tinoco2018pheromone}. In ten hours of simulation $(t = 10h)$, both models, PheroCom and IACA-DI, have presented a homogeneous visual coverage through all rooms of the environment.

When the inertial strategy is applied (Fig.~\ref{inertial_trails}), the swarm's behaviour, considering the scattering, is superior compared to the outcomes of the random and deterministic strategies. Although in this strategy robots are likely to pass through areas that were recently visited, this probability is much lower when compared to the random strategy. This makes the robots to spread more easily throughout the environment and, beyond that, their movements do not have predictability, as occurs in the deterministic strategy. Starting from one hour of simulation $(t = 1h)$, there are only two rooms left to be visited. With five hours $(t = 5h)$, all rooms were visited, and it is possible to notice that robots passed through areas that other robots have already visited in the past, due to overlapping trails. Although it is an example of execution, it is worth mentioning that, at this point, the PheroCom model covered a larger area compared to the IACA-DI model. Completing ten hours of simulation $(t = 10h)$, the entire environment was covered by both models.

Finally, the last variation has applied a heterogeneous movement strategy (Fig.~\ref{heterogeneous_trails}), in which \nicefrac{2}{3} of the robots have the inertial strategy and \nicefrac{1}{3} the deterministic strategy. In this case, the robots with blue and red markers have the inertial strategy, while the green one has the deterministic strategy. As mentioned before, the main objective of this combination is to associate the main advantages of both strategies~\cite{tinoco2019heterogeneous}. Results have shown that the PheroCom and the IACA-DI models maintained a similar behaviour. With one hour of simulation $(t = 1h)$, every room was already visited by the PheroCom model, and only one left to be visited by IACA-DI model. In the charts of five $(t = 5h)$ and ten hours $(t = 10h)$, all rooms were visited with a good scattering throughout the environment.

\section{Assessment and Discussion}
\label{sec:assessment-discussion}
The assessments and discussions are carried out taking into account the basic characteristics of the proposed model, advantages and disadvantages, as well as the localisation challenge to be faced in its real application.

\subsection{Overview of the PheroCom model}
As an initial discussion, it is worth going back to the inspirational aspects of ViBIT protocol (see Sec.~\ref{sec:bio-inspiration}). The objective is not to propose a new metaphor for a communication protocol, but to propose a protocol sub-type, i.e., it is within the scope of gossiping protocols and broadcast communication. There are three points to be considered: (i) the mimicry of vibroacoustic communication is important to emphasise that, in biological systems, there are also forms of direct communication. Although many works state that communication within a swarm must be indirect, the opposite possibility exists. Moreover, the existence of data transmissions through sounds would allow the application of wireless communication, what would keep the characteristics of swarms; (ii) there are some fundamental differences between our proposal and a broadcast communication. In our model, there is a virtual communication radius, and it does not portray the real radius of the robots' wireless mechanism, but rather an information exchange limiter. This limiter is extremely important to ensure that robots do not exchange information beyond what is needed, since this information is equivalent to the pheromone data that is within its virtual radius. Furthermore, in broadcast, there is no propagation of information, since boundaries within the local network are not physically defined; and, finally, (iii) there is an influence of models based on gossiping, as they have as their main objective to propagate data (as in epidemic algorithms to replicate databases). This propagation offloads the initial transmitter through subsequent transmissions. Besides, there is a similarity between the characteristics of gossip-based protocols and swarm intelligence, since both can be characterised as scalable, fault-tolerant, robust, convergent and consistent, extremely decentralised, etc. These characteristics are important for swarms, and, in our proposal, the ease for decentralisation has stood out. Nevertheless, gossiping protocols are peer-to-peer, i.e., there is a connection between the transmitter and the receiver, and the information propagated in most models that rely on gossiping is not mutable. In our model, there is no peer-to-peer connection and the disseminated data contains information from the initial transmitter, however, this data is composed (mutated) by the data from each current transmitter.

Beyond that, it is necessary to open a general assessment on the strengths and weaknesses of the PheroCom model. Its main advantage is the ability to simulate pheromone dynamics without the need of intelligent environments, additional hardware or chemical applications. This simulation is fully virtualised, decentralised and asynchronous. As a result, in addition to the accuracy in the pheromone simulation, the model is easy to implement and extremely accessible. Nevertheless, it is important to remember that the application of wireless communications results in higher bandwidth/energy consumption, being the latter a scarce resource for robotics. Besides, considering that robots need to be in the neighbourhood of each other for information exchange, in some scenarios the emergence of global behaviour is not fully guaranteed (e.g., the coverage problem with a low-density swarm, or foraging over long distances making robots less likely to find each other). It is also worth noting that we took for granted that the model has the fundamental characteristics of swarm robotics (robustness, flexibility and scalability), however, in a real world application, other points must be considered (e.g., external attacks and resiliency to Byzantine robots) to maintain these characteristics.

Finally, the comparative analysis between the PheroCom and the IACA-DI models has proved to be quite consistent. This is due to the fact that, besides keeping the same information for all robots, the IACA-DI model performs a larger number of communications and data exchanges to keep this information available and up-to-date. Thus, since PheroCom model (decentralised, asynchronous and performing a smaller amount of data dissemination) has achieved a similar performance, it is possible to conclude that its outcomes exceed those presented by the IACA-DI. It is noteworthy that, simulations on the Webots platform used only three robots because it required a high processing capacity. However, as seen in the experiments on the MaSS, the PheroCom model is prepared to scale the number of robots to a swarm, without requiring any changes to its source-code. In addition, although the time of the experiments is long, the robotic platform used (e-puck) is not specific for surveillance \cite{mondada2009puck}. The time of the simulations could be drastically reduced by adapting the PheroCom model to ground monitoring robots or, even better, to a swarm of UAVs (Unmanned Aerial Vehicles).

\subsection{Considerations in the spatial localisation challenge}
The path of our research is based on mimicking strong and punctual attributes of biological systems (in this case, intelligent swarms) into robotic systems. This must be done in such a way that essential attributes of these biological systems are not lost. Furthermore, it is worth considering that some aspects of biological systems are extremely difficult to replicate due to their organic complexity (e.g., pheromone secretion, sensing glands, and the pheromone's chemical dynamics). On the other hand, some aspects could be strengthened with the technology we have at our disposal (e.g., availability of a large amount of memory and applications with wireless communication), always keeping the fundamental characteristics of the swarms.

Social insects do not record a sketch of the environment, but the current robotic platforms would be fully capable of doing so, which has allowed us to propose a viable method to conduct pheromone mimicry. Considering our proposed model PheroCom, maintaining a pheromone grid and carrying out its maintenance would be an extremely simple task for the robots. It is noteworthy that, the purpose of the pheromone grid is only to represent the environment, not to describe it with extreme precision. Thus, its granularity could be adapted according to the amount of memory available, i.e., the size of the cells could change. Evidently, with a smaller granularity, the efficiency in the execution of the task would be proportionally and negatively affected, however, the effectiveness would not be distressed.

The obstacle that arises from this strategy is the need for an individual localisation system. In robotics, there are several techniques that allow some type of geographic location (e.g., odometry and landmarks recognition). However, individually, these techniques do not deliver a satisfactory accuracy. This gets more complicated when it comes to swarm robotics. In our previous work~\cite{tinoco2017improved}, it was applied a localisation system based on a centralising agent, however, this breaks robustness, essential to swarms. Nevertheless, although it is not within the scope of this work, as can be seen in the works of \citeonline{zhu2020formation} and \citeonline{jamshidpey2020multi}, a localisation system based on UAVs swarms could be responsible for providing the location of the ground robots. This strategy would guarantee the necessary prerogatives of Swarm Intelligence, with a reasonably precise geolocation. Furthermore, as well as in the granularity of the cells, there is no need for extreme precision, i.e., robots do not need access to a perfect global localisation. Instead, robots only need to have the perception of the concentration of pheromone in their neighbourhood to carry out the decision-making process.

\section{Conclusion and Future Work}
\label{conclusions}
\vspace{-0.1cm}
This work has proposed and investigated a model to virtually simulate, represent and control the dynamics of stigmergic substances (pheromone) used by social animals in communication and self-organisation. The main objective of the proposition of this model is related to its application in the coordination of swarms of robots. Denominated as PheroCom, its inspiration is intrinsically related to the way ants transmit information by area through vibroacoustic signals. The inspiration in vibroacoustic signals allowed the proposition of an information dissemination protocol for the PheroCom model, called Vibroacoustic Based Indirect Transmission (ViBIT). The PheroCom model is decentralised and asynchronous, maintaining the major characteristics of swarm intelligence, different from our precursor model IACA-DI~\cite{tinoco2017improved}, which in turn was centralised and demanded the synchronisation of robots' actions. In this work, the PheroCom model was applied to the surveillance task in indoor environments. However, it can be adapted to other tasks, e.g., exploring unknown environments, foraging, search/rescue, among others.

In the PheroCom model, each robot has in its local memory a representation of the environment through a grid of cells of a cellular automaton. This grid is used for motion and pheromone control. Swarm synergy occurs as soon as the robots begin to disseminate information from their local grids. The pheromone information transmitted corresponds to the dissemination area of each robot, i.e., the size of the message depends on the size of the coverage area of the robot's communication mechanism. Moreover, each time a robot aggregates pheromone information, it updates its local map checking which information is most up-to-date. This type of information transmission made possible the proposition of a decentralised and asynchronous model, along with the emergence of a global behaviour. It is noteworthy that the information of the pheromone concentration does not represent a history of only the robot that holds it, but also of all the other robots that already interacted and contributed with the local map of this robot.

Centralisation is not a desirable feature in swarm robotics, as it undermines the effectiveness of the task since all robots rely on a centralising agent. Thus, the Pherocom model proposal was based on the following objectives: (i) autonomy of the robots, where each one has its own decision-making process; (ii) asynchronicity, independent of third party decisions; (iii) occurrence of local interactions only, without global knowledge of the pheromone information; (iv) robots cooperate with each other to achieve a common goal; and (v) unlike the IACA-DI model, there is no global controller.

The main experiments and analysis have shown that the model is a relevant alternative to represent, in a decentralised and asynchronous way, the pheromone dynamics to coordinate a swarm of robots. This conclusion is mainly related to the fact that the results of the PheroCom model are similar to those achieved by the centralised model (IACA-DI). As already mentioned, in the centralised model, the decision-making process is based on more consistent information, nevertheless, the PheroCom model was able to achieve similar results. The behaviour of the robots in different environments with different swarm sizes was also analysed. The model proved to be stable, where the swarm only had to adapt to the new characteristics. Thus, the main objective of the work was achieved: to propose a model capable of virtualising, in an asynchronous and decentralised way, the dynamics of pheromones applied to swarm robotics.

More specifically, the experiments sought to compare two main characteristics: efficiency and cost of the models. Regarding efficiency, the PheroCom model has exhibited being able to adapt itself to different environments, swarm sizes and decision strategies. Although the deterministic strategy in PheroCom was costlier to achieve the performance of the IACA-DI model, this goal was reached by increasing a few units in its transmission radius. In its turn, the communication cost of the PheroCom model was extremely lower compared with the IACA-DI model, where the former was able to perform the same tasks with around $1.69\%$ of the amount of information transmitted by the latter. This shows that the viability of the PheroCom model is far superior. Finally, three different movement strategies were applied to validate the PheroCom model: two homogeneous (deterministic and inertial) and one heterogeneous (\nicefrac{2}{3} inertial and \nicefrac{1}{3} deterministic). In the three strategies, the PheroCom model has shown a positive adaptation and achieved a performance similar to the IACA-DI model.

Considering the promising outcomes of the PheroCom model, several paths can be followed in future works. These paths can include: (i) to search an automatic method to define the amount of information exchanged between the robots, i.e., the size of messages that robots must exchange to ensure the efficiency of the task performed; (ii) also automate the values of some system variables, such as the number of robots and the evaporation rate, both related to the dimensions and characteristics of different environments; (iii) to implement and analyse the occurrence of events during the execution of the task, e.g., robot failure, the entry of new robots and declines in communication; (iv) knowing that the model scales to any number of robots, future experiments intend to extend the results to larger teams (more than 100 robots) to more accurately characterise the dynamics of a swarm; (v) to implement the PheroCom model in a swarm-dedicated simulator (e.g., ARGoS~\cite{pinciroli2011argos}, since it allows real-time dynamical simulations of up to 10.000 robots, with a suitable compromise between fidelity and performance); (vi) to develop a single metric to evaluate model, e.g. a harmonic mean, that combines coverage homogeneity and task-points; and (vii) to adapt the PheroCom model to be applied to a swarm of unmanned aerial vehicles.

\bibliographystyle{apalike}
\bibliography{main}

\end{document}